\documentclass[journal]{IEEEtran}
\usepackage{amsmath} 
\usepackage{amssymb} 
\usepackage{amsfonts} 
\usepackage[numbers]{natbib}
\usepackage{graphicx}
\usepackage{epstopdf}
\usepackage{booktabs}
\usepackage{algorithm}
\usepackage{algorithmic}
\usepackage{tcolorbox}
\usepackage{listings}
\usepackage{makecell}
\usepackage{placeins}
\usepackage{microtype}      
\usepackage{tabularx}
\usepackage{threeparttable}
\usepackage{hyperref}
\usepackage{cleveref}
\usepackage{longtable}

\makeatletter
\let\MYcaption\@makecaption
\makeatother

\usepackage[font=footnotesize]{subcaption}

\makeatletter
\let\@makecaption\MYcaption
\makeatother

\definecolor{varcolor}{rgb}{0,0,0.6}
\definecolor{codegray}{gray}{0.95}
\lstdefinestyle{mystyle}{
    backgroundcolor=\color{codegray},
    commentstyle=\color{gray},
    keywordstyle=\color{blue},
    numberstyle=\tiny\color{gray},
    stringstyle=\color{orange},
    basicstyle=\ttfamily\footnotesize,
    breakatwhitespace=false,
    breaklines=true,
    captionpos=b,
    keepspaces=true,
    numbers=left,
    numbersep=5pt,
    showspaces=false,
    showstringspaces=false,
    showtabs=false,
    tabsize=4,
    frame=single,
    language=Python
}
\BeforeBeginEnvironment{lstlisting}{\vspace{1em}}
\DeclareMathOperator*{\Crossover}{LLM\_Crossover}
\DeclareMathOperator*{\Mutate}{LLM\_Mutate}
\crefname{listing}{Code}{Codes}
\Crefname{listing}{Code}{Codes}
\crefname{figure}{fig}{figs}
\Crefname{figure}{Fig}{Figs}

\ifCLASSINFOpdf
\else
\fi

\usepackage{stfloats}
\begin{document}
%
    \title{LLM-Meta-SR: In-Context Learning for Evolving Selection Operators in Symbolic Regression  }
%
%
%

    \author{
    Hengzhe Zhang,
    Qi Chen,~\IEEEmembership{Member,~IEEE,}
    Bing Xue,~\IEEEmembership{Fellow,~IEEE,}
    Wolfgang Banzhaf,~\IEEEmembership{Member,~IEEE,}
    Mengjie Zhang,~\IEEEmembership{Fellow,~IEEE}
    \thanks{
        H. Zhang, Q. Chen, B. Xue and M. Zhang are with the Centre for Data Science and Artificial Intelligence \& School of Engineering and Computer Science, Victoria University of Wellington, PO Box 600, Wellington 6140, New Zealand (e-mails: hengzhe.zhang@ecs.vuw.ac.nz; qi.chen@ecs.vuw.ac.nz; bing.xue@ecs.vuw.ac.nz; and mengjie.zhang@ecs.vuw.ac.nz).\\
        \indent W. Banzhaf is with the Department of Computer Science and Engineering, Michigan State University, East Lansing, MI 48824, USA (e-mails: banzhafw@msu.edu).
    }
}

%
%

    \markboth{Journal of \LaTeX\ Class Files,~Vol.~14, No.~8, August~2015}%
    {Shell \MakeLowercase{\textit{et al.}}: Bare Demo of IEEEtran.cls for IEEE Journals}
%



    \maketitle

    \begin{abstract}
        Large language models (LLMs) have revolutionized algorithm development, yet their application in symbolic regression, where algorithms automatically discover symbolic expressions from data, remains limited. In this paper, we propose a meta-learning framework that enables LLMs to automatically design selection operators for evolutionary symbolic regression algorithms. We first identify two key limitations in existing LLM-based algorithm evolution techniques: lack of semantic guidance and code bloat. The absence of semantic awareness can lead to ineffective exchange of useful code components, while bloat results in unnecessarily complex components; both can hinder evolutionary learning progress or reduce the interpretability of the designed algorithm. To address these issues, we enhance the LLM-based evolution framework for meta-symbolic regression with two key innovations: a complementary, semantics-aware selection operator and bloat control. Additionally, we embed domain knowledge into the prompt, enabling the LLM to generate more effective and contextually relevant selection operators. Our experimental results on symbolic regression benchmarks show that LLMs can devise selection operators that outperform nine expert-designed baselines, achieving state-of-the-art performance. Moreover, the evolved operator can further improve a state-of-the-art symbolic regression algorithm, achieving the best performance among 28 symbolic regression and other machine learning algorithms across 116 regression datasets. This demonstrates that LLMs can exceed expert-level algorithm design for symbolic regression.
    \end{abstract}


%
    \IEEEpeerreviewmaketitle

        \section{Introduction}
        \label{sec: Introduction}
        Symbolic regression (SR) is the task of discovering mathematical expressions that accurately model a given dataset. 
        It has the potential to produce interpretable models when appropriate complexity control mechanisms are employed, and its ability to uncover underlying relationships makes it widely applicable across various domains.
        Formally, given a dataset $\mathcal{D} = \{\mathbf{x}^{(i)}, y^{(i)}\}_{i=1}^{N}$, where $\mathbf{x}^{(i)} \in \mathbb{R}^d$ are input features and $y^{(i)} \in \mathbb{R}^m$ are target values, the goal is to find a symbolic expression $f^* \in \mathcal{F}$ from a space of candidate expressions $\mathcal{F}$ that minimizes a loss function $\mathcal{L}$:
        \begin{equation}
            f^* = \arg\min_{f \in \mathcal{F}} \mathcal{L}(f(\mathbf{x}), y).
        \end{equation}
        Many SR algorithms have been proposed, including evolutionary~\cite{fong2024symbolic}, neural~\cite{kamienny2022end}, and Monte Carlo tree search-based~\cite{yu2025symbolic} approaches. Among them, genetic programming (GP) remains one of the most widely used SR paradigms due to its flexible representation and global, gradient-free search mechanism. In recent years, there has been growing interest in hybrid methods that combine the flexibility of GP with the learning capabilities of neural networks~\cite{han2025transformer, zhang2025ragsr}.

        In both evolutionary~\cite{fong2024symbolic} and neural-evolutionary~\cite{han2025transformer, zhang2025ragsr} SR algorithms, the system follows an iterative learning process in which solutions are progressively refined by selecting promising candidates. In this process, the selection operator, which identifies candidate solutions worth refining, plays a key role in system effectiveness. However, current selection operators are typically manually designed by experts, requiring significant effort and trial-and-error, which can hinder progress. Ideally, an automated algorithm design framework capable of generating effective selection operators for SR would substantially enhance research and development efficiency in this area. Other evolutionary components—such as mutation and crossover operators—are tightly coupled to the underlying representation, whether tree-based, graph-based, or grammar-based, which limits transferability across SR systems. In contrast, the selection operator is largely representation-agnostic. Thus, this paper focuses on automated selection design without entangling the framework with representation-specific design choices.

        Large language models (LLMs) have recently emerged as powerful tools for automated heuristic design in various optimization tasks~\cite{romera2024mathematical, liu2024evolution, ye2024reevo}. Recent studies have shown that LLMs can effectively replace traditional crossover and mutation operators to generate candidate solutions in SR~\cite{romera2024mathematical, grayeli2024symbolic, zhang2025ragsr}. However, selection operators are still manually designed by experts, creating a gap in automated algorithm design. This motivates the exploration of automated selection operator design using LLMs, which is the focus of this work. We adopt a \emph{meta-learning} framework based on LLM-driven algorithm evolution to automatically design a core algorithmic component—specifically, the selection operator—in evolutionary SR, aiming to achieve strong performance across a wide range of regression tasks. In this meta-learning framework, a population of algorithms is evolved, where each individual in the population represents a complete selection operator, implemented as a piece of code that defines a selection strategy, rather than a solution to a specific SR problem. The population thus maintains a collection of candidate algorithms, and the evolutionary process aims to discover improved selection operators through crossover and mutation operations guided by LLMs.

\begin{figure}[!t]
            \centering
            \begin{minipage}[t]{0.48\columnwidth}
                \centering
                \includegraphics[width=\columnwidth, trim=5pt 5pt 5pt 5pt, clip]{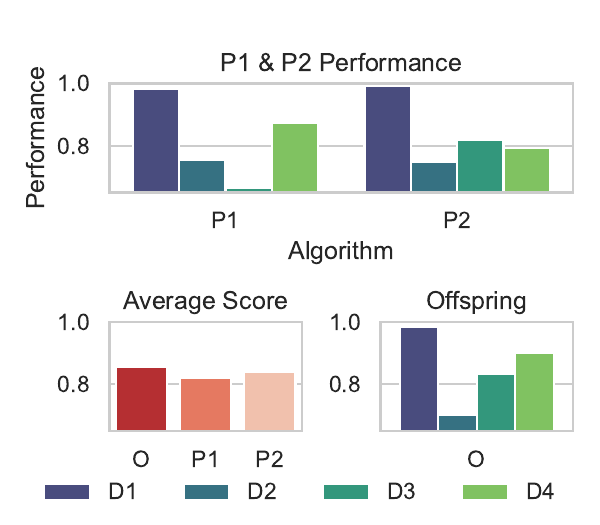}
                \caption{Illustration of fine-grained semantic differences between algorithms during meta-evolution. P1 and P2 are parent selection operator algorithms, O is an offspring algorithm, and D is a dataset.}
                \label{fig:semantics}
            \end{minipage}
            \hfill
            \begin{minipage}[t]{0.48\columnwidth}
                \centering
                \includegraphics[width=\columnwidth]{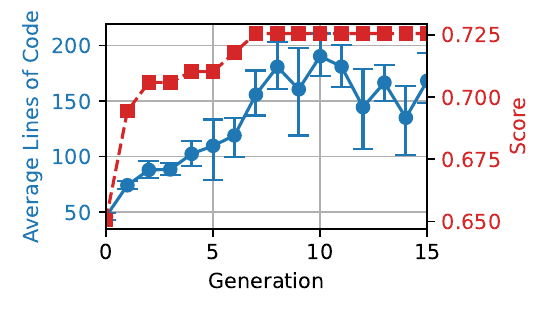}
                \caption{Average code length of solutions in the population over generations without bloat control techniques, and the score of the best solution.}
                \label{fig:bloat}
            \end{minipage}
        \end{figure}

        While LLM-guided algorithm search frameworks such as FunSearch~\cite{romera2024mathematical}, EoH~\cite{liu2024evolution}, and ReEvo~\cite{ye2024reevo} have laid the groundwork for algorithm evolution, two challenges remain underexplored in LLM-driven algorithm design. The first challenge is the \emph{underutilization of semantic information} when generating algorithms. Here, we define semantic information as the performance of an algorithm on each task instance. Typically, only the average performance across all training instances is provided~\cite{huang2025autonomous}, which overlooks fine-grained behavioral differences. As illustrated in \Cref{fig:semantics}, consider a scenario with multiple algorithms in the meta-learning population: some perform well on the first dataset but poorly on the second, while others show the opposite pattern. Despite their contrasting behaviors, these algorithms may exhibit similar average performance. In LLM-driven algorithm evolution, we present multiple algorithms to the LLM so it can infer useful patterns and generate improved variants. However, algorithms that behave similarly at the instance level offer little benefit for this purpose as they succeed and fail on the same datasets and therefore provide redundant information. In contrast, presenting complementary algorithms enables the LLM to integrate distinct strengths, increasing the likelihood of generating an algorithm that performs well on both datasets. Therefore, incorporating semantic information is crucial for effective algorithm evolution.

        The second challenge is \emph{complexity bias}, illustrated in \Cref{fig:bloat}. LLMs tend to generate overly long or intricate code during optimization, similar to code bloat in GP~\cite{banzhaf1998genetic}. This results in the accumulation of redundant or non-functional logic, which impairs interpretability, wastes a large number of tokens, and slows optimization. As shown in \Cref{fig:bloat}, such complexity can lead to performance stagnation, ultimately limiting the effectiveness of evolution.

        In this paper, we propose an LLM-driven method for meta SR that automatically designs selection operators for evolutionary SR algorithms~\footnote{Source Code: \url{https://anonymous.4open.science/r/LLM-Meta-SR/}}. The main contributions are as follows:
        \begin{itemize}
            \item We propose an LLM-driven meta SR framework to automatically design selection operators using in-context learning. The LLM learns from design and evaluation history and automatically discovers generalizable operators that consistently outperform human-designed counterparts across a wide range of unseen datasets.

            \item We identify the issue of bloat in LLM-based code generation and introduce a bloat control strategy, which improves both the interpretability of the evolved code and the effectiveness of the evolutionary process.

            \item We propose semantic feedback and complementary selection mechanisms to fully leverage semantic information during LLM-driven generation, explicitly guiding algorithm evolution, integrating effective building blocks, and enhancing learning performance.

            \item We define desirable properties of selection operators based on domain knowledge to craft more effective prompts, which can guide LLMs in generating high-quality selection operators.

            \item
            We integrate the evolved selection operator into a state-of-the-art Transformer-assisted SR algorithm, RAG-SR~\cite{zhang2025ragsr}, demonstrating that the LLM-evolved operator can enhance the performance of modern SR systems and achieve the best performance among 28 algorithms.
        \end{itemize}
        The remainder of this paper is organized as follows. \Cref{sec: Background and Related Work} reviews related work on SR and LLM-driven algorithm design. \Cref{sec: Algorithm Framework} introduces the proposed LLM-Meta-SR framework and its key components. Experimental settings are described in \Cref{sec: Experimental Settings}, results and ablation studies are presented in \Cref{sec: Experimental Results}, and further analyses are discussed in \Cref{sec: Further Analysis}. Finally, \Cref{sec: conclusion} concludes the paper and outlines future work.

        \section{Background and Related Work}
        \label{sec: Background and Related Work}

        \subsection{Selection Operators in GP}
        Selection mechanisms play a critical role in GP for SR by determining how candidate solutions are preserved and recombined during evolution. Various selection operators have been explored, including tournament selection~\cite{xie2012parent}, Boltzmann sampling~\cite{shojaee2025llmsr,romera2024mathematical}, lexicase selection~\cite{la2019probabilistic}, and complementary phenotype selection~\cite{dolin2002opposites}. Among these, tournament selection is the most commonly used due to its simplicity and efficiency; Boltzmann sampling applies a temperature-controlled softmax over fitness values to adaptively adjust selection pressure; lexicase selection rewards specialists that perform well on subsets of training instances~\cite{la2019probabilistic}; and complementary phenotype selection leverages synergistic individuals during crossover~\cite{dolin2002opposites}. Beyond these, novelty search~\cite{lehman2010efficiently} embodies a distinct philosophy that uses behavioral novelty rather than problem-solving performance as the fitness metric, revealing that diversity plays an important role in evolutionary search.

        From a diversity perspective, lexicase selection is a representative operator and has been widely used and studied by researchers~\cite{cava2018learning,zhang2025ragsr}. Inspired by the lexicase paradigm, several extensions have been proposed to enhance its robustness and adaptability, including probabilistic lexicase selection (PLex)~\cite{ding2023probabilistic}, lexicase-like selection via diverse aggregation (DALex)~\cite{ni2024dalex}, $\varepsilon$-lexicase selection with dynamic split (D-Split)~\cite{imai2024minimum},
        random down-sampled lexicase selection~\cite{hernandez2019random}, informed down-sampled lexicase selection~\cite{boldi2024informed},
        and random down-sampled tournament selection (RDS-Tour)~\cite{geigerperformance}.
        These operators share the common idea of emphasizing case-wise performance diversity while introducing various mechanisms—such as probabilistic sampling, aggregation, or adaptive thresholds—to improve effectiveness. Batch tournament selection~\cite{de2019batch} groups fitness cases into batches and applies tournaments on each batch-specific fitness, producing lexicase-like diversity and solution quality while preserving the computational efficiency of standard tournament selection.

        Besides improving the effectiveness of evolution, selection operators have also been designed to control program size, addressing the issue of bloat—the uncontrolled growth of solutions without fitness gains. A well-known approach for mitigating this effect during selection is lexicographic parsimony pressure~\cite{luke2002lexicographic}, which prioritizes smaller trees when fitness values are equal. Other methods include proportional and double tournament selection~\cite{luke2006comparison}, where the latter applies the tournament process twice—first to select by fitness, then to favor smaller individuals—showing superior performance on benchmarks such as multiplexer and SR tasks. Pareto tournament selection~\cite{kotanchek2006pursuing} randomly samples a pool of candidates and selects all nondominated individuals in that pool based on fitness–complexity Pareto dominance, providing multi-objective pressure that focuses search on the evolving Pareto front while preserving diversity. Lexicase-based selection operators have also been extended to implicitly regulate bloat~\cite{de2022lexi2,zhang2023double}. For example, Lexi$^2$~\cite{de2022lexi2} favors smaller trees in tie cases, while Double Lexicase Selection~\cite{zhang2023double} repeats the lexicase process and probabilistically retains the less complex individual. Finally, adaptive depth-limit techniques that adjust permissible program sizes based on the evolving fitness distribution~\cite{silva2009dynamic,silva2012operator} provide an adaptive way of controlling bloat during evolution.

        Despite the rich landscape of selection operators surveyed above, all existing methods are manually designed by human experts. In GP-based SR, widely used selection operators such as tournament selection~\cite{xie2012parent}, lexicase selection and its variants~\cite{la2019probabilistic,ding2023probabilistic,ni2024dalex}, and parsimony-based approaches~\cite{luke2002lexicographic,luke2006comparison} are all hand-crafted. Similarly, in LLM-based SR, methods such as FunSearch~\cite{romera2024mathematical} and LLM-SR~\cite{shojaee2025llmsr} also rely on manually designed selection mechanisms, e.g., Boltzmann sampling~\cite{romera2024mathematical, shojaee2025llmsr}, to select promising candidates during their search process. Generative-model-based SR methods such as GenSR~\cite{li2026gensr} likewise rely on manually designed selection mechanisms, e.g., truncation selection~\cite{li2026gensr}. While these hand-crafted operators typically optimize a subset of desirable properties, such as diversity maintenance, interpretability, or bloat control, each of them addresses only some of these properties. Designing a unified operator that captures all of them requires substantial domain expertise and extensive trial-and-error. An ideal selection operator should simultaneously consider diversity, interpretability, dynamic selection pressure, complementarity, and vectorization efficiency. A comprehensive comparison of existing methods with respect to these design principles is provided in \Cref{tab:related_work_properties}. The absence of such unified designs highlights the need for an automated approach to generate selection operators that jointly capture these properties---which is the central goal of this work.

        \begin{table}[!tb]
            \caption{Comprehensive comparison of selection operators with respect to the proposed design principles. A checkmark ($\checkmark$) indicates that the method considers the corresponding property. Interpretability is reflected by model size. For the Vectorization column, ``N/A'' denotes that the method uses scalar fitness and does not need to be vectorized.}
            \label{tab:related_work_properties}
            \centering
            \renewcommand{\arraystretch}{1.05}
            \setlength{\tabcolsep}{2.5pt}
            \resizebox{\linewidth}{!}{
                \begin{tabular}{lccccc}
                    \toprule
                    \textbf{Method}                                                    & \textbf{Diversity} & \textbf{Interpretability} & \textbf{Dynamic Pressure} & \textbf{Complementarity} & \textbf{Vectorization} \\
                    \midrule
                    \multicolumn{6}{l}{\textit{Lexicase-based Selection Operators}} \\
                    AutoLex~\cite{la2019probabilistic}                                 & $\checkmark$       &                           & $\checkmark$              &                          &                        \\
                    PLex~\cite{ding2023probabilistic}                                  & $\checkmark$       &                           & $\checkmark$              &                          & $\checkmark$           \\
                    DALex~\cite{ni2024dalex}                                           & $\checkmark$       &                           &                           &                          & $\checkmark$           \\
                    D-Split~\cite{imai2024minimum}                                     & $\checkmark$       &                           & $\checkmark$              &                          &                        \\
                    Down-sampled Lexicase~\cite{hernandez2019random,boldi2024informed} & $\checkmark$       &                           &                           &                          &                        \\
                    Lexi$^2$~\cite{de2022lexi2}                                        & $\checkmark$       & $\checkmark$              &                           &                          &                        \\
                    DLS~\cite{zhang2023double}                                         & $\checkmark$       & $\checkmark$              & $\checkmark$              &                          &                        \\
                    \midrule
                    \multicolumn{6}{l}{\textit{Traditional Selection Operators}} \\
                    Tournament~\cite{xie2012parent}                                    &                    &                           &                           &                          & N/A                    \\
                    Batch Tournament~\cite{de2019batch}                                & $\checkmark$       &                           &                           &                          &                        \\
                    Novelty Search~\cite{lehman2010efficiently}                        & $\checkmark$       &                           &                           &                          &                        \\
                    CPS~\cite{dolin2002opposites}                                      &                    &                           &                           & $\checkmark$             & $\checkmark$           \\
                    Boltzmann~\cite{shojaee2025llmsr,romera2024mathematical}           &                    &                           & $\checkmark$              &                          & N/A                    \\
                    RDS-Tour~\cite{geigerperformance}                                  & $\checkmark$       &                           &                           &                          & $\checkmark$           \\
                    \midrule
                    \multicolumn{6}{l}{\textit{Selection for Bloat Control}} \\
                    Lexicographic Parsimony~\cite{luke2002lexicographic}               &                    & $\checkmark$              &                           &                          & N/A                    \\
                    Double Tournament~\cite{luke2006comparison}                        &                    & $\checkmark$              &                           &                          & N/A                    \\
                    Pareto Tournament~\cite{kotanchek2006pursuing}                     &                    & $\checkmark$              &                           &                          & N/A                    \\
                    Adaptive Depth-Limit~\cite{silva2009dynamic,silva2012operator}     &                    & $\checkmark$              & $\checkmark$              &                          & N/A                    \\
                    \bottomrule
                \end{tabular}
            }
        \end{table}

        \subsection{LLMs for SR}
        Early work on language models for SR focused on specialized models~\cite{kamienny2022end}, often assisted by Monte Carlo Tree Search~\cite{shojaee2024transformer} or evolutionary algorithms~\cite{zhang2025ragsr}. With the advent of general-purpose LLMs, their use in SR has attracted increasing attention~\cite{shojaee2025llmsr}, either through the FunSearch framework~\cite{romera2024mathematical} or via integration with evolutionary algorithm frameworks~\cite{grayeli2024symbolic}.
        In these studies, LLMs effectively replace traditional crossover and mutation operators to generate candidate solutions, but selection operators are still manually designed. This motivates the exploration of automated selection operator design using LLMs. Although there have been rapid advancements in LLM-based program evolution, the fine-grained behavior of candidate programs is often overlooked and reduced to aggregate scores, which can obscure meaningful information and hinder evolutionary progress.

        \subsection{Automated Evolutionary Algorithm Design}
        \label{sec: Automated Algorithm Design}
        Before the LLM era, GP had been widely used for automated evolutionary algorithm design, including the design of fitness functions~\cite{fong2024metasr}, parameter adaptation techniques~\cite{stanovov2022automatic}, update strategies~\cite{lones2021evolving}, and black-box optimizers~\cite{stanovov2024designing}. These efforts are closely related to program synthesis~\cite{sobania2022comprehensive}, where grammar-guided GP~\cite{forstenlechner2018extending} and PushGP~\cite{stanovov2024designing} have been employed to synthesize executable code from specifications.
        More recently, reinforcement learning has also been applied to designing knowledge transfer strategies~\cite{wang2025multi}. However, these approaches are typically limited to generating only small code segments~\cite{fong2024metasr} or opaque rules~\cite{wang2025multi}.
        The emergence of LLMs has made automated algorithm design an alternative and ultimately more feasible approach. For example, LLMs have been used to design a differential evolution algorithm competitive with state-of-the-art continuous optimization methods~\cite{van2024llamea}. Similarly, LLMs have been employed to improve the self-organizing migrating algorithm~\cite{pluhacek2024using}. This growing interest in LLMs motivates the development of efficient LLM-based approaches to automated algorithm design, where semantic awareness and code bloat are two key issues that remain underexplored and warrant further investigation.

        \section{Algorithm Framework}
        \label{sec: Algorithm Framework}
        This work introduces a meta-learning framework that leverages large language models (LLMs) within a meta-SR setting to automate the design of selection operators through in-context learning. The framework's foundation is defined by the solution structure and the overall evolutionary workflow as detailed in \Cref{sec: Solution Representation} and \Cref{sec: Meta-Evolution}, where LLMs generate new operators by learning from historical design and evaluation results provided in prompts. Building upon this foundation, \Cref{sec: Semantics-Aware Evolution}, \Cref{sec: Bloat Control}, and \Cref{sec: Operator Design Principle} introduce key improvements, including semantics-aware evolution, bloat control, and the incorporation of domain knowledge to enhance the in-context learning process and guide the LLM toward generating more effective and efficient operators.

        \subsection{Solution Representation}
        \label{sec: Solution Representation}
        In the meta-learning framework, each candidate solution is a piece of code representing a selection operator. The input to the selection operator is a list of individuals. Each individual contains a list representing squared errors, a list of predicted values, and a list of nodes, which can be used to compute the height and depth of the symbolic tree. These inputs provide the necessary information for the selection operator to satisfy the desired properties summarized in \Cref{tab:related_work_properties}: the squared error and predictions enable the selection operator to consider semantics for preserving diversity as well as the complementarity property, the nodes enable control of complexity, and the evolutionary status—expressed as the ratio between the current and total generations—supports the dynamic selection pressure property. The expected output of the selection operator is a list of promising symbolic trees. The structure of the selection operator is presented in \Cref{lst:template} of the supplementary material.

        \subsection{Meta-Evolution Workflow}
        \label{sec: Meta-Evolution}

        \begin{figure*}[!t]
            \centering
            \includegraphics[width=\textwidth, trim=5pt 10pt 5pt 10pt, clip]{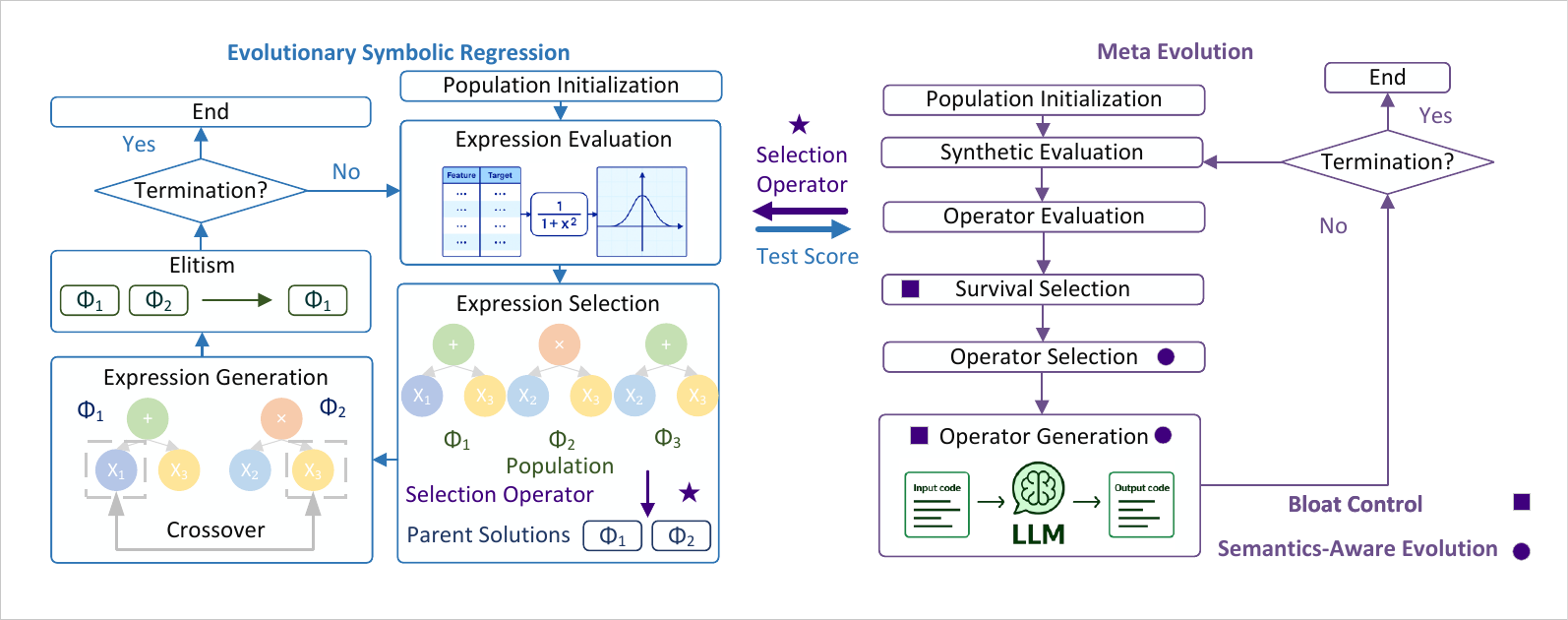}
            \caption{Workflow of LLM-driven selection operator evolution. The right-hand side shows the outer meta-evolution loop that generates candidate selection operators, while the left-hand side shows the inner SR loop that uses each candidate selection operator to evolve symbolic expressions and evaluates its performance.}
            \label{fig: workflow}
        \end{figure*}

        The proposed framework performs \emph{meta-evolution}: it optimizes selection operators rather than symbolic expressions.
        Unlike existing SR methods, which typically rely on manually designed selection operators, the proposed method uses LLM-driven meta-evolution to design them automatically.
        \par Specifically, the method consists of two nested loops. In the outer meta-evolution loop, the LLM generates candidate selection operators. For each candidate, the inner SR loop runs an evolutionary symbolic regression algorithm on multiple meta-training datasets to evaluate how effectively that selection operator guides the search.

        More concretely, for each candidate operator, the algorithm launches an SR run on every meta-training dataset. In each run, a population of symbolic expressions is initialized, and the candidate operator is invoked whenever parents must be chosen for variation. The selected expressions then undergo the standard SR loop, including variation and elitist survival, so that the population gradually improves over generations. These evolved expressions are intermediate solutions used only to assess the quality of the candidate selection operator, rather than the final object of optimization. Performance on each dataset's evaluation split is aggregated and returned as the fitness value of the candidate selection operator. The outer loop then compares candidate operators using these inner-loop outcomes and applies LLM-based crossover and mutation to generate improved selection operators for the next generation.

        Let \( \mathcal{P}^{(t)} = \{O_1^{(t)}, O_2^{(t)}, \dots, O_N^{(t)}\} \) denote the population of selection operators at generation \( t \), where \( N \) is the population size and each \( O_i^{(t)} \) is a selection operator. The fitness of each operator \( O_i^{(t)} \), denoted \( f(O_i^{(t)}) \), is evaluated based on its performance in the SR loop when that operator is used. The meta-evolution process includes the following components:
        \begin{itemize}
            \item \textbf{Population Initialization:} The initial population $\mathcal{P}_{*}^{(0)}$ is generated by prompting the LLM to produce $N$ random selection operators. The details of the initialization prompts are provided in \Cref{sec: Prompt} of the supplementary material.
            \item \textbf{Synthetic Evaluation:} To ensure the validity and efficiency of LLM-generated operators, each \( O_i^{(t)} \in \mathcal{P}_{*}^{(t)} \) is first evaluated in a synthetic testing environment before being applied to real SR tasks. Two synthetic test cases are designed to provide an inexpensive proxy for assessing an operator's validity. They require only minimal computation to detect syntax errors, runtime failures, and inefficient implementations. This screening step is important because the downstream SR-based evaluation of a candidate operator takes hundreds of seconds on average, as shown in \Cref{sec: evaluation_time} of the supplementary material, so evaluating invalid or pathological operators directly on the real tasks would waste substantial computation. The first case consists of 100 solutions whose predicted values and training errors are independently sampled from integers in the range $[1, 10]$, representing a diverse scenario with varying fitness values. The second case also includes 100 solutions but assigns identical fitness to all individuals by repeating a single integer value sampled from $[1, 10]$, representing an edge case of uniform performance. An operator is discarded if it contains syntax or runtime errors, exceeds a 300-second execution time, or is more than 100$\times$ slower than the fastest operator in the batch. Let \( \mathcal{P}^{(t)} \subseteq \mathcal{P}_{*}^{(t)} \) denote the filtered population after synthetic evaluation.
            \item \textbf{Solution Evaluation:} For each selection operator \( O_i^{(t)} \in \mathcal{P}^{(t)} \) that survived the synthetic evaluation, its fitness is computed as $f(O_i^{(t)}) = \frac{1}{T} \sum_{j=1}^{T} \texttt{SR}(O_i^{(t)}, \mathcal{D}_j),$ where $\mathcal{D}_1, \dots, \mathcal{D}_T$ are meta-learning SR datasets, and $\texttt{SR}(O_i^{(t)}, \mathcal{D}_j)$ denotes the SR performance of $O_i^{(t)}$ on dataset $\mathcal{D}_j$. Because these are meta-learning datasets used for meta-evolution, each dataset is split into training and evaluation sets only, with no separate test set. SR is performed on the training set, and the fitness of the selection operator is computed based on its performance on the evaluation set. Details of the solution evaluation procedure are provided in \Cref{sec: ESR} of the supplementary material.
            \item \textbf{Survival Selection:} After solution generation, the current population $\mathcal{P}^{(t)}$ is merged with the parent population from the previous generation $\mathcal{P}_{\cup}^{(t-1)}$, and survival selection is applied to form $\mathcal{P}_{\cup}^{(t)} = \sigma(\mathcal{P}^{(t)} \cup \mathcal{P}_{\cup}^{(t-1)})$, where $\sigma$ denotes the survival selection operator that maintains the population size by reducing it from $2|\mathcal{P}|$ to $|\mathcal{P}|$. Operators from $\mathcal{P}_{\cup}^{(t)}$ then serve as parents for generating the next generation $\mathcal{P}_{*}^{(t+1)}$. To control bloat, a multi-objective survival selection strategy is employed, as described in \Cref{sec: Bloat Control}.
            \item \textbf{Solution Selection:} Two parent operators $O_a^{(t)}, O_b^{(t)} \in \mathcal{P}^{(t)}$ are selected for recombination. The selection strategy incorporates semantic information, as detailed in \Cref{sec: Semantics-Aware Evolution}.

            \item \textbf{Solution Generation:} Based on the selected parents, $N$ candidate operators are generated per generation. Of these, $N - M$ are generated via crossover and $M$ via mutation, where this paper sets $M = 1$:
            \begin{itemize}
                \item \textbf{Operator Crossover:} For each pair \( (O_a^{(t)}, O_b^{(t)}) \), a new operator is generated as $O_{\text{new}}^{(t+1)} = \Crossover\big(O_a^{(t)}, O_b^{(t)}\big)$. The LLM prompt provides the parent code and their dataset-specific performance scores as context for generating improved operators that combine strengths, with details shown in \Cref{fig: llm_crossover} of the supplementary material.
                \item \textbf{Operator Mutation:} Let $O^* = \arg\max_{O_i^{(t)} \in \mathcal{P}^{(t)}} f(O_i^{(t)})$ be the best-performing operator in generation $t$~\cite{ye2024reevo}. Mutated variants are then generated as \(O_{\text{mut}}^{(t+1)} = \Mutate(O^*)\), where the LLM prompt provides the elite operator's code as context for generating variants, with details shown in \Cref{fig: llm_mutation} of the supplementary material.
            \end{itemize}
            Both crossover and mutation leverage in-context learning~\cite{gao2024customizing} by analyzing design and evaluation results in prompts, enabling LLMs to generate better solutions by identifying patterns in historical execution results.
        \end{itemize}
        This process continues until a predefined number of generations $T_{\text{max}}$ is reached. The operator with the highest fitness value across all generations is selected as the final selection operator.

        \subsection{Semantics-Aware Evolution}
        \label{sec: Semantics-Aware Evolution}
        \textbf{Semantics-based Selection:}
        In the meta-evolution scenario, the goal of the crossover operator is to combine the strengths of two LLM-generated selection operators. As introduced in \Cref{sec: Introduction}, crossover benefits more from combining solutions with complementary strengths than from pairing generalists.

        To this end, we propose a semantics-aware selection strategy, with the pseudocode provided in \Cref{alg:sel-complement}. Because each selection operator $O_i \in \mathcal{P}^{(t)}$ has been evaluated across $T$ datasets and is associated with a score vector $\mathbf{s}_i = [s_{i,1}, \dots, s_{i,T}]$, we can compute a complementarity score for each candidate in the population. The process begins by randomly selecting the first parent \( O_a \in \mathcal{P}^{(t)} \). This random selection strategy, as used in ReEvo~\cite{ye2024reevo} and HSEvo~\cite{dat2025hsevo}, helps mitigate premature convergence~\cite{dat2025hsevo}. Then, for each of the other individuals \( O_i \in \mathcal{P}^{(t)} \), we compute a complementarity score:
        \begin{equation}
            \mu_i = \frac{1}{T} \sum_{j=1}^T \max(s_{a,j}, s_{i,j}),
        \end{equation}
        where the expression averages the element-wise maxima of the two score vectors, so the resulting value reflects the potential combined performance of \( O_a \) and \( O_i \) and is high only when $O_i$ improves on $O_a$ for at least some datasets.
        The second parent \( O_b \) is retrieved as the operator with the highest complementarity score, i.e., \( O_b = O_{i^*} \) where \( i^* = \arg\max_i \mu_i \). In other words, complementarity serves as the retrieval criterion to identify the most complementary code from the population for guiding code generation.

        \begin{algorithm}[!t]
        \caption{Semantics-Aware Selection}
        \label{alg:sel-complement}
        \begin{algorithmic}[1]
            \REQUIRE \( \mathcal{P}^{(t)} = \{O_1, \dots, O_N\} \): population where each \( O_i \) has score vector \( \mathbf{s}_i \)
            \STATE \( O_a \gets \) Random sample from \( \mathcal{P}^{(t)} \)
            \FOR{each \( O_i \in \mathcal{P}^{(t)} \)}
            \STATE Compute \( \mu_i \gets \frac{1}{T} \sum_{j=1}^T \max(s_{a,j}, s_{i,j}) \)
            \ENDFOR
            \STATE \( i^* \gets \arg\max_i \mu_i \)
            \STATE \( O_b \gets O_{i^*} \) \hfill // Retrieve complementary operator
            \RETURN \( O_b \)
        \end{algorithmic}
        \end{algorithm}

        \textbf{Semantic Feedback:}
        The semantic selection strategy ensures that crossover is guided by semantically diverse behaviors. Furthermore, to enhance semantic awareness in the in-context learning process, we provide the full score vector $\mathbf{s}_i$ of dataset-specific scores directly in the prompts used for solution generation, rather than averaging them into a single aggregate value. By including this fine-grained semantic information as context, the LLM can reason explicitly about behavioral differences across tasks and generate operators that better integrate complementary capabilities.

        \subsection{Bloat Control}
        \label{sec: Bloat Control}

        \textbf{Prompt-based Length Limit:}
        To mitigate code bloat in the evolution of selection operators, we incorporate a length constraint in the prompt during solution generation. For any operator \( O_i \in \mathcal{P}^{(t)} \), its code length is denoted by \( \ell(O_i) \), measured as the number of non-empty, non-comment lines in the implementation. We choose to constrain the number of lines, rather than the number of tokens, because line count is less sensitive to variations in variable name length. Unlike traditional GP, which enforces a hard upper bound \( \ell(O_i) \leq L_{\max} \) by sampling and discarding solutions that exceed the limit, LLM-based generation enables us to state the constraint directly in the prompt. For instance, the LLM is instructed: ``Write a selection operator with code length \( \leq \ell_{\text{target}} \).'' This prompt-guided strategy encourages the model to produce more concise programs.

        \textbf{Multi-Objective Survival Selection:}
        Beyond prompt-based constraints, we employ a multi-objective survival selection strategy based on both operator fitness and code length to further control bloat, since LLMs do not always follow instructions to generate code within a specified length. Each operator \( O_i \) is represented as a tuple \( (f(O_i), \ell(O_i)) \), where \( f(O_i) \) denotes its average task performance across \( T \) tasks, and \( \ell(O_i) \) denotes its code length. The merged set $\mathcal{P}_{\cup}^{(t)} = \sigma(\mathcal{P}^{(t)} \cup \mathcal{P}_{\cup}^{(t-1)})$ is formed using a dominance-dissimilarity selection mechanism~\cite{yao2025multi}, and operators from $\mathcal{P}_{\cup}^{(t)}$ serve as parents for generating the next generation $\mathcal{P}_{*}^{(t+1)}$.

        The key idea of the selection mechanism is to compute a dominance score for each operator based on weak Pareto dominance~\cite{yuan2017objective} and code similarity. Since fitness \( f \) is maximized and code length \( \ell \) is minimized, an operator \( O_i \) is said to weakly Pareto dominate another operator \( O_j \), denoted \( O_i \succeq O_j \), if and only if \( f(O_i) \geq f(O_j) \) and \( \ell(O_i) \leq \ell(O_j) \). For each such pair \( (O_i, O_j) \), the score of \( O_j \) is penalized by the code similarity \( \text{sim}(O_i, O_j) \), computed using the CodeBLEU metric~\cite{ren2020codebleu}. CodeBLEU evaluates generated code by combining token overlap, syntax structure, and semantic data flow so that the score better matches human judgment. The CodeBLEU score between candidate code $C$ and reference code $R$ is defined as:
        \begin{equation}
            \begin{split}
                \text{CodeBLEU}(C, R) &= \alpha\, \text{BLEU}(C, R) + \beta\, \text{BLEU}_{\text{weight}}(C, R) \\
                &\quad + \gamma\, \text{Match}_{\text{ast}}(C, R) + \delta\, \text{Match}_{\text{df}}(C, R),
            \end{split}
        \end{equation}
        where $\text{BLEU}$ and $\text{BLEU}_{\text{weight}}$ measure token overlap, $\text{Match}_{\text{ast}}$ captures syntax structure similarity via abstract syntax tree matching, and $\text{Match}_{\text{df}}$ evaluates semantic similarity through data-flow matching. Intuitively, this penalty discourages retaining operators that are simultaneously worse (dominated) and redundant (highly similar) compared to others. The total dominance score of \( O_j \) is then defined as \(s(O_j) = \sum_{O_i \succeq O_j} -\text{sim}(O_i, O_j)\). Operators are then ranked, and the top-$N$ operators with higher scores are selected, as they are less frequently dominated and exhibit greater dissimilarity from dominating counterparts. Unlike Pareto tournament selection~\cite{kotanchek2006pursuing}, which selects enough individuals over multiple tournaments, this mechanism selects enough individuals in one round. This selection strategy encourages a trade-off between maximizing task performance and minimizing code complexity, effectively reducing operator bloat while maintaining diversity and effectiveness.

        \subsection{Incorporating Domain Knowledge into Prompts}
        \label{sec: Operator Design Principle}
        Algorithm evolution requires domain expertise, which general-purpose LLMs often lack. To compensate, domain knowledge\footnote{Here, domain knowledge refers to common knowledge shared by machine learning and GP experts, rather than knowledge specific to a particular application domain.} is commonly incorporated into prompts to enhance LLM-based algorithm evolution~\cite{romera2024mathematical, liu2024evolution}. For the design of selection operators, we embed the following principles directly into the prompt text so that the LLM explicitly sees the desired properties when generating solutions:

        \textbf{Diversity-aware:} Diversity means that the selection operator should drive the SR algorithm to discover diverse solutions, rather than referring to the diversity of the selection operator code itself. Purely objective-driven selection can cause premature convergence. To counter this, it is desirable to select models that perform well on different training instances to maintain population diversity.

        \textbf{Interpretability-aware:} Interpretability means that the selection operator should guide the SR algorithm to discover interpretable solutions, rather than emphasizing the interpretability of the selection operator itself. Interpretability, often measured by the number of nodes in a symbolic expression, is a critical criterion. Let $n_i$ denote the number of nodes in solution $p_i$. The selection operator should favor solutions with smaller $n_i$ to promote simpler and more interpretable models.

        \textbf{Dynamic Selection Pressure:} Selection pressure should adapt over time. In early generations ($t \ll T_{\text{max}}$), it should favor exploration. In later generations ($t \approx T_{\text{max}}$), it should prioritize exploitation of high-fitness solutions to encourage convergence.

        \textbf{Complementarity-aware:} To effectively recombine useful traits, crossover should favor solutions with complementary strengths. Given performance vectors $\mathbf{s}_i$ and $\mathbf{s}_j$ over $T$ datasets, complementarity means selecting pairs with low correlation. Such combinations integrate distinct capabilities and can yield offspring that perform well across a broader range of datasets.

        \textbf{Vectorization-aware:} Vectorized operations are preferred, as they can be efficiently accelerated using NumPy or GPU computation. For example, lexicase processes individuals and cases sequentially, while DALex~\cite{ding2023probabilistic} formulates weighting summations so that selection reduces to matrix operations. This improves the runtime efficiency of the evolved operator and reduces evaluation overhead during meta-evolution.

        These domain knowledge principles describe the desired properties that selection operators designed by the LLM should satisfy. The prompt that incorporates this domain knowledge is provided in \Cref{fig: domain knowledge} of the supplementary material.

        \section{Experimental Settings}
        \label{sec: Experimental Settings}

        \subsection{Datasets}
        The experiments are conducted using datasets from the contemporary SR benchmark (SRBench)~\cite{la2021contemporary}. Following a meta-learning setup, four datasets are selected as meta-training tasks for evolving the selection operators, corresponding to the four highest-dimensional datasets in SRBench with OpenML IDs 505, 4544, 588, and 650. High-dimensional datasets are chosen because they pose greater challenges for SR algorithms, and operators optimized under these conditions are more likely to generalize effectively to both easy and hard problems. For meta-testing, experiments are conducted on 116 out of the 120 datasets in SRBench, excluding the four used during meta-training to prevent potential data leakage. Each algorithm is evaluated using ten independent runs per dataset to ensure statistically reliable comparisons.

        \subsection{Evaluation Protocol}
        \label{sec: Evaluation Protocol}
        \textbf{Meta-Evolution:} In the evaluation of selection operators generated by the LLM, all other algorithmic components—including solution initialization, evaluation, generation, and elitism—follow standard practices in evolutionary SR. For each dataset \( \mathcal{D}_j \), the data is randomly split into training and evaluation subsets using an 80:20 split~\cite{la2021contemporary}. The regression model is optimized to maximize the training \( R^2 \) score, while the evaluation \( R^2 \) score is used as the fitness measure of the selection operator. This design encourages the discovery of operators that exhibit implicit regularization and generalization capability. 
        Each algorithm variant is run three times independently, following a widely adopted convention in LLM-based algorithm evolution where each run incurs substantial computational cost, mainly due to iterative querying of the LLM as well as inner-loop evaluations. This practice is consistent with prior work such as ReEvo~\cite{ye2024reevo}, EoH~\cite{liu2024evolution}, multi-objective EoH~\cite{yao2025multi}, LLM-LNS~\cite{ye2025large}, and MCTS-AHD~\cite{zheng2025monte}.

        \textbf{Discovered Operators:} For the final evaluation of the evolved selection operator on SRBench, each dataset is split into training and test sets using an 80:20 split, following the SRBench protocol~\cite{la2021contemporary}. For datasets with more than 10,000 training instances, subsampling is applied to limit the maximum number of training instances to 10,000~\cite{la2021contemporary}. SR is performed on the training set and evaluated on the test set. All input features are standardized, as feature standardization has been shown to improve SR performance~\cite{owen2022standardization}. The standardization parameters are computed from the training data and applied consistently to both the training and test sets.
        The evaluation uses 30 independent runs per dataset, and the pairwise statistical comparisons among discovered operators and baselines in \Cref{sec: Discovered Operators} are based on these runs and are conducted using the Wilcoxon signed-rank test with \( p = 0.05 \).

        \subsection{Baseline Algorithms}
        We compare the proposed method against several representative selection algorithms, and a summary of their key characteristics is provided in Table~\ref{tab:baseline_properties}. Methods such as tournament selection do not require vectorization since their operations rely solely on scalar fitness comparisons. In contrast, semantic or case-based selection operators benefit significantly from vectorized computation, as modern hardware architectures, e.g., SIMD CPUs and GPUs, can efficiently accelerate batch operations. In our implementation, for algorithms that do not natively support vectorization, we employ Numba to JIT-compile critical loops, ensuring computational fairness and avoiding performance disadvantages during comparison.

        \begin{table}[t]
            \caption{Comparison of baseline selection operators with respect to the proposed design principles. A checkmark ($\checkmark$) indicates that the operator satisfies the corresponding property, while ``N/A'' denotes not applicable.}
            \label{tab:baseline_properties}
            \centering
            \renewcommand{\arraystretch}{1.05}
            \setlength{\tabcolsep}{3pt}
            \resizebox{\linewidth}{!}{
                \begin{tabular}{lccccc}
                    \toprule
                    \textbf{Operator}                  & \textbf{Diversity} & \textbf{Interpretability} & \textbf{Dynamic Pressure} & \textbf{Complementarity} & \textbf{Vectorization} \\
                    \midrule
                    AutoLex~\cite{la2019probabilistic} & $\checkmark$       &                           & $\checkmark$              &                          &                        \\
                    PLex~\cite{ding2023probabilistic}  & $\checkmark$       &                           & $\checkmark$              &                          & $\checkmark$           \\
                    DALex~\cite{ni2024dalex}           & $\checkmark$       &                           &                           &                          & $\checkmark$           \\
                    D-Split~\cite{imai2024minimum}     & $\checkmark$       &                           & $\checkmark$              &                          &                        \\
                    RDS-Tour~\cite{geigerperformance}  & $\checkmark$       &                           &                           &                          & $\checkmark$           \\
                    CPS~\cite{dolin2002opposites}      &                    &                           &                           & $\checkmark$             & $\checkmark$           \\
                    Tour-3/7~\cite{banzhaf1998genetic} &                    &                           &                           &                          & N/A                    \\
                    Boltzmann~\cite{shojaee2025llmsr}  &                    &                           & $\checkmark$              &                          & N/A                    \\
                    DLS~\cite{zhang2023double}         & $\checkmark$       & $\checkmark$              & $\checkmark$              &                          &                        \\
                    \bottomrule
                \end{tabular}
            }
        \end{table}

        \begin{itemize}
            \item \textbf{Automatic $\varepsilon$-Lexicase Selection (AutoLex)}~\cite{la2019probabilistic}: Employs a median absolute deviation (MAD)-based adaptive $\varepsilon$ threshold to automatically regulate case-wise filtering strength. AutoLex maintains the diversity–performance balance of lexicase selection while simplifying parameter control through statistical adaptation.
            \item \textbf{Probabilistic Lexicase Selection (PLex)}~\cite{ding2023probabilistic}: Extends lexicase selection by first identifying non-dominated individuals across test cases using a MAD-based adaptive $\varepsilon$ threshold, and then applying softmax-based probabilistic sampling to these candidates. The temperature parameter controls selection pressure, offering a tunable trade-off between exploration and exploitation.
            \item \textbf{Lexicase-Like Selection via Diverse Aggregation (DALex)}~\cite{ni2024dalex}: Aggregates normalized case-wise errors using Gaussian-sampled importance weights, selecting individuals that minimize these weighted errors. This method promotes diversity through stochastic aggregation rather than case ordering, providing a simpler alternative to lexicase filtering.
            \item \textbf{$\varepsilon$-Lexicase Selection with Dynamic Split (D-Split)}~\cite{imai2024minimum}: Determines an adaptive threshold $\tau^*$ for each case by minimizing intra-group error variance in lexicase selection.
            \item \textbf{Random Down-Sampling Tournament Selection (RDS-Tour)}~\cite{geigerperformance}: Conducts tournaments using a randomly sampled subset of test cases to compute temporary fitness values. By evaluating only a fraction of cases, RDS-Tour introduces stochasticity to maintain diversity.
            \item \textbf{Complementary Phenotype Selection (CPS)}~\cite{dolin2002opposites}: Selects the first parent (mother) using a standard tournament and then identifies a second parent (father) that minimizes element-wise offspring error via $\min(F, M)$. CPS encourages complementary error reduction between parents, promoting synergistic offspring generation.
            \item \textbf{Tournament Selection (Tour-3/7)}~\cite{banzhaf1998genetic}: The most commonly used selection operator in evolutionary computation, where individuals compete in tournaments of size three or seven, and the best performer is selected.
            \item \textbf{Boltzmann Selection with Temperature Scheduling (Boltzmann)}~\cite{shojaee2025llmsr}: Applies a temperature-controlled softmax over fitness values, gradually shifting from stochastic to selective behavior as evolution progresses. This thermodynamic-inspired mechanism adaptively adjusts selection pressure across generations.
            \item \textbf{Double Lexicase Selection (DLS)}~\cite{zhang2023double}: Repeats the lexicase selection process ten times to enhance robustness and control program bloat. Among the individuals selected across these runs, the less complex one is probabilistically favored for reproduction. This two-phase mechanism preserves the diversity advantages of lexicase selection while introducing implicit parsimony pressure that promotes more compact yet competitive solutions.
        \end{itemize}

        \subsection{Parameter Settings}
        \textbf{Meta-Evolution:} The meta-evolution process is implemented using GPT-4.1 Mini. In the outer loop, the population size \(N\) is set to 20, with one offspring generated by mutation and 19 offspring generated by crossover per generation. The total number of generations is 20, and each candidate operator is limited to a maximum of 30 lines to control code complexity. In the inner loop, the population size is 100 and the number of generations is 30, defining the inner SR loop used to evaluate each generated operator. All experiments are repeated three times with random seeds 0, 1, and 2, and the median performance with confidence intervals is reported following~\cite{yao2025multi}. The detailed meta-evolution parameters are summarized in Table~\ref{tab:meta_params}.

        \textbf{Symbolic Regression:} For SR evaluation, standard GP with linear scaling is adopted, consistent with widely used frameworks in SR research~\cite{virgolin2021improving,kronberger2022shape}. The parameter settings are identical to those used in the inner loop, except that the number of generations is increased to 100 to allow more extensive evolution. More implementation details are provided in \Cref{sec: ESR} of the supplementary material.

        \begin{table}[!t]
            \caption{Parameter settings for the meta-evolution framework.}
            \label{tab:meta_params}
            \centering
            \begin{tabular}{lc}
                \toprule
                \textbf{Parameter}                      & \textbf{Value}          \\
                \midrule
                LLM model                               & GPT-4.1 Mini/GPT-5 Mini \\
                Outer-loop population size ($N$)        & 20                      \\
                Mutation offspring per generation ($M$) & 1                       \\
                Crossover offspring per generation      & 19                      \\
                Outer-loop generations                  & 20                      \\
                Maximum operator length                 & 30 lines                \\
                Inner-loop population size              & 100                     \\
                Inner-loop generations                  & 30                      \\
                \bottomrule
            \end{tabular}
        \end{table}

        \textbf{Baseline Selection Operators:} The hyperparameters for the baseline selection operators follow the default settings specified in their respective original papers. For RDS-Tour, the sampling ratio is set to 10\% and the tournament size to 7~\cite{geiger2025tournament}. For PLex selection, the temperature parameter is set to 1.0~\cite{ding2023probabilistic}. For DALex, a particularity pressure of 3 is used~\cite{ni2024dalex}. For CPS, the first parent is selected via tournament, with the tournament size set to 7~\cite{xu2022genetic}. For Boltzmann sampling with temperature scheduling~\cite{shojaee2025llmsr,romera2024mathematical}, a temperature decay rule is applied to balance exploration and exploitation over time, similar to the approaches used in LLM-SR~\cite{shojaee2025llmsr} and FunSearch~\cite{romera2024mathematical}. The sampling probability for individual $i$ is defined as:
        \begin{equation}
            P_i = \frac{\exp\left(\frac{s_i}{\tau_c}\right)}{\sum_{i'} \exp\left(\frac{s_{i'}}{\tau_c}\right)}, \quad
            \tau_c = \tau_0 \left(1 - \frac{t \bmod T_{\max}}{T_{\max}} \right),
        \end{equation}
        where $s_i$ denotes the score of individual $i$, $\tau_0=0.1$ is the initial temperature~\cite{shojaee2025llmsr}, $t$ is the current generation, and $T_{\max}$ is the total number of generations. This scheduling rule linearly decays the temperature over generations and gradually shifts the selection pressure from exploration (high temperature) to exploitation (low temperature) as evolution progresses.

        \section{Experimental Results}
        \label{sec: Experimental Results}
        In this section, we evaluate the proposed LLM-Meta-SR framework and its evolved operators through three complementary analyses. \Cref{sec: Meta-Evolution Results} presents the meta-evolution and ablation results, highlighting the effectiveness of semantic evolution, bloat control, and domain knowledge in guiding LLM-driven search. \Cref{sec: Discovered Operators} then focuses on the selection operator with the highest evaluation score in \Cref{sec: Meta-Evolution Results}, denoted Omni, and examines its performance against expert-designed baselines to demonstrate its superiority in accuracy, efficiency, and interpretability. Finally, \Cref{sec: Analysis on SOTA SR Algorithms} integrates the evolved operator into a state-of-the-art SR framework, confirming its general applicability and practical value in modern algorithmic settings.

        \subsection{Meta-Evolution Results}
        \label{sec: Meta-Evolution Results}
        \begin{table*}[!tb]
            \centering
            \caption{Median historical best scores and code lengths over three runs when using \textit{GPT-4.1 Mini} as the LLM backend.}
            \label{tab:historical-best-gpt4}
            \begin{tabular}{lcccccc}
                \toprule
                & LLM-Meta-SR     & W/O Semantics   & W/O SE+BC       & W/O Knowledge   & W/O DK+SE       & W/O DK+SE+BC \\
                \midrule
                Score         & 0.86 $\pm$ 0.03 & 0.84 $\pm$ 0.02 & 0.84 $\pm$ 0.03 & 0.79 $\pm$ 0.01 & 0.75 $\pm$ 0.03 & 0.75         \\
                Lines of Code & 48 $\pm$ 4      & 43 $\pm$ 16     & 80 $\pm$ 9      & 55 $\pm$ 5      & 41 $\pm$ 21     & 222          \\
                \bottomrule
            \end{tabular}
        \end{table*}
        \begin{table*}[!tb]
            \centering
            \caption{Median historical best scores and code lengths over three runs when using \textit{GPT-5 Mini} as the LLM backend.}
            \label{tab:historical-best-gpt5}
            \begin{tabular}{lcccccc}
                \toprule
                & LLM-Meta-SR     & W/O Semantics   & W/O SE+BC       & W/O Knowledge   & W/O DK+SE       & W/O DK+SE+BC    \\
                \midrule
                Score         & 0.86 $\pm$ 0.01 & 0.85 $\pm$ 0.01 & 0.85 $\pm$ 0.01 & 0.83 $\pm$ 0.02 & 0.81 $\pm$ 0.02 & 0.80 $\pm$ 0.01 \\
                Lines of Code & 47 $\pm$ 8      & 50 $\pm$ 6      & 124 $\pm$ 64    & 43 $\pm$ 12     & 36 $\pm$ 12     & 211 $\pm$ 122   \\
                \bottomrule
            \end{tabular}
        \end{table*}

        In this section, we evaluate the proposed meta-evolution framework, LLM-Meta-SR, against several ablated variants, including variants without semantic evolution (W/O Semantics), without semantic evolution and bloat control (W/O SE+BC), without domain knowledge (W/O Knowledge), without domain knowledge and semantic evolution (W/O DK+SE), and without domain knowledge, semantic evolution, and bloat control (W/O DK+SE+BC). Because domain knowledge might not be familiar to non-GP experts, we also include ablation studies under settings without domain knowledge to examine the framework’s generalization ability in unseen domains. In the W/O Semantics setting, random selection with identical-objective prevention from ReEvo~\cite{ye2024reevo} is used as the baseline, meaning the second parent is repeatedly resampled until its objective differs from that of the first parent. In the W/O SE+BC setting, survival selection is replaced with direct population replacement and elitism following the strategy used in ReEvo~\cite{ye2024reevo}. In the W/O Knowledge setting, semantic evolution and bloat control are retained, but the domain knowledge prompt is removed.

\begin{figure*}[!t]
            \centering
            \begin{subfigure}[t]{0.32\textwidth}
                \centering
                \includegraphics[width=\textwidth]{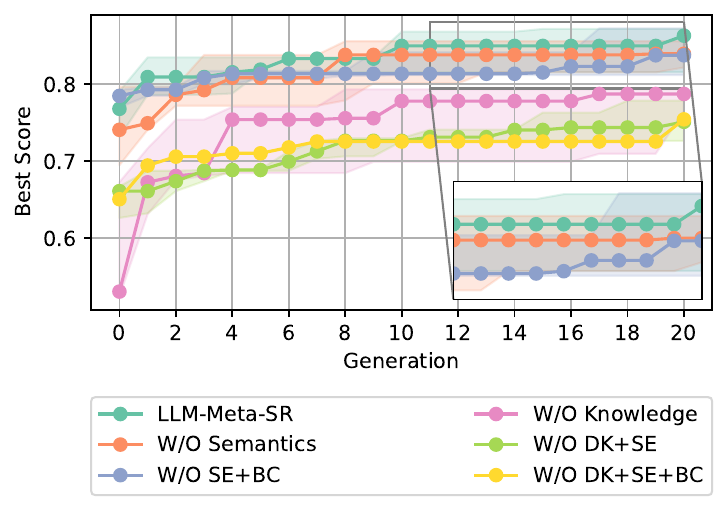}
                \caption{Evaluation $R^2$ of the best solution.}
                \label{fig:llm_evolution}
            \end{subfigure}
            \hfill
            \begin{subfigure}[t]{0.32\textwidth}
                \centering
                \includegraphics[width=\textwidth]{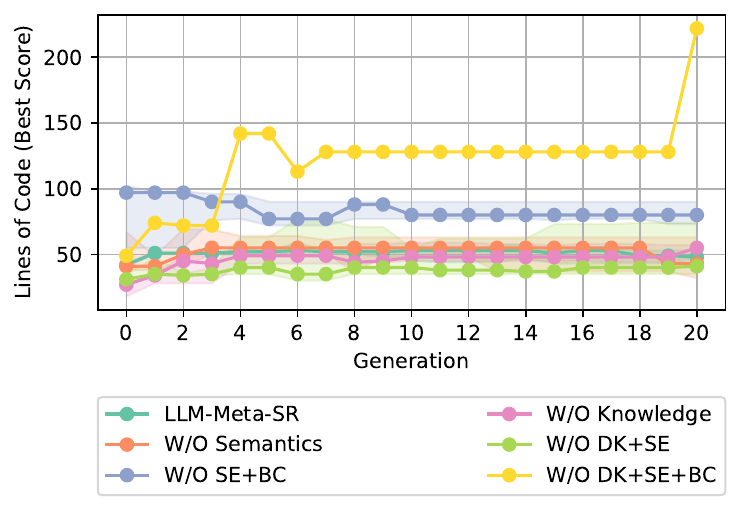}
                \caption{Code length of the best solution.}
                \label{fig:llm_code_length_evolution}
            \end{subfigure}
            \hfill
            \begin{subfigure}[t]{0.32\textwidth}
                \centering
                \includegraphics[width=\textwidth]{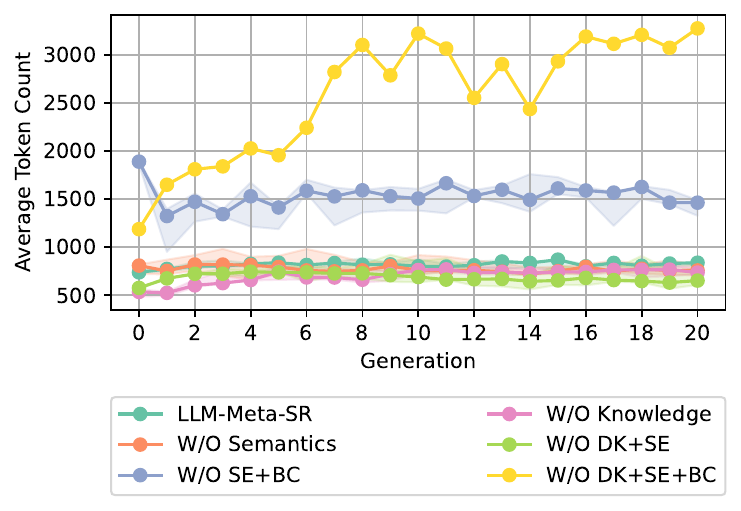}
                \caption{Average token count of generated code.}
                \label{fig:token_count}
            \end{subfigure}
            \caption{Comparison across generations for different LLM-driven search strategies using \textit{GPT-4.1-Mini}.}
            \label{fig:llm_evolution_comparison}
        \end{figure*}

        \textbf{Objective Score:} The objective score is measured by the average test $R^2$ value of each selection operator across the four training datasets. For the W/O DK+SE+BC setting with \textit{GPT-4.1 Mini}, only one run completed within 24 hours and is therefore reported; all other settings report the median over three runs. The results in \Cref{fig:llm_evolution} and \Cref{tab:historical-best-gpt4,tab:historical-best-gpt5} show that removing domain knowledge leads to the largest performance drop, which highlights the importance of domain knowledge introduced in \Cref{sec: Operator Design Principle} for guiding algorithm evolution. This finding is consistent with observations from FunSearch~\cite{romera2024mathematical} and EoH~\cite{liu2024evolution}, where LLMs also relied on external knowledge for effective search. However, domain knowledge alone does not guarantee high performance. As summarized in \Cref{tab:historical-best-gpt4,tab:historical-best-gpt5}, semantic evolution still provides consistent gains relative to its ablated counterparts.

\begin{figure*}[!t]
            \centering
            \includegraphics[width=0.65\textwidth, trim=0pt 5pt 0pt 0pt, clip]{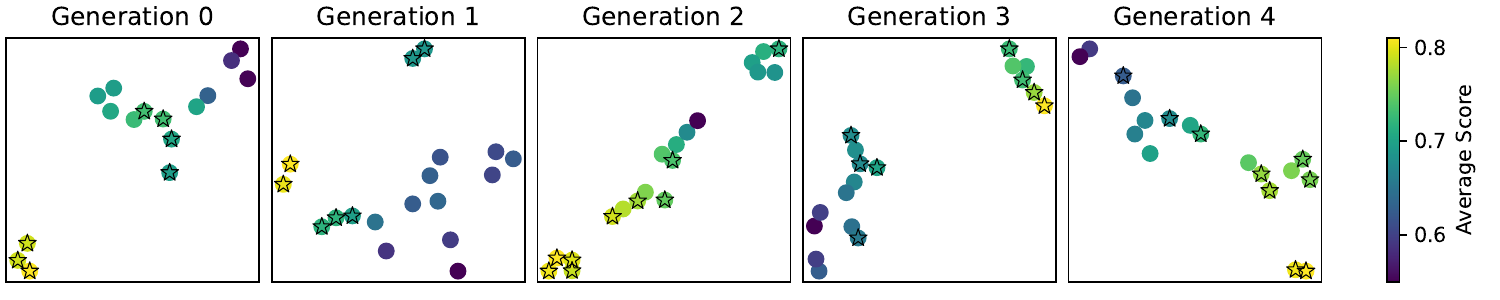}
            \caption{t-SNE visualization of evolved operator semantics. The shape of each point indicates whether it achieved top-3 performance on any task: stars denote top-3 performance on at least one task, while circles indicate otherwise.}
            \label{fig: tsne}
        \end{figure*}

        To further analyze the effect of semantic evolution, \Cref{fig: tsne} visualizes the semantic distribution of solutions during the first five generations. Each point represents a solution and is colored by its average score across the four tasks. Each point is positioned using t-SNE to project its four-dimensional performance vector. Although the ultimate goal of meta-evolution is to maximize the aggregated score across tasks, the visualization reveals that relying solely on aggregate scores during evolution can be misleading. In particular, selection operators that rank in the top three on at least one task, as indicated by star markers in \Cref{fig: tsne}, are not necessarily globally top-ranked individuals. This suggests that task-specific specialists with complementary strengths can be overlooked when parent selection is based only on aggregate rankings. This observation directly motivates our semantics-aware approach, which leverages fine-grained per-task performance vectors to identify and combine complementary operators during evolution, ultimately leading to better aggregated performance. Additional qualitative analysis of semantic-aware crossover is provided in \Cref{sec: Semantic-Aware Crossover}, which further demonstrates that semantic-aware evolution encourages diverse and high-quality offspring.

        \textbf{Code Length:} \Cref{fig:llm_code_length_evolution} presents the evolution of code length for the best-performing solutions. When bloat control is removed, the code length grows excessively throughout optimization, especially without domain knowledge. Introducing domain knowledge helps constrain the structure of the generated operators, but the length remains large. By contrast, adding bloat control effectively mitigates the LLM’s tendency to generate unnecessarily long code, producing concise operators of approximately 50 lines that also achieve faster improvements in objective score.

        \textbf{Token Count:} The token count directly affects the computational cost of algorithm evolution, as language models are billed per token. Token counting is performed using the \texttt{cl100k\_base} tokenizer, which is consistent with the OpenAI implementation. As shown in \Cref{fig:token_count}, removing bloat control leads to significantly higher token counts, whereas incorporating bloat control consistently reduces token generation in both the LLM-Meta-SR and W/O Knowledge settings. These results demonstrate that bloat control not only improves the interpretability of evolved code but also reduces the token cost of LLM-driven evolution.

        \begin{figure*}[!t]
            \centering
            \begin{subfigure}[t]{0.32\textwidth}
                \centering
                \includegraphics[width=\textwidth, trim=5pt 5pt 5pt 5pt, clip]{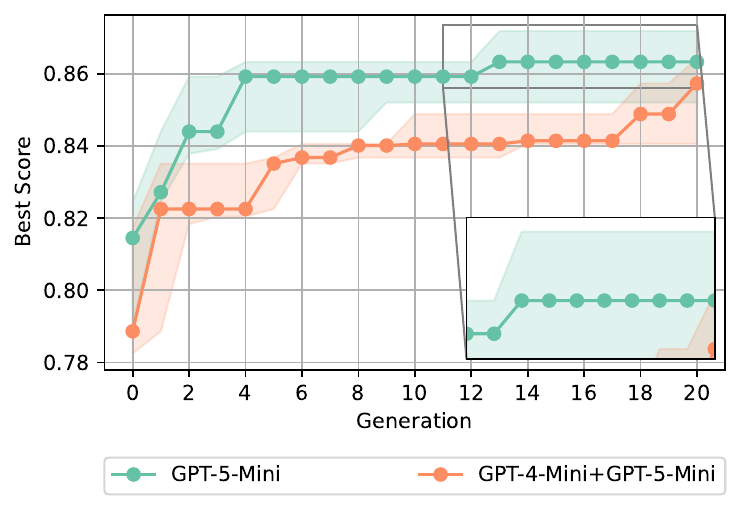}
                \caption{Model comparison.}
                \label{fig:gpt5_comparison}
            \end{subfigure}
            \hfill
            \begin{subfigure}[t]{0.32\textwidth}
                \centering
                \includegraphics[width=\textwidth, trim=5pt 5pt 5pt 5pt, clip]{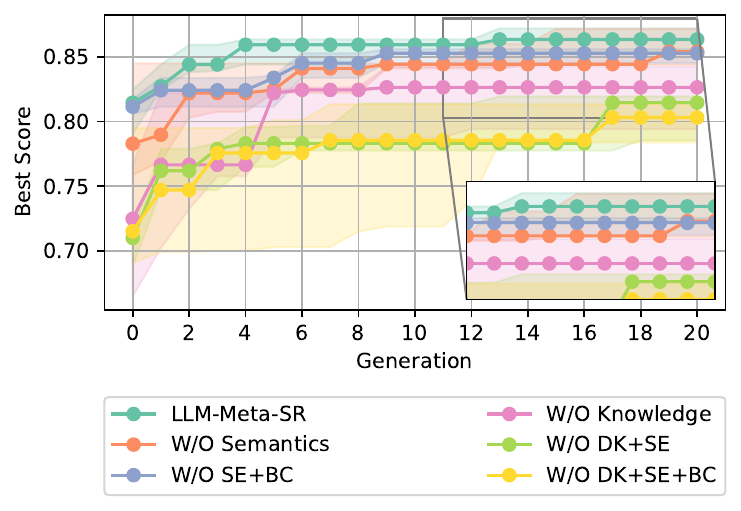}
                \caption{Evaluation $R^2$ of the best solution.}
                \label{fig:gpt5_r2}
            \end{subfigure}
            \hfill
            \begin{subfigure}[t]{0.32\textwidth}
                \centering
                \includegraphics[width=\textwidth, trim=5pt 5pt 5pt 5pt, clip]{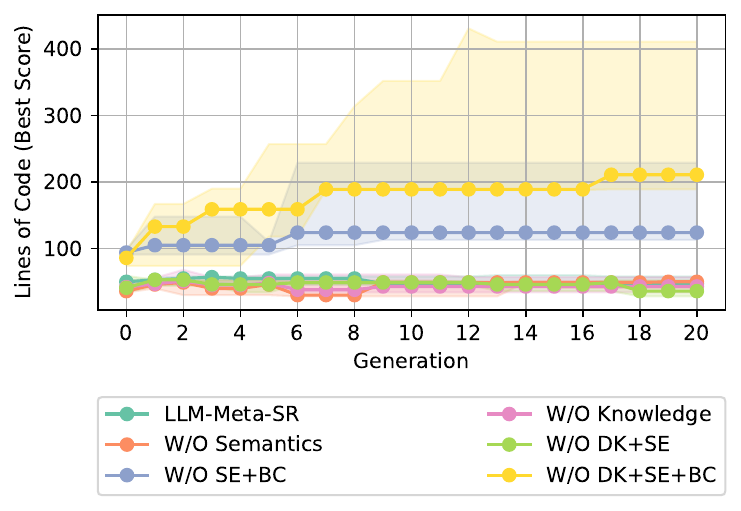}
                \caption{Code length of the best solution.}
                \label{fig:gpt5_code_length}
            \end{subfigure}
            \caption{Comparison across generations for different LLM-driven search strategies using \textit{GPT-5-Mini}.}
            \label{fig:gpt5_ablation}
        \end{figure*}

        \textbf{LLM Model Comparison:} To assess the impact of model capability, we compare GPT-5-Mini with a hybrid configuration that combines GPT-4.1-Mini and GPT-5-Mini, where each model is invoked with equal probability during evolution. Minor prompt refinements are introduced to align the process more closely with real algorithm development practices, and the detailed settings are provided in \Cref{sec: GPT-5 Prompt}. As shown in \Cref{fig:gpt5_comparison}, GPT-5-Mini consistently outperforms the hybrid configuration, demonstrating that stronger reasoning and code synthesis capabilities contribute to more effective operator discovery. Furthermore, \Cref{fig:gpt5_r2,fig:gpt5_code_length} confirm that the ablation trends observed with GPT-4.1-Mini remain consistent under GPT-5-Mini, indicating that the proposed semantic evolution, bloat control, and domain knowledge strategies generalize across different LLM backbones and prompts.

        \subsection{Discovered Operators}
        \label{sec: Discovered Operators}
        In this section, we present experiments on datasets from contemporary SR benchmarks to evaluate the performance of the discovered operators. Since the LLM-designed operator incorporates several desirable characteristics of selection operators—including diversity-awareness and dynamic selection pressure—it is referred to as Omni selection. The pseudocode for the Omni selection operator is provided in \Cref{alg:omni-selection}, and the full Python implementation is given in \Cref{lst:omni-selection} in the supplementary material. We focus on the operator with the best objective score on the meta-training tasks, with analysis of more operators in \Cref{sec: More Operators}.

\begin{figure*}[!t]
            \centering
            \begin{minipage}[t]{0.32\textwidth}
                \centering
                \includegraphics[width=\textwidth]{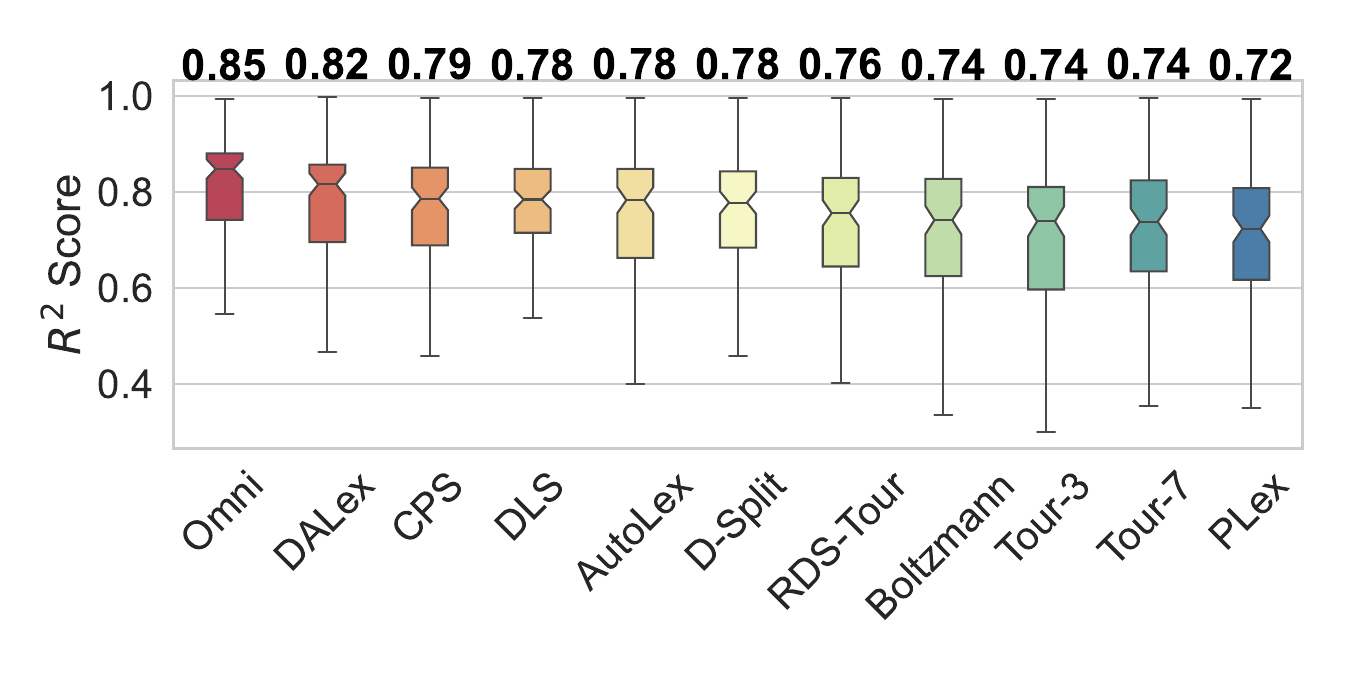}
                \caption{Test $R^2$ scores of different selection operators on SR benchmarks. Median values are annotated.}
                \label{fig: r2}
            \end{minipage}
            \hfill
            \begin{minipage}[t]{0.32\textwidth}
                \centering
                \includegraphics[width=\textwidth]{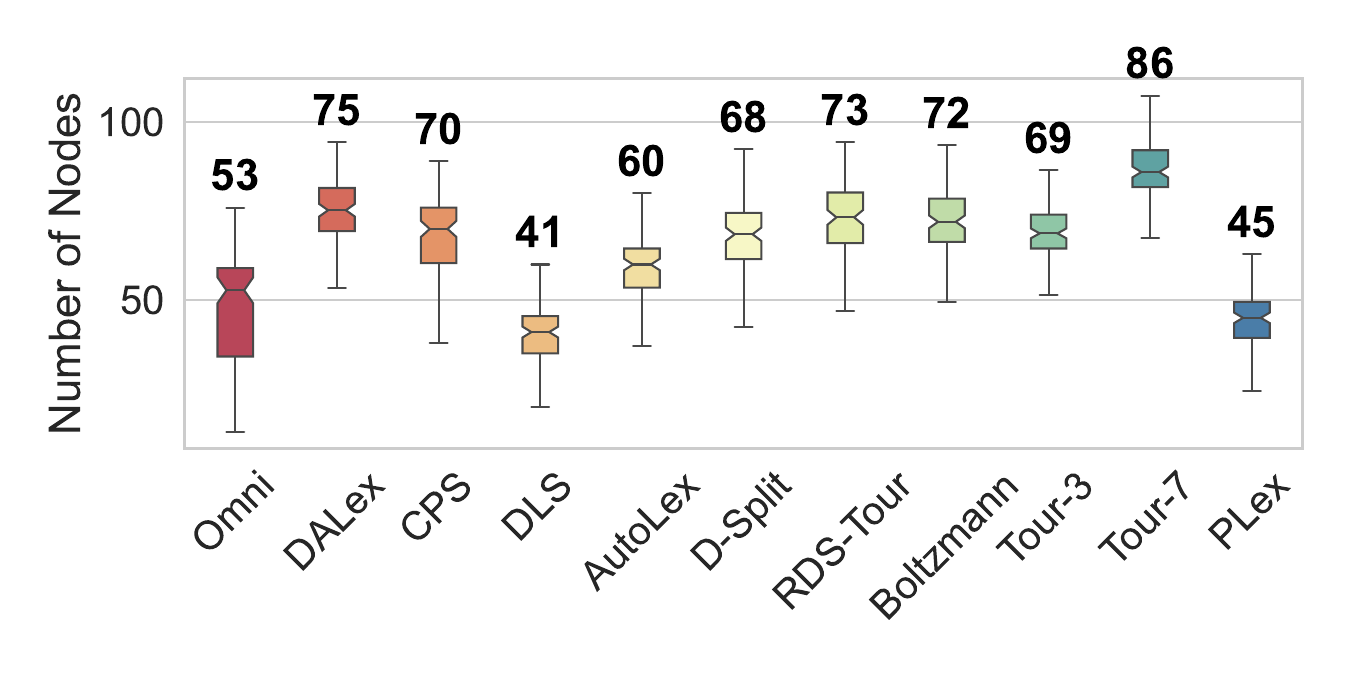}
                \caption{Tree sizes of selection operators on SR benchmarks.}
                \label{fig: complexity}
            \end{minipage}
            \hfill
            \begin{minipage}[t]{0.32\textwidth}
                \centering
                \includegraphics[width=\textwidth]{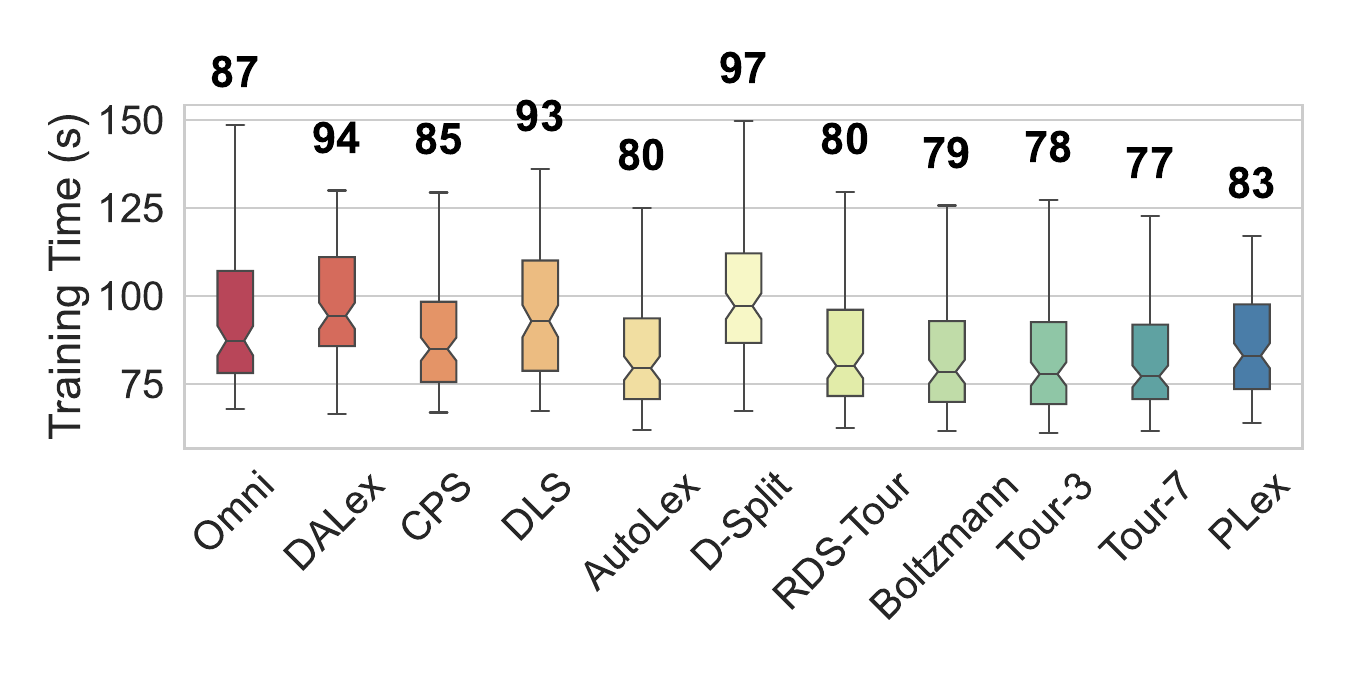}
                \caption{Training times of selection operators on SR benchmarks.}
                \label{fig: training_time}
            \end{minipage}
        \end{figure*}

\begin{figure}[!t]
            \centering
            \includegraphics[width=0.8\columnwidth]{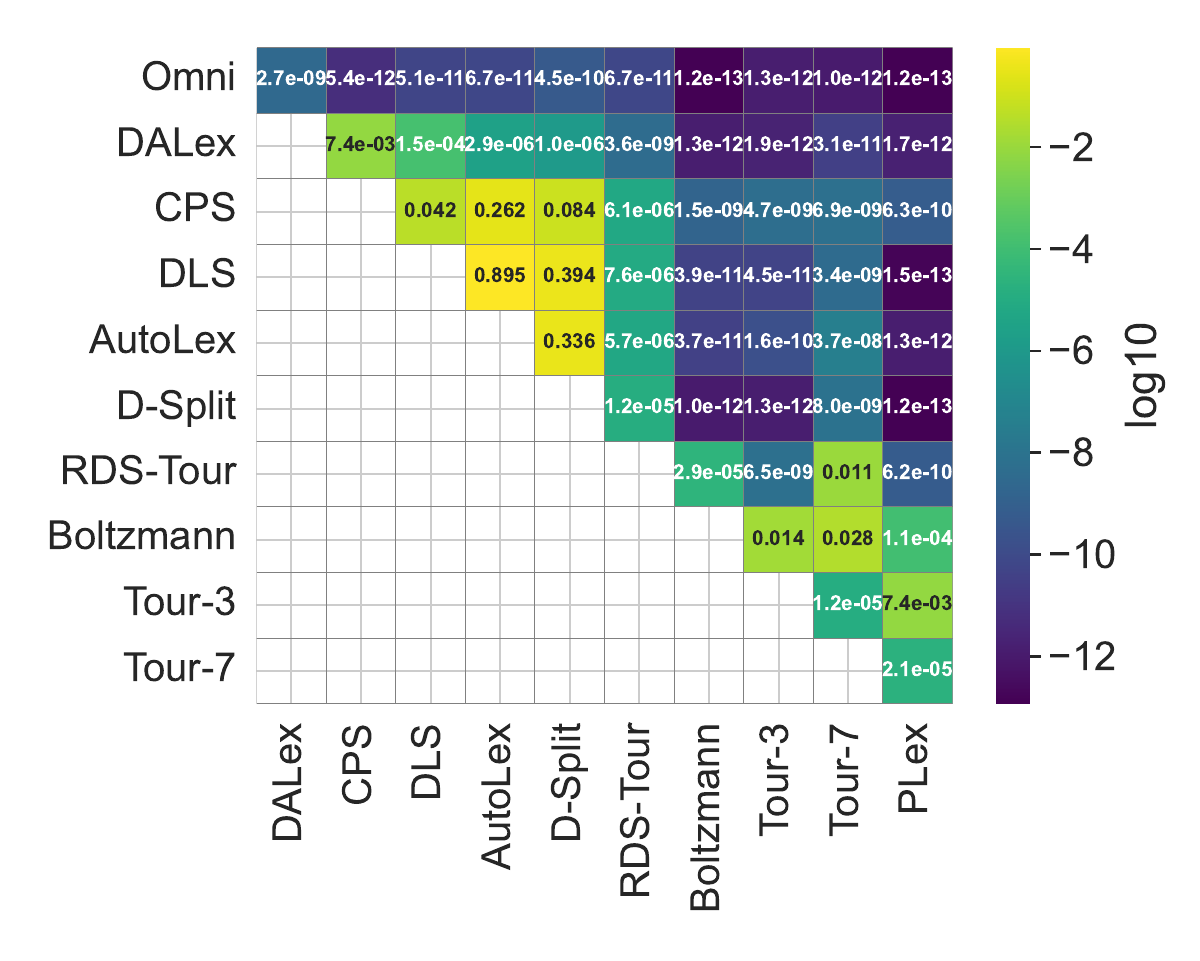}
            \caption{Pairwise statistical comparison of selection operators using the Wilcoxon signed-rank test with Benjamini--Hochberg correction.}
            \label{fig: p-value}
        \end{figure}

        \textbf{Test Accuracy:} The results in \Cref{fig: r2} show that the LLM-generated selection operator outperforms the expert-designed baselines in terms of $R^2$ scores and achieves the best overall performance. A Wilcoxon signed-rank test with Benjamini--Hochberg correction is presented in \Cref{fig: p-value}, and the detailed pairwise comparison results are provided in \Cref{tab:stats}. Detailed results on each dataset are reported in \Cref{tab:stats1} and \Cref{tab:stats2} in the supplementary material. These results demonstrate that the evolved Omni selection operator performs significantly better than existing expert-designed operators, many of which cannot be statistically distinguished from one another. This confirms that LLMs can effectively discover selection operators that surpass those created by domain experts.
        \begin{table*}[!t]
            \centering
            \caption{Pairwise statistical comparison of test \( R^2 \) scores for different selection operators on SR benchmarks. Each cell shows the number of datasets where the row method is statistically better (+), equivalent ($\sim$), or worse ({-}) than the column method based on the Wilcoxon signed-rank test with \( p = 0.05 \). The format is: number of wins (+)/number of ties ($\sim$)/number of losses ({-}).}
            \label{tab:stats}
            \resizebox{\linewidth}{!}{
                \begin{tabular}{ccccccccccc}%
                    \toprule%
                    & \textbf{DALex}           & \textbf{CPS}            & \textbf{DLS}             & \textbf{AutoLex}        & \textbf{D-Split}         & \textbf{RDS-Tour}& \textbf{Boltzmann}& \textbf{Tour-3}& \textbf{Tour-7}& \textbf{PLex}\\%
                    \midrule%
                    \textbf{Omni}      & 43(+)/60($\sim$)/13({-}) & 47(+)/61($\sim$)/8({-}) & 69(+)/37($\sim$)/10({-})& 66(+)/41($\sim$)/9({-})& 61(+)/42($\sim$)/13({-})& 70(+)/32($\sim$)/14({-})& 75(+)/34($\sim$)/7({-})& 71(+)/36($\sim$)/9({-})& 73(+)/35($\sim$)/8({-})& 78(+)/31($\sim$)/7({-})\\%
                    \textbf{DALex}     & ---                      & 18(+)/93($\sim$)/5({-}) & 29(+)/82($\sim$)/5({-})  & 24(+)/90($\sim$)/2({-})& 33(+)/81($\sim$)/2({-})& 43(+)/67($\sim$)/6({-})& 69(+)/43($\sim$)/4({-})& 67(+)/45($\sim$)/4({-})& 47(+)/66($\sim$)/3({-})& 76(+)/33($\sim$)/7({-})\\%
                    \textbf{CPS}       & ---                      & ---                     & 14(+)/94($\sim$)/8({-})  & 11(+)/99($\sim$)/6({-}) & 19(+)/92($\sim$)/5({-})& 28(+)/80($\sim$)/8({-})& 51(+)/63($\sim$)/2({-})& 56(+)/53($\sim$)/7({-})& 38(+)/70($\sim$)/8({-})& 68(+)/44($\sim$)/4({-})\\%
                    \textbf{DLS}       & ---                      & ---                     & ---                      & 7(+)/107($\sim$)/2({-}) & 8(+)/99($\sim$)/9({-})   & 24(+)/86($\sim$)/6({-})& 45(+)/69($\sim$)/2({-})& 52(+)/58($\sim$)/6({-})& 35(+)/74($\sim$)/7({-})& 76(+)/39($\sim$)/1({-})\\%
                    \textbf{AutoLex}   & ---                      & ---                     & ---                      & ---                     & 11(+)/102($\sim$)/3({-}) & 28(+)/82($\sim$)/6({-})& 44(+)/71($\sim$)/1({-})& 52(+)/58($\sim$)/6({-})& 31(+)/79($\sim$)/6({-})& 76(+)/40($\sim$)/0({-})\\%
                    \textbf{D-Split}   & ---                      & ---                     & ---                      & ---                     & ---                      & 16(+)/92($\sim$)/8({-})  & 35(+)/79($\sim$)/2({-})& 49(+)/64($\sim$)/3({-})& 25(+)/86($\sim$)/5({-})& 69(+)/46($\sim$)/1({-})\\%
                    \textbf{RDS-Tour}  & ---                      & ---                     & ---                      & ---                     & ---                      & ---                      & 18(+)/97($\sim$)/1({-}) & 26(+)/88($\sim$)/2({-})& 6(+)/110($\sim$)/0({-})& 43(+)/72($\sim$)/1({-})\\%
                    \textbf{Boltzmann} & ---                      & ---                     & ---                      & ---                     & ---                      & ---                      & ---                     & 3(+)/108($\sim$)/5({-}) & 6(+)/95($\sim$)/15({-})& 24(+)/88($\sim$)/4({-})\\%
                    \textbf{Tour-3}    & ---                      & ---                     & ---                      & ---                     & ---                      & ---                      & ---                     & ---                     & 4(+)/98($\sim$)/14({-}) & 24(+)/85($\sim$)/7({-}) \\%
                    \textbf{Tour-7}    & ---                      & ---                     & ---                      & ---                     & ---                      & ---                      & ---                     & ---                     & ---                     & 40(+)/69($\sim$)/7({-}) \\%
                    \bottomrule%
                \end{tabular}%
            }
        \end{table*}

        \begin{figure*}[!tbp]
            \centering
            \includegraphics[width=0.85\textwidth]{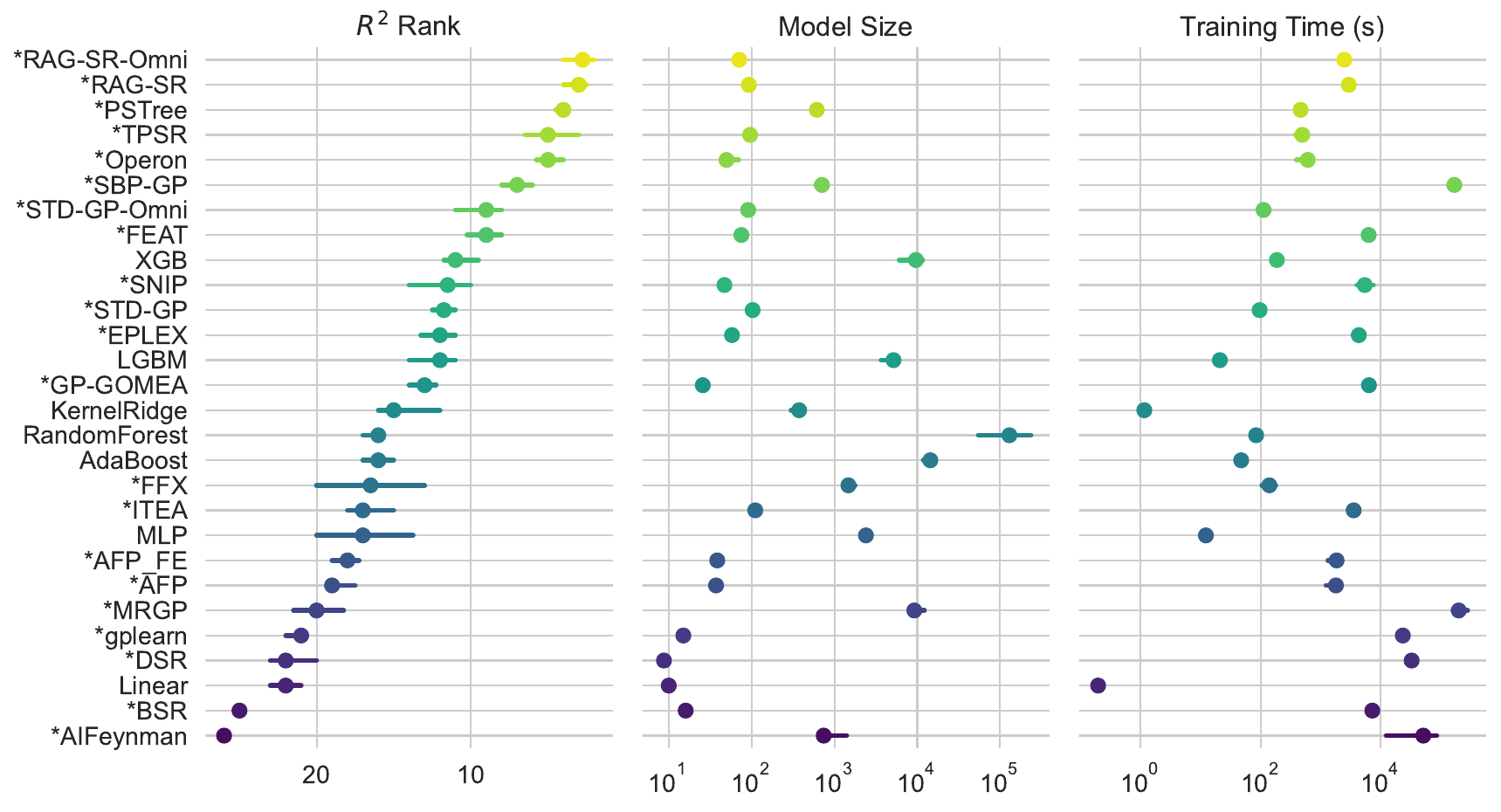}
            \caption{Median ranks of test $R^2$, model sizes, and training times of 28 algorithms on the SR benchmark. Algorithms marked with * are symbolic regression algorithms.}
            \label{fig: SRBench Results}
        \end{figure*}

        \textbf{Tree Size:} The distribution of model sizes is presented in \Cref{fig: complexity}. The results show that the evolved operator produces smaller models compared to top-performing selection operators like AutoLex and CPS. This is primarily because the evolved operator incorporates model size into the selection process, biasing the search toward regions where symbolic expressions are more compact. Compared to DLS, which is explicitly designed for bloat control, the evolved operator yields slightly larger but still competitive model sizes. However, unlike Omni, the DLS operator focuses mainly on controlling bloat, and as a result, its $R^2$, as shown in \Cref{fig: r2}, is not substantially higher than that of standard lexicase selection. Overall results in \Cref{fig: complexity} and \Cref{fig: r2} indicate that model size and performance are not always in conflict. By carefully designing the selection operator, it is possible to evolve symbolic models that achieve both high accuracy and small model size, indicating high interpretability.

        \textbf{Training Time:} As shown in \Cref{fig: training_time}, the proposed Omni selection operator is efficient, even though integrating multiple criteria into the selection process makes it slightly more time-consuming than some other operators in terms of overall SR time. However, it is important to consider not only the training phase but also the cost of evaluating and deploying candidate models. Since Omni selection tends to favor smaller models with fewer nodes, it reduces computational cost during evaluation and deployment, especially on resource-constrained devices. Thus, despite its multifaceted design leading to a slight increase in selection time, Omni remains an efficient and practical selection operator.

        \subsection{Comparing with State-of-the-Art SR Algorithms}
        \label{sec: Analysis on SOTA SR Algorithms}
        \textbf{Experimental Settings:}
        We apply the evolved operator to a state-of-the-art Transformer-assisted SR algorithm, namely retrieval-augmentation-generation-based SR (RAG-SR)~\cite{zhang2025ragsr}, to evaluate the effectiveness of the proposed operator in the context of modern SR. The only modification is the replacement of the automatic epsilon lexicase selection~\cite{la2019probabilistic} with the repaired variant Omni-R of the Omni selection operator, described in \Cref{sec:omni_small_dataset_repair}, which corrects a limitation of the original Omni on small datasets. All other components remain unchanged. The resulting algorithm is referred to as RAG-SR-Omni. For the experimental datasets, following the setup in \Cref{sec: Experimental Settings}, we use 116 out of 120 regression problems, excluding the four used during meta-evolution. To assess whether the evolved operator also benefits SR methods with more moderate performance, we additionally compare standard GP with lexicase selection and standard GP with Omni-R, using single-tree, linear scaling, and the same primitives and evaluation protocol as in \Cref{sec: Discovered Operators}.

        \textbf{Experimental Results:}
        The results are presented in \Cref{fig: SRBench Results}. The figure shows that RAG-SR-Omni outperforms RAG-SR in terms of median ranks of test $R^2$, model sizes, and training time, achieving the best performance among 28 symbolic regression and machine learning algorithms. These results indicate that the LLM-evolved selection operator can be seamlessly integrated into state-of-the-art SR methods to further enhance their performance.

        Recall that RAG-SR uses automatic epsilon lexicase selection~\cite{zhang2025ragsr}, which only considers diversity awareness. These results suggest that even for a state-of-the-art SR algorithm, it is still desirable to consider multiple aspects, such as diversity, dynamic selection pressure, complementarity, and interpretability, during selection to achieve stronger performance. Nonetheless, manually designing such an operator is challenging, which highlights the value of leveraging LLMs in this context.
        To investigate whether the evolved operator also benefits SR methods that are currently less competitive, we further compare standard GP with lexicase selection and standard GP with Omni selection. Standard GP with Omni outperforms standard GP with lexicase selection. Moreover, using SNIP, a deep learning-based symbolic regression method~\cite{meidani2024snip}, as a reference, standard GP with lexicase selection performs worse than SNIP, whereas standard GP with Omni performs better than SNIP in terms of $R^2$ on the same 116 SRBench datasets. This result indicates that the performance gains from the evolved operator generalize beyond already strong frameworks.

        \section{Further Analysis}
        \label{sec: Further Analysis}

        \begin{table*}[!t]
            \small
            \setlength{\tabcolsep}{4pt}
            \caption{Comparison of key selection components between parents and offspring.}
            \begin{threeparttable}
                \label{tab:selection_fusion}
                \centering
                \begin{tabularx}{\textwidth}{l *{3}{>{\raggedright\arraybackslash}X}}
                \toprule
                \textbf{Component} & \textbf{Parent 1}                                                               & \textbf{Parent 2} & \textbf{Generated Code} \\
                \midrule
                Complexity penalty
                & $(s+h)\times\dfrac{0.3+0.7\,\text{stage}}{100}$
                & $\displaystyle \dfrac{s+h}{30+70\,\text{stage}}$
                & $(s+h)\times\dfrac{0.4+0.6\,\text{stage}}{40+60\,\text{stage}}$                                                               \\

                Subset size
                & $\displaystyle \max\!\bigl(7,\lfloor\tfrac{n_c}{\tfrac{k}{2}+3}\rfloor\bigr)$
                & $\displaystyle \max\!\bigl(10,\lfloor\tfrac{n_c}{\tfrac{k}{2}+1}\rfloor\bigr)$
                & $\displaystyle \max\!\bigl(8,\lfloor\tfrac{n_c}{\tfrac{k}{2}+2}\rfloor\bigr)$                                                 \\

                Parent A scoring
                & $(1-\text{stage})\times(\mathrm{mse}_{\text{full}}-\mathrm{mse}_{\text{sub}})\;
                +\;\text{stage}\times\dfrac{1}{\mathrm{mse}_{\text{full}}}\;
                -\;\mathrm{comp\_pen}$
                & $(1-\text{stage})\times\dfrac{1}{\mathrm{mse}_{\text{sub}}}
                +\;\text{stage}\times\dfrac{1}{\mathrm{mse}_{\text{full}}}
                -\;\mathrm{comp\_pen}$
                & Same as Parent 1                                                                                                              \\

                Parent B scoring
                & $corr + \dfrac{\mathrm{comp\_pen}}{1 + 5\,\text{stage}}$
                & $corr + \dfrac{\mathrm{comp\_pen}}{10}$
                & $corr + \dfrac{\mathrm{comp\_pen}}{\begin{cases}
                                                         5, & \text{if stage}<0.5 \\10,&\text{otherwise}
                \end{cases}}$                                             \\
                \bottomrule
                \end{tabularx}
                \begin{tablenotes}
                    \footnotesize
                    \item[*] Here $s$ denotes size, $h$ denotes height, $n_c$ is the number of cases, $k$ is the number of individuals in the population, $\text{stage}$ is the normalized generation (ranging from 0 to 1), $\mathrm{comp\_pen}$ is the complexity penalty, $\mathrm{mse}_{\text{full}}$ is the mean squared error on the full training set, $\mathrm{mse}_{\text{sub}}$ is the mean squared error on the subset, and \textit{corr} denotes the normalized residual correlation.
                \end{tablenotes}
            \end{threeparttable}
        \end{table*}

        In this section, we provide a deeper analysis of the proposed LLM-Meta-SR framework across two complementary aspects. \Cref{sec: Semantic-Aware Crossover} analyzes the semantics-aware evolution mechanism, demonstrating how LLM-based crossover effectively fuses complementary code fragments. In \Cref{sec: Code Evolved by LLM-Meta-SR}, we examine the characteristics and behavior of the evolved Omni selection operator through static and benchmark analyses, highlighting its interpretability, diversity, and efficiency.

        \subsection{Analysis of Semantics-Aware Evolution}
        \label{sec: Semantic-Aware Crossover}
        Our meta-learning framework explicitly selects complementary parent algorithms for crossover and illustrates this effect with the example presented in this section. Prior work typically combines operators through dynamic algorithm selection~\cite{xue2022multi}, which stores a set of algorithms to select from under different conditions and can lead to exponential code growth after repeated merges of algorithms~\cite{martins2018solving}. In contrast, our method fuses meaningful code fragments from both parents, yielding structured offspring rather than simple concatenations. The LLM-based crossover provides a unique opportunity for intelligent fusion. Compared with mutation, which performs a local random search around an elite operator, the LLM-based crossover can provide a directed way to blend complementary capabilities and produce operators that generalize across datasets.

        \Cref{tab:selection_fusion} shows how the LLM synthesizes the offspring operator by combining key components from both parents. The complexity penalty in the offspring, $(s+h)\times\frac{0.4+0.6\,\text{stage}}{40+60\,\text{stage}}$, interpolates between parent 1's $(s+h)\times\frac{0.3+0.7\,\text{stage}}{100}$ and parent 2's $\frac{s+h}{30+70\,\text{stage}}$. More importantly, the parent B scoring formula demonstrates dynamic selection pressure adaptation: the offspring uses $corr + \frac{\mathrm{comp\_pen}}{\begin{cases}
                                                                                                                                                                                                                                                                                                                                                                                                                                                                                                                                              5, & \text{if stage}<0.5 \\10,&\text{otherwise}
        \end{cases}}$, which combines parent 1's penalty $corr + \frac{\mathrm{comp\_pen}}{1 + 5\,\text{stage}}$ with parent 2's penalty $corr + \frac{\mathrm{comp\_pen}}{10}$. This blending strategy enables the offspring to balance selection pressures across different search phases. Detailed code comparisons are provided in \Cref{lst:offspring}, \Cref{lst:parent a}, and \Cref{lst:parent b} of the supplementary material.

        \begin{figure}[!t]
            \centering
            \includegraphics[width=0.6\columnwidth]{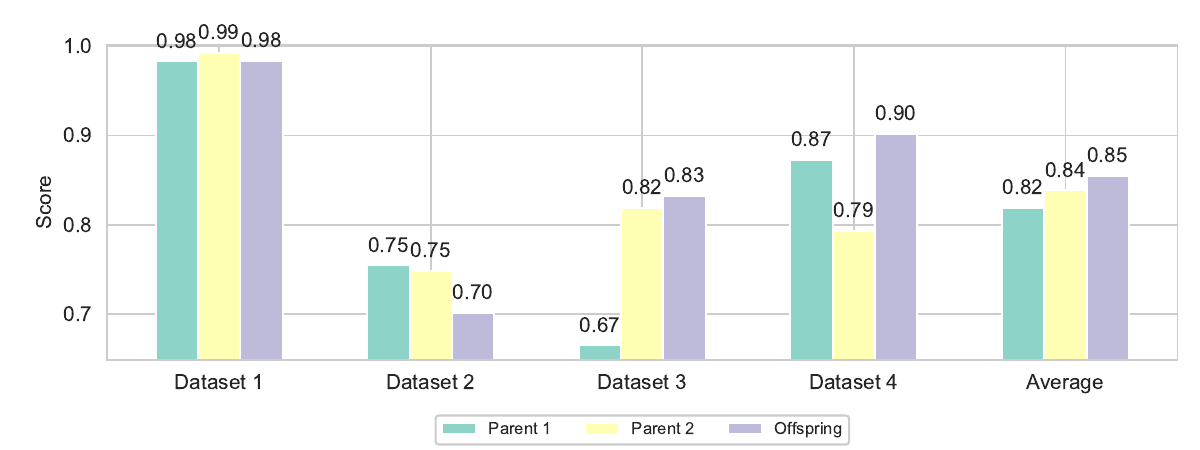}
            \caption{Evaluation $R^2$ scores of parent and offspring selection operators on different datasets.}
            \label{fig: score}
        \end{figure}

        As shown in \Cref{fig: score}, parent 1 excels on dataset 4 with 50 features, which corresponds to a relatively small search space, while parent 2 excels on dataset 3 with 100 features, which corresponds to a relatively large search space. Parent 1 employs a more relaxed selection pressure, while parent 2 uses a stricter selection pressure. By combining their complementary strengths through the mixing strategy performed by the LLM in \Cref{tab:selection_fusion}, the offspring operator balances these selection pressures, achieving strong performance on both datasets 3 and 4, thereby yielding better average performance. Unlike random subtree crossover in tree-based GP~\cite{banzhaf1998genetic} or one-point crossover in linear GP~\cite{huang2024toward}, LLM-based crossover can synthesize code that is semantically intermediate between the two parent programs by leveraging its code understanding capability. This highlights the importance of semantic diversity: if both parents are highly similar, even if they are performant, the resulting offspring would be nearly identical to the parents, leading to redundant evaluations and wasted computation. Therefore, semantics-aware evolution is vital for automated algorithm design.

        \subsection{Code Evolved by LLM-Meta-SR}
        \label{sec: Code Evolved by LLM-Meta-SR}

        \begin{algorithm}[!t]
        \caption{Omni Selection}
        \label{alg:omni-selection}
        \begin{algorithmic}[1]
            \REQUIRE Population $\mathcal{F} = \{\Phi_1, \dots, \Phi_\lambda\}$; population size $\lambda$; evolutionary stage $\alpha \in [0,1]$; target values $\mathbf{y}$
            \STATE $K \gets \lambda / 2$ \hfill // Parent pairs
            \STATE $m \gets \max(7, \lfloor |\mathbf{y}| / (2K) \rfloor)$ \hfill // Subset size
            \STATE $S_j^{\text{seq}} \gets \mathbf{y}[(j-1) \cdot m : j \cdot m]$ for $j = 1, \dots, K/2$ \hfill // Sequential slices
            \STATE $S_j^{\text{rnd}} \gets \text{RandomSample}(\{0,\dots,|\mathbf{y}|\!-\!1\},\; m)$ for $j = 1, \dots, K/2$ \hfill // Random subsets
            \STATE $\mathcal{S} \gets \{S_1^{\text{seq}}, \dots\} \cup \{S_1^{\text{rnd}}, \dots\}$ \hfill // Total $K$ subsets
            \STATE \textbf{// Compute residuals and complexity}
            \FOR{each $\Phi_i \in \mathcal{F}$}
            \STATE $\mathbf{r}_i \gets \mathbf{y} - \hat{\mathbf{y}}_i$ \hfill // Residual vector
            \STATE $c_i \gets (\text{size}(\Phi_i) + \text{depth}(\Phi_i)) \,/\, c_{\max}$ \hfill // Normalized complexity
            \ENDFOR
            \STATE $\mu \gets 0.25 + 0.25 \cdot \alpha$ \hfill // Generation-dependent complexity weight
            \STATE \textbf{// Select Parent A}
            \FOR{each subset $S_j \in \mathcal{S}$}
            \STATE $\text{MSE}_{i,j} \gets \frac{1}{|S_j|} \sum_{t \in S_j} r_{i,t}^2$ for all $\Phi_i$
            \STATE $A_j \gets \arg\min_{\Phi_i} (\text{MSE}_{i,j},\; c_i)$ \hfill // Lexicographic sorting
            \ENDFOR
            \STATE \textbf{// Select Parent B}
            \FOR{each $A_j$}
            \STATE $\text{sim}_i \gets |\cos(\mathbf{r}_{A_j},\; \mathbf{r}_i)|$ for all $\Phi_i$
            \STATE $B_j \gets \arg\min_{\Phi_i} \left(\text{sim}_i + \mu \cdot c_i\right)$
            \ENDFOR
            \RETURN $\{(A_1, B_1), (A_2, B_2), \dots, (A_K, B_K)\}$ \hfill // $\lambda$ parents
        \end{algorithmic}
        \end{algorithm}

        \begin{figure}[!t]
            \centering
            \captionsetup[subfigure]{skip=2pt}
            \begin{subfigure}[b]{\columnwidth}
                \centering
                \includegraphics[width=\columnwidth]{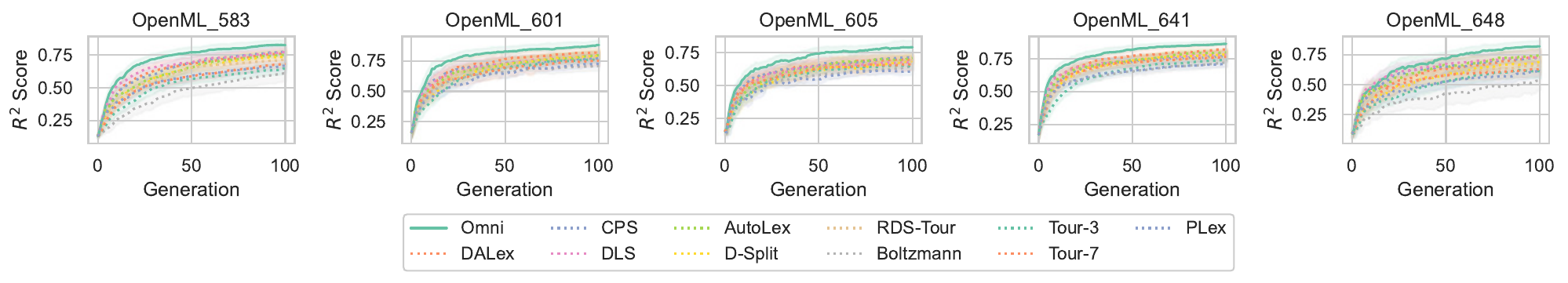}
                \caption{Test $R^2$ scores.}
                \label{fig: test r2}
            \end{subfigure}
            \begin{subfigure}[b]{\columnwidth}
                \centering
                \includegraphics[width=\columnwidth]{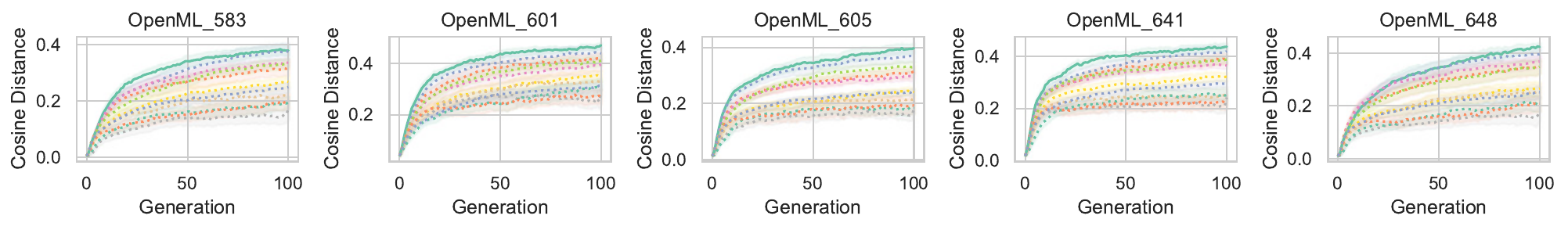}
                \caption{Population cosine semantic diversity.}
                \label{fig: cosine distance}
            \end{subfigure}
            \begin{subfigure}[b]{\columnwidth}
                \centering
                \includegraphics[width=\columnwidth]{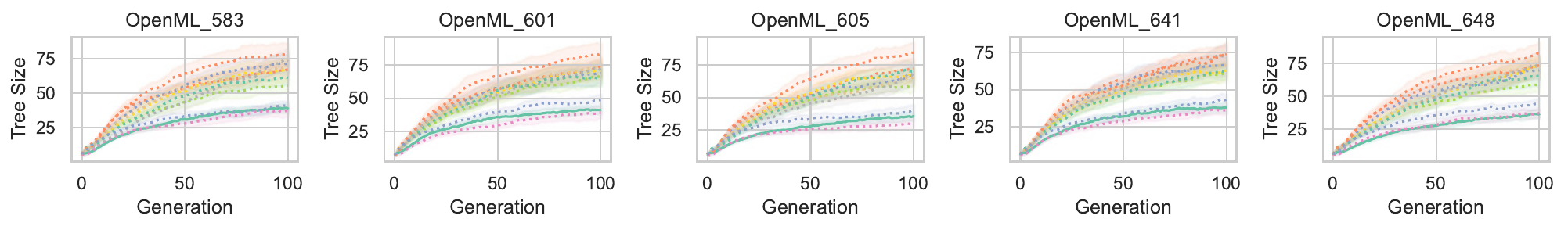}
                \caption{Tree sizes.}
                \label{fig: tree size}
            \end{subfigure}
            \caption{Evolution plots of different selection operators across generations.}
            \label{fig: evolution}
        \end{figure}

        \subsubsection{Code Analysis}
        \Cref{alg:omni-selection} presents the highest-performing operator evolved through meta-evolution. The operator first constructs $K$ subsets of the training data. Half of these subsets are structured (contiguous blocks), and the other half are sampled randomly. Moreover, a normalized complexity measure is calculated for each individual (lines~2--10). During the selection of the first parent (lines~12--16), candidates are lexicographically ranked by their subset-specific mean squared error followed by their complexity, thereby prioritizing specialists on individual subsets while retaining a secondary preference for parsimonious models when errors are comparable, thus enhancing interpretability without sacrificing accuracy.

        The second parent (lines~18--21) is selected by computing the absolute cosine similarity between the residual vector of each candidate and that of the first parent~A, to which a generation-dependent complexity penalty is added. Consequently, early generations with small penalties emphasize semantic complementarity, whereas later generations with larger penalties bias the search toward concise yet complementary models. This mechanism explicitly couples diversity preservation with dynamic selection pressure.

        Overall, the operator satisfies all design principles in \Cref{sec: Operator Design Principle}: interpretability through explicit complexity control, diversity and complementarity via residual-based similarity, dynamic selection pressure through the adaptive penalty, and vectorization because all semantics-related computations are performed using vectorized operations.

        \subsubsection{Benchmark Analysis}
        \label{sec: Benchmark Analysis}
        To further analyze the behavior of the Omni selection operator, we plot the test $R^2$ scores on five representative datasets in \Cref{fig: test r2}. The diversity of the population is shown in \Cref{fig: cosine distance}, and the tree size is shown in \Cref{fig: tree size}.

        \textbf{Test $R^2$ Scores:}
        The test $R^2$ scores demonstrate that the Omni selection operator gains an early advantage and maintains it throughout the search, indicating that it is an effective selection operator for anytime performance~\cite{ye2022automated}.

        \textbf{Population Diversity:}
        To understand why the Omni selection operator achieves this performance, we plot the diversity trajectory of the symbolic-regression population in \Cref{fig: cosine distance}. Diversity is measured by the cosine distance between GP individuals, which is preferred over Euclidean distance because the latter can be trivially large if some solutions have very large errors. The results show that population diversity is well maintained across generations and is better preserved than in the baseline. In other words, the Omni selection operator maintains a more diverse population of SR models, with individuals making errors on different subsets of instances. This helps explain the superior performance of the Omni selection operator compared to the baseline.

        \textbf{Tree Size:}
        Finally, we plot the tree size trajectory in \Cref{fig: tree size}. The results show that Omni selection not only achieves better performance but also exhibits less rapid growth in tree size than other selection operators throughout the entire evolution process. This suggests that the Omni selection operator explores regions of the search space containing parsimonious solutions more exhaustively than other operators do, which is ideal for finding effective and parsimonious solutions in SR.

        \section{Conclusions and Future Work}
        \label{sec: conclusion}
        In this paper, we demonstrate that LLMs can discover selection operators that outperform those crafted by domain experts. The evolved Omni selection operator surpasses expert-designed baselines and boosts the performance of state-of-the-art SR algorithms, achieving the best performance among 28 symbolic regression and machine learning algorithms across a diverse range of tasks. Omni also controls model growth competitively. Ablation studies confirm that addressing three key challenges—limited semantic awareness, code bloat, and lack of domain knowledge—is crucial for success. Semantic selection and feedback mechanisms enable fine-grained semantic distinction, while prompt-based length control and multi-objective survival selection effectively mitigate bloat. Incorporating domain knowledge principles into prompts helps guide the LLM to generate operators with desired properties.

        We emphasize that LLM-Meta-SR is a general-purpose framework for automated algorithm component design rather than a method limited to evolving selection operators for GP-based symbolic regression. The framework can be readily applied to design other algorithmic components, such as crossover and mutation operators, and can be extended to broader GP-based tasks beyond regression, including classification. In this work, selection operator design serves as a representative case study. For future work, it is worth exploring the automatic design of crossover and mutation operators across diverse evolutionary computation domains to further reduce the effort required for developing novel algorithms. 
        Another promising direction is adapting the meta-evolution evaluation protocol to use a fixed computational budget rather than a fixed number of generations, which would enable the automated discovery of effective selection operators under a specific time constraint.

\bibliographystyle{IEEETransDOI}
\bibliography{aaai2026}

    \clearpage
    \appendices

    \ifCLASSOPTIONcaptionsoff
    \newpage
    \fi

    \clearpage
    \appendices

        The supplementary material provides additional details and analyses that complement the main paper. The appendices are organized as follows:
        \begin{itemize}
            \item \Cref{sec: ESR} describes the evolutionary SR algorithm workflow and parameter settings used in the experiments, supporting \Cref{sec: Algorithm Framework} and \Cref{sec: Experimental Settings}.
            \item \Cref{sec: Analysis of Evolved Operator} provides the complete code of the Omni selection operator, analyzes an operator evolved without domain knowledge, presents a repaired version for small datasets, and reports results on additional discovered operators, supporting \Cref{sec: Discovered Operators}.
            \item \Cref{sec: More Analysis on SR Benchmark} provides further analysis of the SR benchmark results, including top-5 algorithm comparisons and Pareto front analysis, supporting \Cref{sec: Analysis on SOTA SR Algorithms}.
            \item \Cref{sec: Crossover and Mutation Ratio} investigates the effect of different crossover and mutation ratios on the meta-evolution process, supporting \Cref{sec: Meta-Evolution Results}.
            \item \Cref{sec: evaluation_time} reports an evaluation-time analysis of the meta-evolution process, supporting \Cref{sec: Meta-Evolution Results}.
            \item \Cref{sec: GPT-5 Prompt} describes the prompting improvements made for the GPT-5 experiment, supporting \Cref{sec: Meta-Evolution Results}.
            \item \Cref{sec: Prompt} presents the complete prompts used for algorithm evolution, including system, initialization, crossover, and mutation prompts, supporting \Cref{sec: Algorithm Framework}.
            \item \Cref{sec: infrastructure} lists the computing infrastructure and software packages used in the experiments, supporting \Cref{sec: Experimental Settings}.
        \end{itemize}

        \section{Evolutionary SR}
        \label{sec: ESR}

        \subsection{Algorithm Workflow}
        The SR algorithm is implemented using the GP framework~\cite{banzhaf1998genetic}. Let \( \mathcal{F}^{(t)} = \{ \Phi_1, \Phi_2, \dots, \Phi_\lambda \} \) denote the population of \( \lambda \) symbolic expressions at generation \( t \). The evolutionary process proceeds as follows:
        \begin{itemize}
            \item \textbf{Population Initialization:} The initial population \( \mathcal{F}^{(0)} \) is generated using the ramped half-and-half method~\cite{banzhaf1998genetic}.

            \item \textbf{Fitness Evaluation:} Each expression \( \Phi_i \in \mathcal{F}^{(t)} \) is evaluated on the dataset \( (X, Y) \). Let \( \mathbf{z}_i = \Phi_i(X) \in \mathbb{R}^n \) denote the vector of symbolic outputs. The linear coefficients \( \alpha_i \) and \( \beta_i \) are computed by fitting a ridge regression model of the form \( \hat{Y}_i = \alpha_i \cdot \mathbf{z}_i + \beta_i \). The leave-one-out cross-validation (LOOCV)~\cite{allen1974relationship} squared error is computed efficiently as $\mathcal{E}_i = \sum_{j=1}^{n} \left( \frac{r_{ij}}{1 - H_{ijj}} \right)^2,$ where \( \mathbf{r}_i = Y - \hat{Y}_i \) is the residual vector, and \( H_i = Z_i(Z_i^\top Z_i + \lambda I)^{-1} Z_i^\top \) is the hat matrix for ridge regression with input \( Z_i = [\mathbf{z}_i \ \mathbf{1}] \in \mathbb{R}^{n \times 2} \). The full error vector is retained for use in selection.

            \item \textbf{Elitism:} The best-performing expression \( \Phi^* = \arg\min_{\Phi_i \in \mathcal{F}^{(t)}} \mathcal{E}_i \) is preserved in an external archive and used for solution selection to maintain historical performance.

            \item \textbf{Solution Selection:} The LLM-evolved selection operator is invoked to select a set of promising parent expressions from the current population and the elite archive.

            \item \textbf{Solution Generation:} A new population \( \mathcal{F}^{(t+1)} \) is generated from the selected parents using standard GP operators:
            \begin{itemize}
                \item \textbf{Subtree Crossover:} Given two parent expressions \( \Phi_a \) and \( \Phi_b \), offspring are generated by exchanging randomly selected subtrees.
                \item \textbf{Subtree Mutation:} A randomly selected subtree of an expression \( \Phi_i \) is replaced with a newly generated subtree.
            \end{itemize}
            During solution generation, all generated solutions are stored in a hash set. If an offspring is identical to any historical solution, it is discarded and a new solution is generated. The hash set is used because it maintains $O(1)$ query complexity. This mechanism prevents SR from wasting resources by evaluating the same solution multiple times.
        \end{itemize}

        \subsection{Parameter Settings}
        The parameters for SR follow conventional settings. The population size and number of generations are set to match those used in D-Split~\cite{imai2024minimum}. To prevent division-by-zero errors, we use the analytical quotient ($\mathrm{AQ}(x, y) = \frac{x}{\sqrt{1 + y^2}}$)~\cite{ni2012use} in place of the standard division operator.

        \begin{table}[h]
            \centering
            \caption{Parameter settings of SR}
            \label{tab: gp_params}
            \begin{tabular}{ll}
                \toprule
                Parameter             & Value                            \\
                \midrule
                Population size       & 100                              \\
                Maximum generations   & 100                              \\
                Maximum tree depth    & 10                               \\
                Initialization method & Ramped half-and-half (depth 0-6) \\
                Function set &
                \makecell[l]{$+,\, -,\, \times,\, \mathrm{AQ},\, \sqrt{|\cdot|},\, $ \\
                    $\log(1+|\cdot|),\, |\cdot|,\, (\cdot)^2, \sin_\pi(\cdot),\,$        \\
                    $\cos_\pi(\cdot),\, \mathrm{Max},\, \mathrm{Min},\, \mathrm{Neg}$} \\
                Crossover rate        & 0.9                              \\
                Mutation rate         & 0.1                              \\
                \bottomrule
            \end{tabular}
        \end{table}

        \section{Analysis of an Evolved Selection Operator}
        \label{sec: Analysis of Evolved Operator}

        \subsection{Code of Omni Selection}
        \label{sec: omni-selection-code}
        The complete Python implementation of the Omni selection operator is provided in \Cref{lst:omni-selection} for reproducibility. The corresponding pseudocode is presented in \Cref{alg:omni-selection} in the main text.

        \begin{lstlisting}[caption={Omni Selection}, label={lst:omni-selection}]
def omni_selection(population, k=100, status={}):
    stage = np.clip(status.get("evolutionary_stage", 0), 0, 1)
    n = len(population)
    half_k = k // 2
    y = population[0].y
    n_cases = y.size

    ssize = max(7, n_cases // max(1, 2 * half_k))
    half_struct = half_k // 2
    structured = [
        np.arange(i * ssize, min((i + 1) * ssize, n_cases)) for i in range(half_struct)
    ]
    random_ = [
        np.random.choice(n_cases, ssize, replace=False)
        for _ in range(half_k - half_struct)
    ]
    subsets = structured + random_

    residuals = np.vstack([ind.y - ind.predicted_values for ind in population])
    complexity = np.array([len(ind) + ind.height for ind in population], float)
    complexity /= max(1, complexity.max())
    subset_mse = np.array(
        [
            [((residuals[i, s]) ** 2).mean() if s.size else np.inf for s in subsets]
            for i in range(n)
        ]
    )
    comp_factor = 0.25 + 0.25 * stage

    parent_a = [
        population[np.lexsort((complexity, subset_mse[:, i]))[0]]
        for i in range(len(subsets))
    ]

    idx_map = {ind: i for i, ind in enumerate(population)}
    norms = np.linalg.norm(residuals, axis=1) + 1e-12

    parent_b = []
    for a in parent_a:
        i_a = idx_map[a]
        res_a = residuals[i_a]
        cors = (residuals @ res_a) / (norms * norms[i_a])
        cors[i_a] = 1
        comp_score = np.abs(cors) + comp_factor * complexity
        b_idx = np.argmin(comp_score)
        parent_b.append(population[b_idx])

    return [ind for pair in zip(parent_a, parent_b) for ind in pair][:k]
        \end{lstlisting}

        \subsection{Code Evolved by LLM-Meta-SR without Domain Knowledge}
        In this section, we analyze the code evolved by LLM-Meta-SR without domain knowledge to understand how an LLM can generate algorithms based solely on its internal knowledge. The evolved code is shown in \Cref{lst:omni-selection-free}. We refer to the resulting algorithm as Omni-Zero because it evolves the operator from scratch, in the spirit of AutoML-Zero~\cite{real2020automl}.

        \subsubsection{Code Analysis}
        Based on the code in \Cref{lst:omni-selection-free}, the LLM can evolve selection operators with desirable properties even without domain-specific guidance. These properties include diversity-awareness, dynamic selection pressure, and interpretability-awareness. For diversity-awareness, the selection operator uses cosine similarity to measure pairwise distances between individuals. Formally, it defines the novelty score as:
        \begin{equation}
            \mathrm{nov}_i = \frac{1 - \frac{1}{n} \sum_{k=1}^n \left\langle \frac{\mathbf{x}_i - \bar{x}_i}{\|\mathbf{x}_i - \bar{x}_i\|_2 + \epsilon}, \frac{\mathbf{x}_k - \bar{x}_k}{\|\mathbf{x}_k - \bar{x}_k\|_2 + \epsilon} \right\rangle}{\max\limits_{j} \left[ 1 - \frac{1}{n} \sum_{k=1}^n \left\langle \frac{\mathbf{x}_j - \bar{x}_j}{\|\mathbf{x}_j - \bar{x}_j\|_2 + \epsilon}, \frac{\mathbf{x}_k - \bar{x}_k}{\|\mathbf{x}_k - \bar{x}_k\|_2 + \epsilon} \right\rangle \right]}
        \end{equation}
        This cosine-distance-based novelty score is better suited for SR than Euclidean-distance-based novelty scores, as it is more robust to outliers and cannot be easily manipulated by introducing an outlier that artificially increases the distance, as discussed in \Cref{sec: Benchmark Analysis}.

        For dynamic selection pressure, the selection operator applies a nonlinear weighting function to control the trade-off between error and novelty. The function is defined as $0.35 + 0.3 \left(1 - \frac{t}{T_{\max}}\right)^{0.8}$, where $t$ is the current generation and $T_{\max}$ is the maximum number of generations, as described in \Cref{lst:omni-selection-free}. This function gradually decreases over time, promoting exploration in early generations and exploitation in later ones. This dynamic trade-off satisfies the dynamic selection pressure criterion defined in \Cref{sec: Operator Design Principle}.

        For interpretability-awareness, the selection operator uses both the number of nodes and the height of the tree to measure solution complexity. A linear weighting function is used to balance interpretability and accuracy during selection.

        A limitation of this operator is that it does not consider diversity from the perspective of specificity; selection is largely dominated by each individual's mean squared error. Furthermore, it does not account for the semantic complementarity between two parents.

        \begin{lstlisting}[caption={Omni Selection Zero}, label={lst:omni-selection-free}]
def omni_selection(population, k=100, status={}):
    import numpy as np
    evo_stage = status.get("evolutionary_stage", 0.)
    n = len(population)
    if n == 0:
        return []

    # Extract metrics
    errs = np.array([np.mean(ind.case_values) for ind in population], dtype=float)
    sizes = np.array([len(ind) for ind in population], dtype=float)
    heights = np.array([ind.height for ind in population], dtype=float)
    residuals = np.array([ind.y - ind.predicted_values for ind in population])
    preds = np.array([ind.predicted_values for ind in population])

    def safe_norm(arr):
        m = max(arr.max(), 1e-10)
        return arr / m

    norm_size, norm_height = safe_norm(sizes), safe_norm(heights)
    res_vars = np.var(residuals, axis=1)
    res_norms = np.linalg.norm(residuals, axis=1)
    norm_var, norm_resnorm = safe_norm(res_vars), safe_norm(res_norms)
    norm_err = safe_norm(errs)

    # Complexity metric combines more residual stats while balancing structural features
    complexity = (norm_size + norm_height + norm_var + norm_resnorm) / 4
    alpha, beta = 1 - evo_stage, evo_stage
    base_score = beta * (1 - norm_err) + alpha * (1 - complexity)
    base_score -= base_score.min()
    if base_score.sum() == 0:
        base_score[:] = 1
    base_probs = base_score / base_score.sum()

    # Novelty score based on residuals and predicted values diversity
    def novelty_score(mat):
        centered = mat - mat.mean(axis=1, keepdims=True)
        norms = np.linalg.norm(centered, axis=1, keepdims=True) + 1e-10
        normed = centered / norms
        sim = normed @ normed.T
        nov = 1 - sim.mean(axis=1)
        max_n = nov.max()
        return nov / (max_n if max_n > 0 else 1)

    novelty_res = novelty_score(residuals)
    novelty_pred = novelty_score(preds)
    novelty = 0.5 * (novelty_res + novelty_pred)

    # Dynamic novelty weighting: stronger early novelty, moderate late novelty
    novelty_weight = 0.35 + 0.3 * (1 - evo_stage)**0.8

    mixed_probs = (1 - novelty_weight) * base_probs + novelty_weight * novelty
    mixed_probs -= mixed_probs.min()
    if mixed_probs.sum() == 0:
        mixed_probs[:] = 1
    mixed_probs /= mixed_probs.sum()

    # Adaptive tournament size encourages pressure on error, preserves diversity
    tour_size = min(3 + int(evo_stage * 2), n)
    selected = []
    while len(selected) < k:
        chosen = np.random.choice(n, size=tour_size, replace=False, p=mixed_probs)
        # Tournament winner: lowest error considering base scores and novelty tie-break
        chosen_errs = errs[chosen]
        chosen_scores = base_probs[chosen]
        min_err_idx = chosen_errs.argmin()
        # Tie-break with highest combined score
        best_candidates = np.flatnonzero(chosen_errs == chosen_errs[min_err_idx])
        if len(best_candidates) > 1:
            best_scores = chosen_scores[best_candidates] + novelty[chosen][best_candidates]
            winner_idx = best_candidates[np.argmax(best_scores)]
        else:
            winner_idx = min_err_idx
        selected.append(population[chosen[winner_idx]])
    return selected
        \end{lstlisting}
        \FloatBarrier

        \subsubsection{Benchmark Analysis}
\begin{figure}[htb]
            \begin{minipage}[t]{0.48\columnwidth}
                \centering
                \includegraphics[width=\columnwidth]{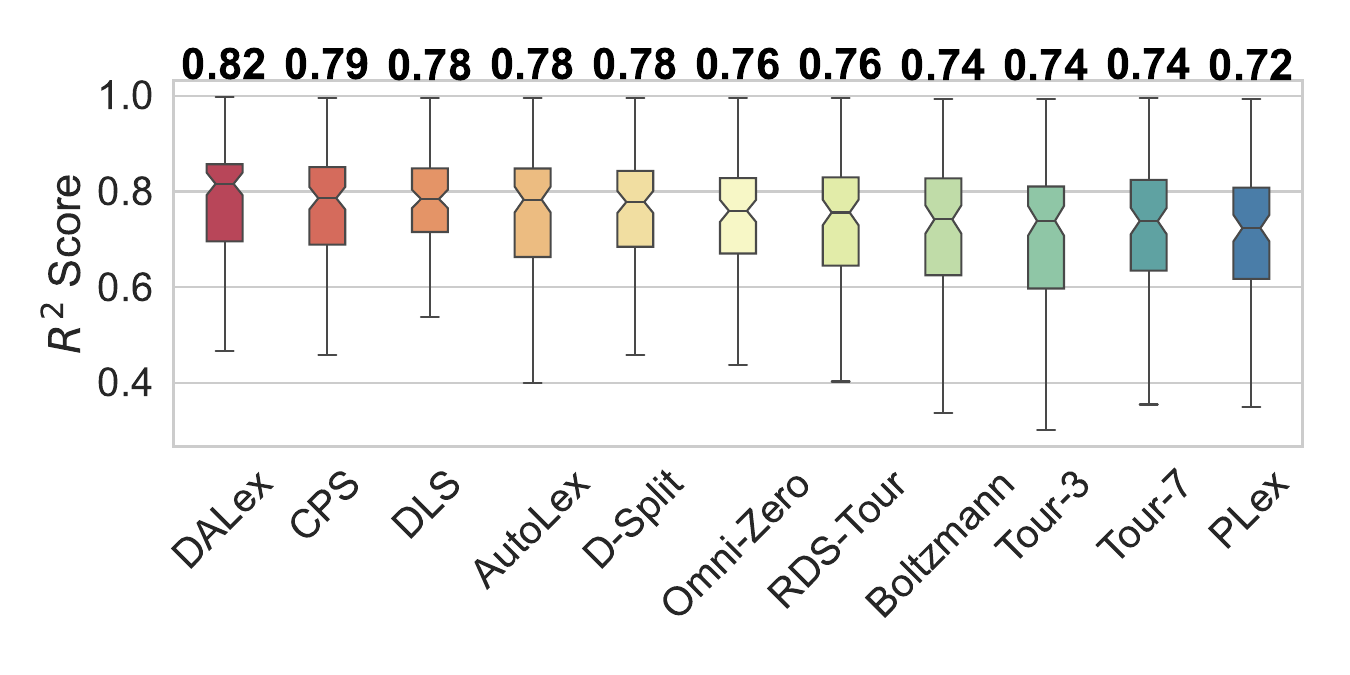}
                \caption{Test $R^2$ scores of different selection operators on symbolic regression benchmarks, including Omni-Zero.}
                \label{fig: r2 zero}
            \end{minipage}
            \hfill
            \begin{minipage}[t]{0.48\columnwidth}
                \centering
                \includegraphics[width=0.9\columnwidth]{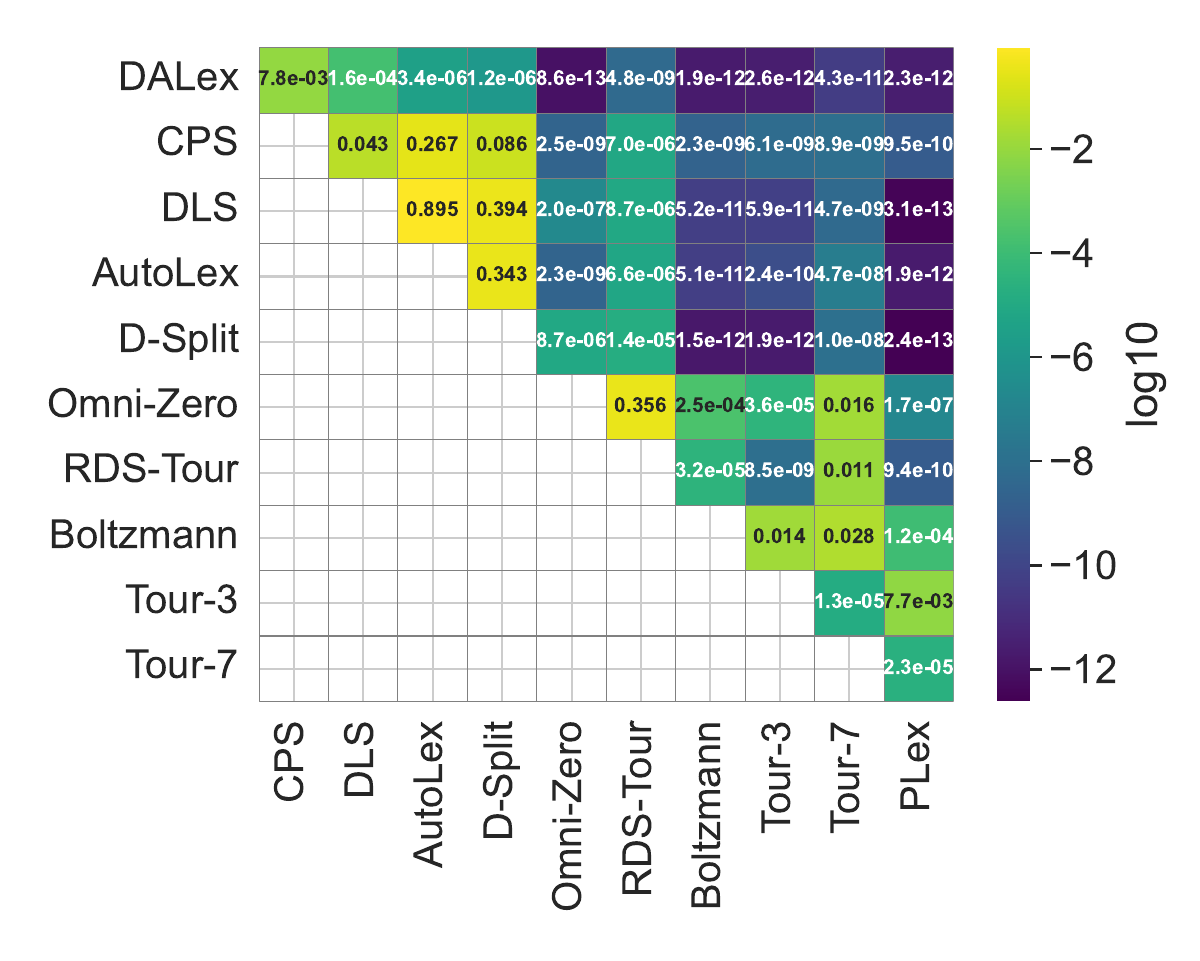}
                \caption{Pairwise statistical comparison of selection operators using the Wilcoxon signed-rank test with Benjamini--Hochberg correction.}
                \label{fig: p-value zero}
            \end{minipage}

        \end{figure}

        The Omni-Zero operator is evaluated on the SRBench datasets~\cite{la2021contemporary}, and the results are shown in \Cref{fig: r2 zero}. The proposed method achieves moderate performance among the benchmarked algorithms. It performs significantly better than tournament selection, as indicated by the p-values in \Cref{fig: p-value zero}. However, compared to selection operators that consider specificity, such as lexicase selection, the evolved selection operator is inferior.

        \begin{figure}[htb]
            \begin{minipage}[t]{0.48\columnwidth}
                \centering
                \includegraphics[width=\columnwidth]{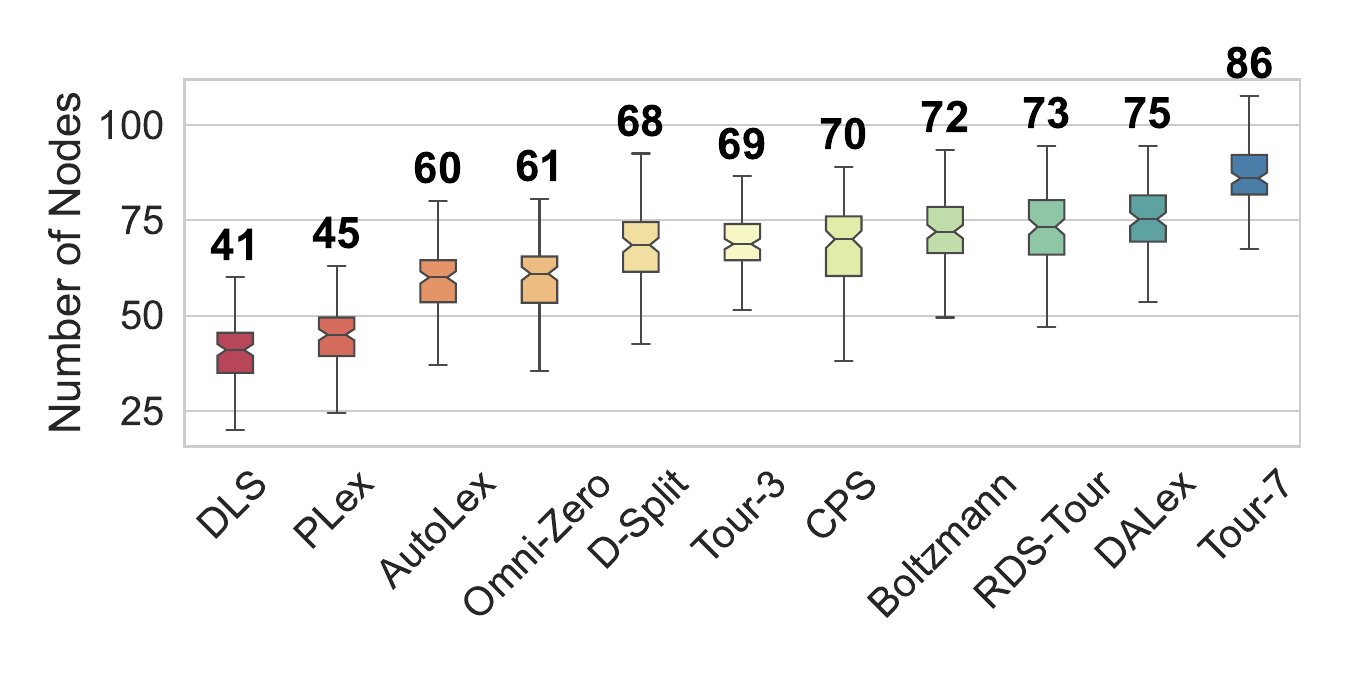}
                \caption{Tree sizes of different selection operators on SR benchmarks, including Omni-Zero.}
                \label{fig: complexity zero}
            \end{minipage}
            \hfill
            \begin{minipage}[t]{0.48\columnwidth}
                \centering
                \includegraphics[width=\columnwidth]{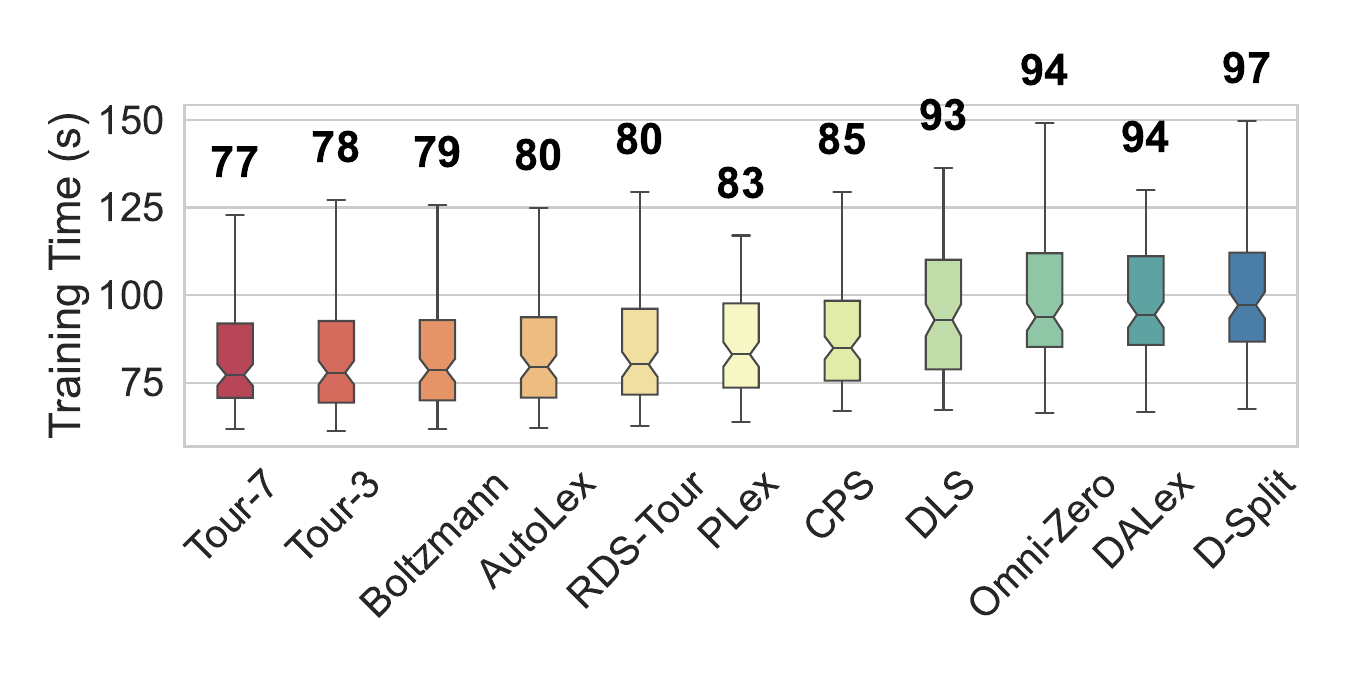}
                \caption{Training times of different selection operators on SR benchmarks, including Omni-Zero.}
                \label{fig: training_time zero}
            \end{minipage}
        \end{figure}

        \subsection{A Repaired Version of Omni Selection for Small Datasets}
        \label{sec:omni_small_dataset_repair}

        \subsubsection{Problem Analysis}
        Upon closely analyzing the code generated by the LLM in \Cref{lst:omni-selection}, we identified a logical bug that can lead to poor performance of the selection operator on small datasets. When the dataset is small, the structured division in line 10 of \Cref{lst:omni-selection} does not yield enough samples to form sufficient subsets, which can result in some subsets being empty. These empty subsets cause the subsequent selection process to rely solely on complexity when selecting individuals, without considering their predictive performance. This issue arises because the datasets used during meta-evolution generally contain a relatively large number of training instances, as shown in \Cref{tab:datasets}. Consequently, the bug has limited impact during training but becomes problematic when applied to smaller datasets.

        \begin{table}[!t]
            \centering
            \caption{Summary of datasets used in meta-evolution.}
            \label{tab:datasets}
            \begin{tabular}{lrr}
                \toprule
                \textbf{Dataset} & \textbf{n\_observations} & \textbf{n\_features} \\
                \midrule
                OPENML\_505      & 240                      & 124                  \\
                OPENML\_4544     & 1059                     & 117                  \\
                OPENML\_588      & 1000                     & 100                  \\
                OPENML\_650      & 500                      & 50                   \\
                \bottomrule
            \end{tabular}
        \end{table}

        \subsubsection{Solution}
        To remedy this limitation, we introduce a corrected variant of the Omni selection operator, termed Omni-R. The only modification is in the subset construction step: empty structured subsets are excluded, and random subsets are added to ensure exactly $K$ subsets in total. The modified subset construction is presented in \Cref{alg:omni-selection-repaired}; the remainder of the procedure is unchanged from \Cref{alg:omni-selection}. The complete Python implementation is provided in \Cref{lst:omni-selection-repaired}.

        \begin{algorithm}[!t]
        \caption{Omni-R: Differing Subset Construction}
        \label{alg:omni-selection-repaired}
        \begin{algorithmic}[1]
            \STATE $K \gets \lambda / 2$, \quad $m \gets \max(7, \lfloor |\mathbf{y}| / (2K) \rfloor)$
            \STATE $\mathcal{S}^{\text{seq}} \gets \emptyset$
            \FOR{$j = 1, \dots, K/2$}
            \STATE $S \gets \text{indices } [(j-1) \cdot m : \min(j \cdot m,\, |\mathbf{y}|)]$
            \STATE \textbf{if} $|S| > 0$ \textbf{then} $\mathcal{S}^{\text{seq}} \gets \mathcal{S}^{\text{seq}} \cup \{S\}$
            \ENDFOR
            \STATE $\mathcal{S}^{\text{rnd}} \gets \emptyset$
            \FOR{$j = 1, \dots, K - |\mathcal{S}^{\text{seq}}|$}
            \STATE $\mathcal{S}^{\text{rnd}} \gets \mathcal{S}^{\text{rnd}} \cup \{\text{RandomSample}(\{0,\dots,|\mathbf{y}|\!-\!1\},\; m)\}$
            \ENDFOR
            \STATE $\mathcal{S} \gets \mathcal{S}^{\text{seq}} \cup \mathcal{S}^{\text{rnd}}$
            \STATE \textbf{// Remainder same as \Cref{alg:omni-selection}}
        \end{algorithmic}
        \end{algorithm}

        \subsubsection{Benchmark Analysis}
\begin{figure}[!tb]
            \begin{minipage}[t]{0.48\columnwidth}
                \centering
                \includegraphics[width=\columnwidth]{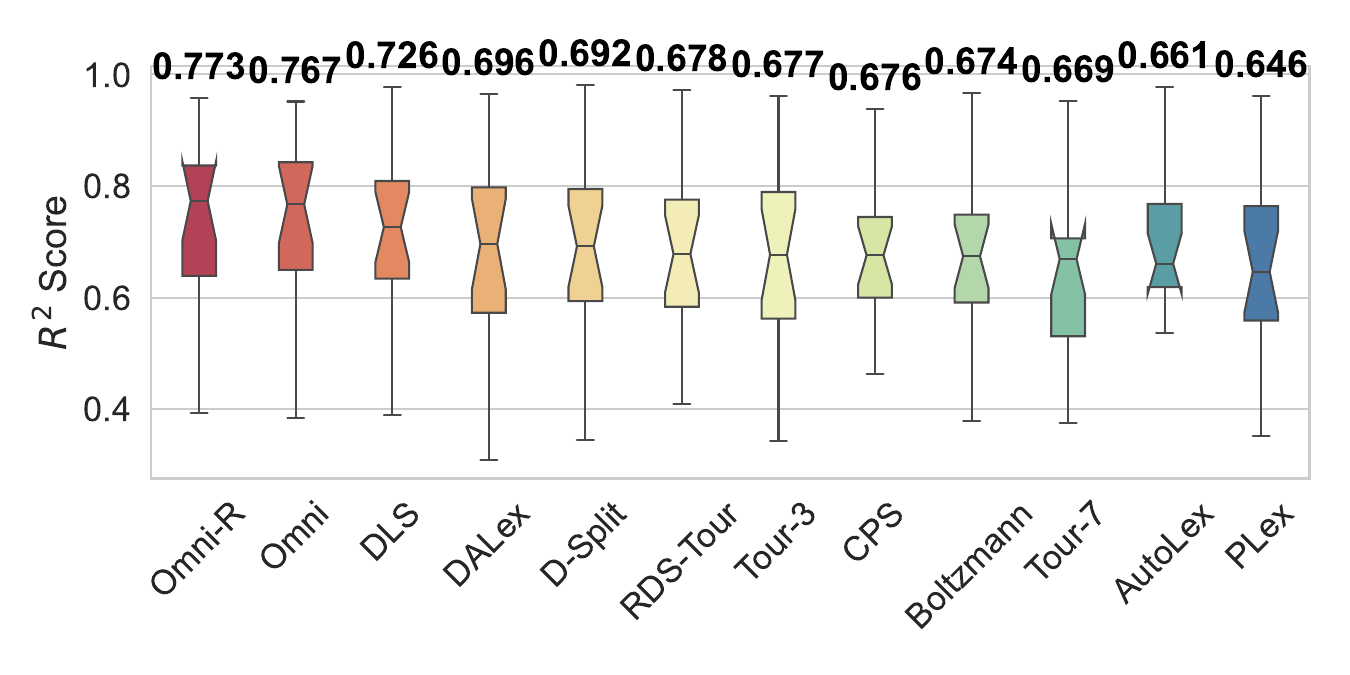}
                \caption{Test $R^2$ scores of different selection operators on small-scale SR benchmarks, including Omni-R.}
                \label{fig: r2 repaired}
            \end{minipage}
            \hfill
            \begin{minipage}[t]{0.48\columnwidth}
                \centering
                \includegraphics[width=\columnwidth]{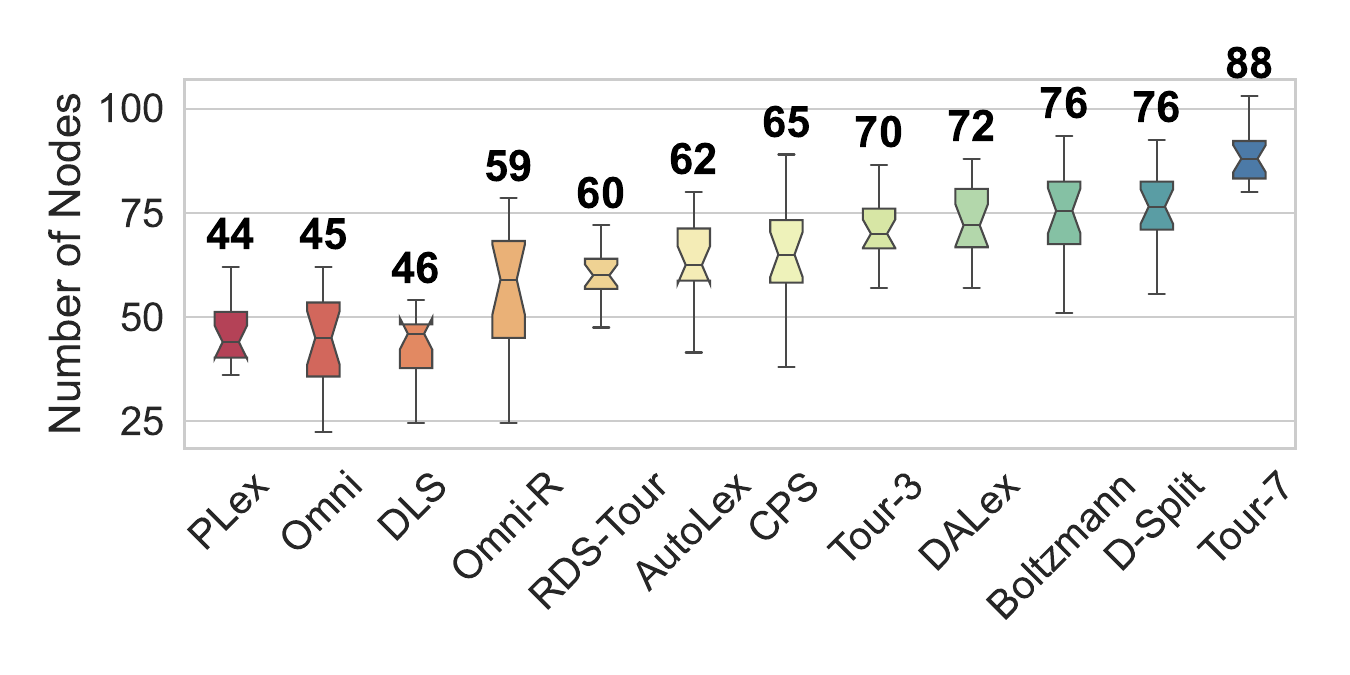}
                \caption{Tree sizes of different selection operators on small-scale SR benchmarks, including Omni-R.}
                \label{fig: complexity repaired}
            \end{minipage}
        \end{figure}

        We compare the performance of the repaired and original versions on datasets with no more than 100 training instances. The results, presented in \Cref{fig: r2 repaired}, show that the repaired version outperforms the original, indicating that the logical bug in the original version is indeed problematic. Regarding model size, as shown in \Cref{fig: complexity repaired}, the repaired version leads to larger trees, which is reasonable because it corrects the previous behavior of selecting individuals solely based on complexity by also considering the error on subsets. Nonetheless, the increase in tree size is modest, and the resulting models remain competitive in size compared to those produced by other selection operators.

        This failure highlights that, when using LLMs for meta-evolution, exposing the model to a wide range of instances is important. Otherwise, the generated code may contain subtle logical bugs that are not easily detected.

        \begin{lstlisting}[caption={The Repaired Version of Omni Selection (Omni-R)}, label={lst:omni-selection-repaired}]
def omni_selection(population, k=100, status={}):
    stage = np.clip(status.get("evolutionary_stage", 0), 0, 1)
    n = len(population)
    half_k = k // 2
    y = population[0].y
    n_cases = y.size

    ssize = max(7, n_cases // max(1, 2 * half_k))
    half_struct = half_k // 2
    structured = [
        np.arange(i * ssize, min((i + 1) * ssize, n_cases)) for i in range(half_struct)
    ]
    # Filter out empty subsets
    structured = [s for s in structured if s.size > 0]
    random_ = [
        np.random.choice(n_cases, ssize, replace=False)
        for _ in range(half_k - len(structured))
    ]
    subsets = structured + random_

    residuals = np.vstack([ind.y - ind.predicted_values for ind in population])
    complexity = np.array([len(ind) + ind.height for ind in population], float)
    complexity /= max(1, complexity.max())
    subset_mse = np.array(
        [
            [((residuals[i, s]) ** 2).mean() if s.size else np.inf for s in subsets]
            for i in range(n)
        ]
    )
    comp_factor = 0.25 + 0.25 * stage

    parent_a = [
        population[np.lexsort((complexity, subset_mse[:, i]))[0]]
        for i in range(len(subsets))
    ]

    idx_map = {ind: i for i, ind in enumerate(population)}
    norms = np.linalg.norm(residuals, axis=1) + 1e-12

    parent_b = []
    for a in parent_a:
        i_a = idx_map[a]
        res_a = residuals[i_a]
        cors = (residuals @ res_a) / (norms * norms[i_a])
        cors[i_a] = 1
        comp_score = np.abs(cors) + comp_factor * complexity
        b_idx = np.argmin(comp_score)
        parent_b.append(population[b_idx])

    return [ind for pair in zip(parent_a, parent_b) for ind in pair][:k]
        \end{lstlisting}

        \begin{lstlisting}[caption={Offspring Code}, label={lst:offspring}]
def omni_selection(population, k=100, status={}):
    stage = status.get("evolutionary_stage", 0)
    n = len(population)
    half_k = k // 2
    rng = np.random.default_rng(12345)

    errs = np.array([ind.case_values for ind in population])           # (n,n_cases)
    residuals = np.array([ind.y - ind.predicted_values for ind in population])
    sizes = np.array([len(ind) for ind in population])
    heights = np.array([ind.height for ind in population])

    n_cases = errs.shape[1]
    comp_pen = (sizes + heights) * (0.4 + 0.6 * stage) / (40 + 60 * stage)  # adaptive comp penalty

    subset_size = max(8, n_cases // (half_k + 2))
    max_tries = 15 * half_k
    tried_subsets = set()
    full_mse = errs.mean(axis=1)

    parent_a = []
    tries = 0
    while len(parent_a) < half_k and tries < max_tries:
        tries += 1
        subset = tuple(sorted(rng.choice(n_cases, subset_size, replace=False)))
        if subset in tried_subsets:
            continue
        tried_subsets.add(subset)

        mse_sub = errs[:, subset].mean(axis=1)
        specialization = full_mse - mse_sub  # + means better on subset specialization
        scores = (
            (1 - stage) * specialization +
            stage * (1 / (full_mse + 1e-10))
        ) - comp_pen
        best_idx = np.argmax(scores)
        parent_a.append(population[best_idx])

    if not parent_a:
        parent_a = rng.choice(population, half_k, replace=True).tolist()

    norm_resid = residuals / (np.linalg.norm(residuals, axis=1, keepdims=True) + 1e-10)
    parent_b = []
    comp_factor = 5 if stage < 0.5 else 10  # stronger complexity pressure late
    for a in parent_a:
        a_idx = population.index(a)
        corr = norm_resid @ norm_resid[a_idx]
        corr[a_idx] = 1  # exclude self
        scores = corr + comp_pen / comp_factor
        comp_idx = np.argmin(scores)
        parent_b.append(population[comp_idx])

    selected = [ind for pair in zip(parent_a, parent_b) for ind in pair]
    return selected[:k]
        \end{lstlisting}

        \begin{lstlisting}[caption={Code of Parent A}, label={lst:parent a}]
def omni_selection(population, k=100, status={}):
    stage = status.get("evolutionary_stage", 0)
    n = len(population)
    half_k = k // 2

    errs = np.array([ind.case_values for ind in population])  # (n, n_cases)
    residuals = np.array([ind.y - ind.predicted_values for ind in population])
    sizes = np.array([len(ind) for ind in population])
    heights = np.array([ind.height for ind in population])
    n_cases = errs.shape[1]
    rng = np.random.default_rng(777)

    # Complexity penalty grows with stage, stronger late to favor interpretability
    comp_pen = (sizes + heights) * (0.3 + 0.7 * stage) / 100

    # Parameters for subset selection
    subset_size = max(7, n_cases // (half_k + 3))
    max_tries = 10 * half_k
    tried_subsets = set()
    full_mse = errs.mean(axis=1)

    parent_a = []
    while len(parent_a) < half_k and max_tries > 0:
        max_tries -= 1
        subset = tuple(sorted(rng.choice(n_cases, subset_size, replace=False)))
        if subset in tried_subsets:
            continue
        tried_subsets.add(subset)

        mse_sub = errs[:, subset].mean(axis=1)
        specialization = full_mse - mse_sub  # positive means better on subset
        # Blend specialization and overall fitness with stage-dependent weights
        scores = (1 - stage) * specialization + stage * (1 / (full_mse + 1e-10))
        scores -= comp_pen
        best_idx = np.argmax(scores)
        parent_a.append(population[best_idx])

    if not parent_a:
        parent_a = rng.choice(population, half_k, replace=True).tolist()

    norm_resid = residuals / (np.linalg.norm(residuals, axis=1, keepdims=True) + 1e-10)
    parent_b = []
    for a in parent_a:
        a_idx = population.index(a)
        corr = norm_resid @ norm_resid[a_idx]
        corr[a_idx] = 1
        # Score complementary: minimize correlation and complexity; complexity effect softer early on
        scores = corr + comp_pen / (1 + stage * 5)
        b_idx = np.argmin(scores)
        parent_b.append(population[b_idx])

    selected = [ind for pair in zip(parent_a, parent_b) for ind in pair]
    return selected[:k]
        \end{lstlisting}

        \begin{lstlisting}[caption={Code of Parent B}, label={lst:parent b}]
def omni_selection(population, k=100, status={}):
    n = len(population)
    half_k = k // 2
    stage = status.get("evolutionary_stage", 0)
    max_tries = 10 * k

    preds = np.array([ind.predicted_values for ind in population])
    cases = np.array([ind.case_values for ind in population])
    residuals = np.array([ind.y - ind.predicted_values for ind in population])
    sizes = np.array([len(ind) for ind in population])
    heights = np.array([ind.height for ind in population])

    n_cases = preds.shape[1]
    subset_size = max(10, n_cases // (half_k + 1))
    idx_pool = np.arange(n_cases)
    np.random.seed(42)

    # Complexity penalty scaled by stage
    comp_pen = (sizes + heights) / (30 + 70 * stage)

    # Score function balances subset/full MSE and complexity
    def score(i, subset_idx):
        mse_subset = np.mean((cases[i, subset_idx] - preds[i, subset_idx]) ** 2)
        mse_full = np.mean((cases[i] - preds[i]) ** 2)
        return ((1 - stage) * (1 / (mse_subset + 1e-10)) + stage * (1 / (mse_full + 1e-10))) - comp_pen[i]

    parent_a = []
    tried_subsets = set()
    tries = 0
    while len(parent_a) < half_k and tries < max_tries:
        tries += 1
        subset_idx = tuple(sorted(np.random.choice(idx_pool, subset_size, replace=False)))
        if subset_idx in tried_subsets:
            continue
        tried_subsets.add(subset_idx)
        scores = np.array([score(i, subset_idx) for i in range(n)])
        best_idx = np.argmax(scores)
        parent_a.append(population[best_idx])

    if not parent_a:
        parent_a = np.random.choice(population, half_k, replace=True).tolist()

    parent_b = []
    norm_residuals = residuals / (np.linalg.norm(residuals, axis=1, keepdims=True) + 1e-10)
    for a in parent_a:
        a_idx = population.index(a)
        a_res = norm_residuals[a_idx]
        corr = norm_residuals @ a_res
        corr[a_idx] = 1.0  # exclude self
        complement_scores = corr + comp_pen / 10  # prefer low corr & low complexity
        comp_idx = np.argmin(complement_scores)
        parent_b.append(population[comp_idx])

    selected_individuals = [ind for pair in zip(parent_a, parent_b) for ind in pair]
    return selected_individuals[:k]
        \end{lstlisting}

        \subsection{Results on More Discovered Operators}
        \label{sec: More Operators}
\begin{figure*}[!t]
            \centering
            \begin{minipage}[t]{0.32\textwidth}
                \centering
                \includegraphics[width=\textwidth]{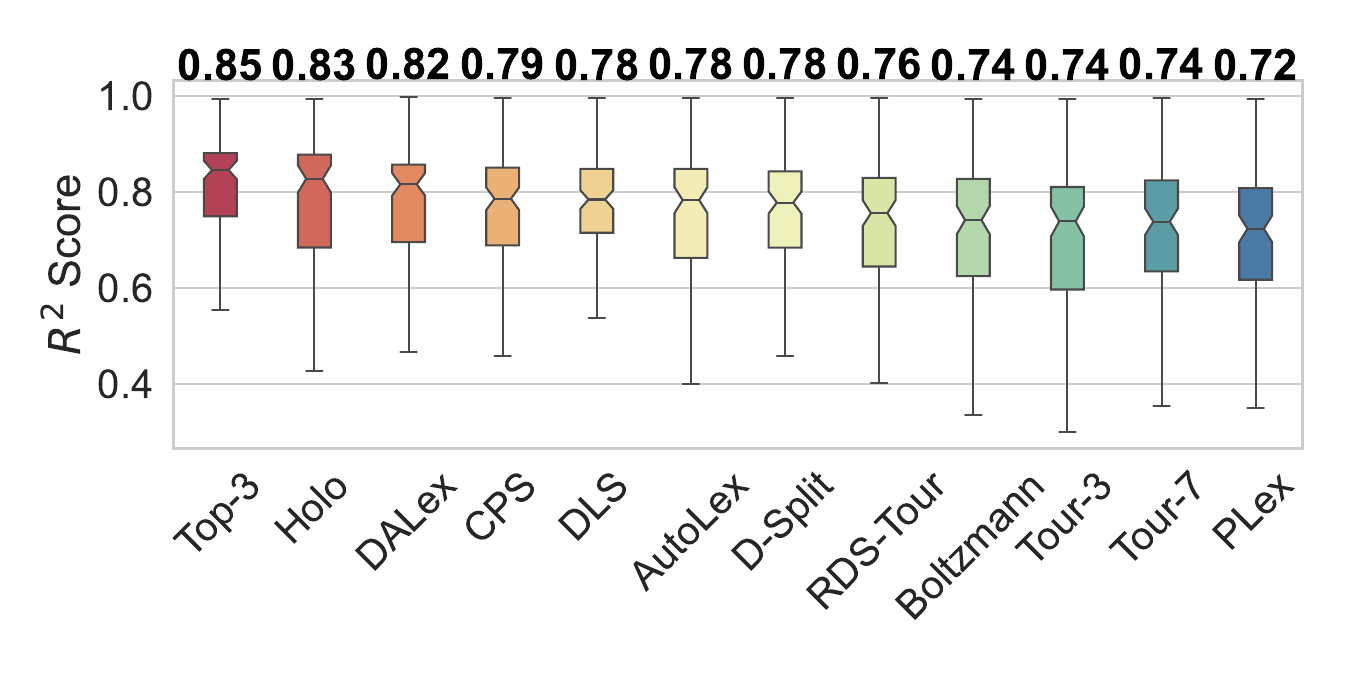}
                \caption{Test $R^2$ scores of different selection operators on SR benchmarks, including the Holo selection operator.}
                \label{fig: r2 v2}
            \end{minipage}
            \hfill
            \begin{minipage}[t]{0.32\textwidth}
                \centering
                \includegraphics[width=0.9\textwidth]{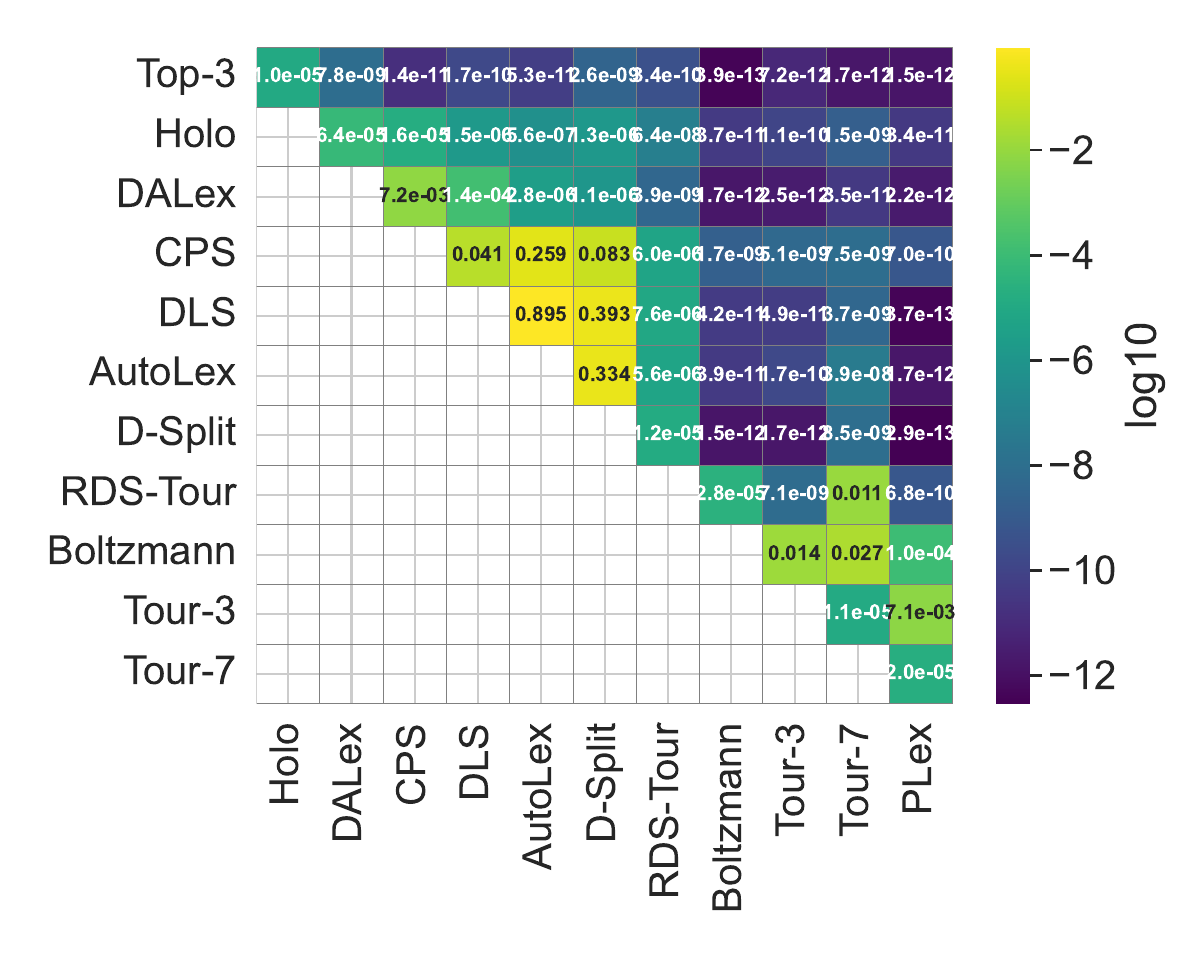}
                \caption{Pairwise statistical comparison of selection operators using the Wilcoxon signed-rank test with Benjamini--Hochberg correction.}
                \label{fig: p-value v2}
            \end{minipage}
            \hfill
            \begin{minipage}[t]{0.32\textwidth}
                \centering
                \includegraphics[width=\textwidth]{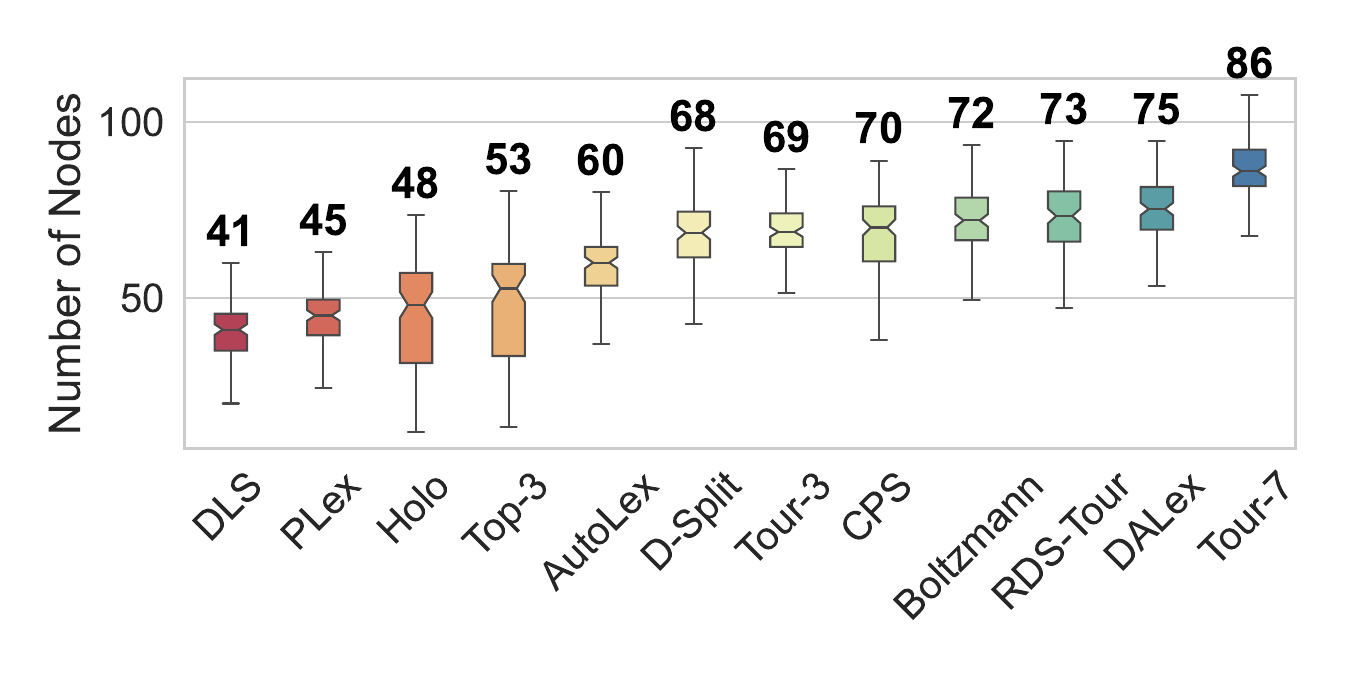}
                \caption{Tree sizes of different selection operators on SR benchmarks, including the Holo selection operator.}
                \label{fig: complexity v2}
            \end{minipage}
        \end{figure*}

        The proposed LLM-Meta-SR framework can discover a diverse range of selection operators beyond the Omni operator. \Cref{lst:offspring} presents another evolved operator, referred to as ``Holo'', which is the offspring operator introduced in \Cref{sec: Semantic-Aware Crossover}. As shown in \Cref{fig: r2 v2} and \Cref{fig: p-value v2}, the Holo operator achieves higher test $R^2$ scores than expert-designed operators, while maintaining competitive model complexity as illustrated in \Cref{fig: complexity v2}. Compared to the Omni operator reported in \Cref{fig: complexity}, Holo produces slightly smaller symbolic expressions, making it preferable when interpretability is crucial. In addition, the average test scores reported in \Cref{fig: r2 v2} show that the three best-performing algorithms also outperform expert-designed operators, indicating that the proposed LLM-Meta-SR framework can generate a wide range of high-quality operators.

        \section{More Analysis on SR Benchmark}
        \label{sec: More Analysis on SR Benchmark}
\begin{figure}[!tb]
            \centering
            \includegraphics[width=0.65\columnwidth]{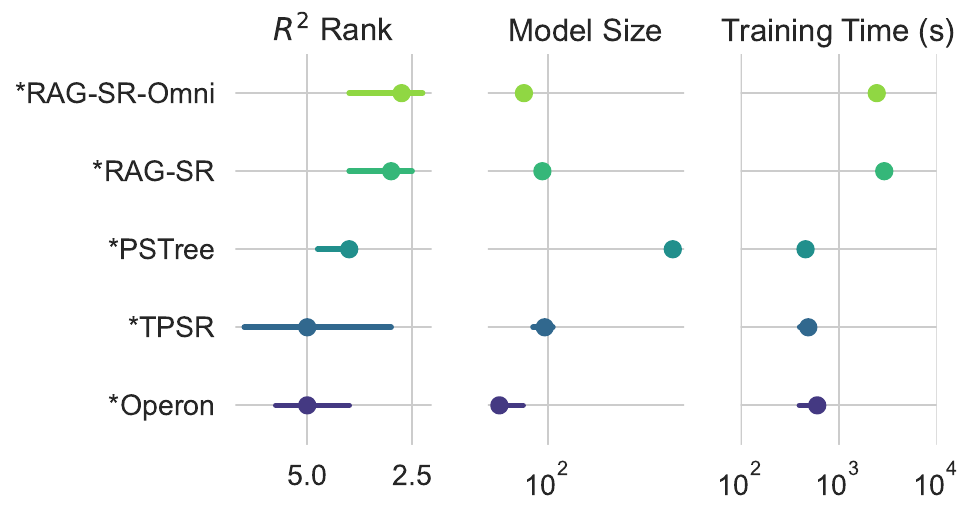}
            \caption{Median ranks of test $R^2$, model sizes, and training times for the top-5 algorithms on the SR benchmark.}
            \label{fig: SRBench Results Top}
        \end{figure}

\begin{figure}[!tb]
            \centering
            \includegraphics[width=0.65\columnwidth]{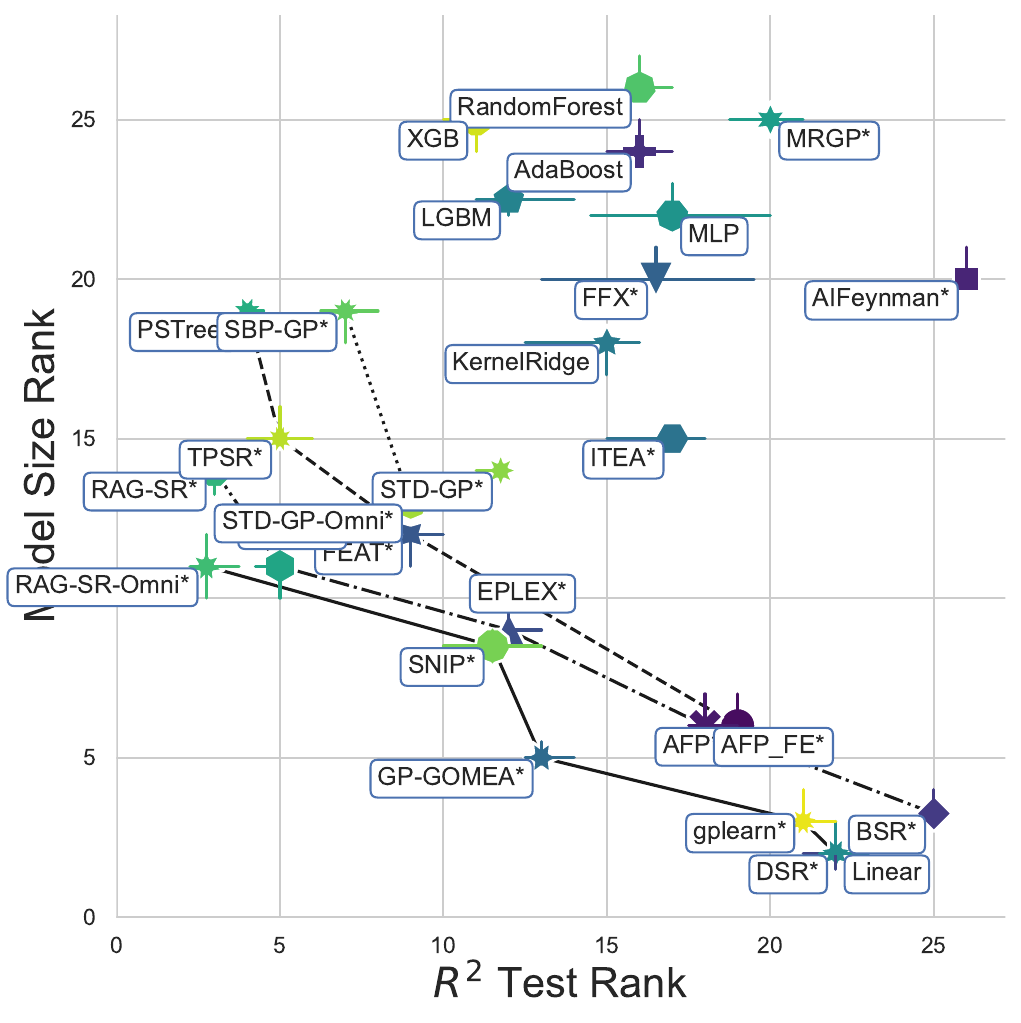}
            \caption{Pareto front of the ranks of test $R^2$ scores and model sizes for different algorithms.}
            \label{fig: Pareto Front}
        \end{figure}

        The results of the top-5 performing algorithms are shown in \Cref{fig: SRBench Results Top} for better clarity, and the Pareto front of the ranks of test $R^2$ scores and model sizes across different algorithms is shown in \Cref{fig: Pareto Front}. These results demonstrate that RAG-SR-Omni is Pareto-optimal among the compared algorithms on SRBench. Specifically, RAG-SR-Omni dominates retrieval-augmentation-generation-based SR (RAG-SR)~\cite{zhang2025ragsr} as well as transformer-based planning for SR~\cite{shojaee2024transformer}. Overall, these results show the effectiveness of the LLM-evolved selection operator.

        \section{Further Investigation of Crossover and Mutation Ratios}
        \label{sec: Crossover and Mutation Ratio}
\begin{figure}[!tb]
            \centering
            \begin{minipage}[t]{0.48\columnwidth}
                \centering
                \includegraphics[width=\columnwidth]{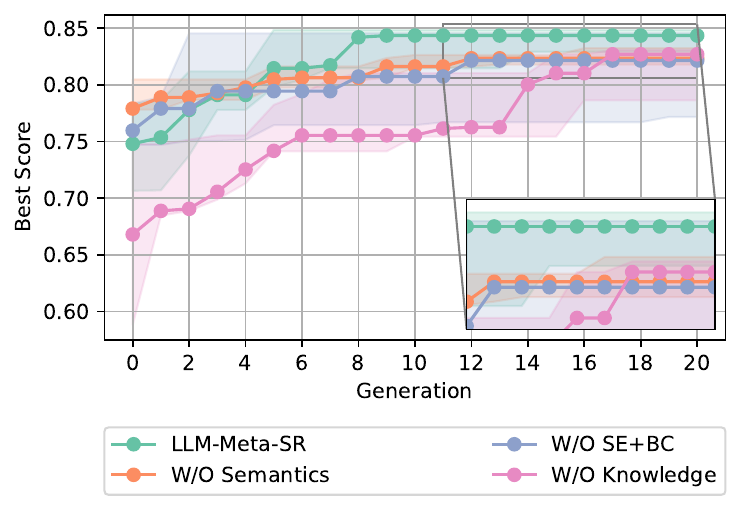}
                \caption{Evaluation $R^2$ of the best solution across generations for different LLM-driven search strategies in the case of a large mutation rate.}
                \label{fig:llm_evolution_5}
            \end{minipage}
            \hfill
            \begin{minipage}[t]{0.48\columnwidth}
                \centering
                \includegraphics[width=\columnwidth]{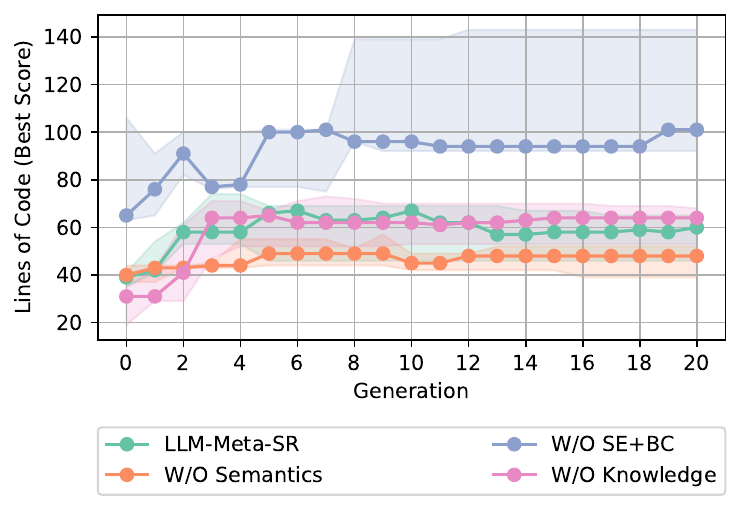}
                \caption{Code length of the best solution across generations for different LLM-driven search strategies in the case of a large mutation rate.}
                \label{fig:llm_code_length_evolution_5}
            \end{minipage}
        \end{figure}

        In the main paper, among the 20 offspring generated in each generation, 19 are produced by crossover and 1 by mutation. In this section, we investigate the proposed method under a different crossover-to-mutation ratio to examine how this ratio affects performance. Specifically, among the 20 offspring generated in each generation, 15 are produced by crossover and 5 by mutation. The results for evaluation score and code length are shown in \Cref{fig:llm_evolution_5} and \Cref{fig:llm_code_length_evolution_5}, respectively.

        First, the experimental results in \Cref{fig:llm_evolution_5} and \Cref{fig:llm_code_length_evolution_5} show that the conclusions from the ablation studies in \Cref{sec: Meta-Evolution Results} generalize to different crossover-to-mutation ratios. These conclusions include the effectiveness of semantics-based evolution, bloat control, and domain-knowledge-based operators.

        Second, the results in \Cref{fig:llm_evolution_5} as well as the numerical values in \Cref{tab:historical-best-5} indicate that the optimal crossover and mutation ratios vary depending on the prompting setting. When domain knowledge is available, a smaller mutation rate, as used in \Cref{sec: Meta-Evolution Results}, achieves slightly better performance than the larger mutation rate used in this section. In contrast, when domain knowledge is not available, a larger mutation rate appears to be more beneficial. One possible reason is that when domain knowledge is available, the search space is more confined and a large mutation rate does not provide additional benefit. Conversely, when domain knowledge is not available, the search space is larger and a higher mutation rate is needed to effectively explore the space.

        \begin{table*}[!tb]
            \centering
            \caption{Historical best scores and corresponding code lengths for each algorithm in the case of a large mutation rate.}
            \begin{tabular}{lcccccc}
                \toprule
                & LLM-Meta-SR & W/O Semantics & W/O SE+BC & W/O Knowledge \\
                \midrule
                Score         & 0.84        & 0.82          & 0.82      & 0.83          \\
                Lines of Code & 64          & 48            & 101       & 64            \\
                \bottomrule
            \end{tabular}%
            \label{tab:historical-best-5}

        \end{table*}

        \section{Analysis of Evaluation Time}
        \label{sec: evaluation_time}
\begin{figure}[!t]
            \centering
            \begin{subfigure}[t]{0.48\columnwidth}
                \centering
                \includegraphics[width=\textwidth, trim=5pt 5pt 5pt 5pt, clip]{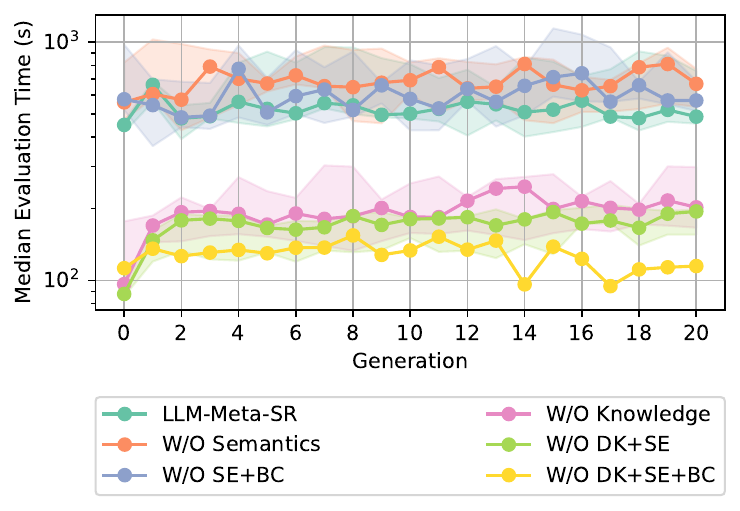}
                \caption{GPT-4.1-Mini ablation.}
                \label{fig:eval_time_gpt4}
            \end{subfigure}
            \hfill
            \begin{subfigure}[t]{0.48\columnwidth}
                \centering
                \includegraphics[width=\textwidth, trim=5pt 5pt 5pt 5pt, clip]{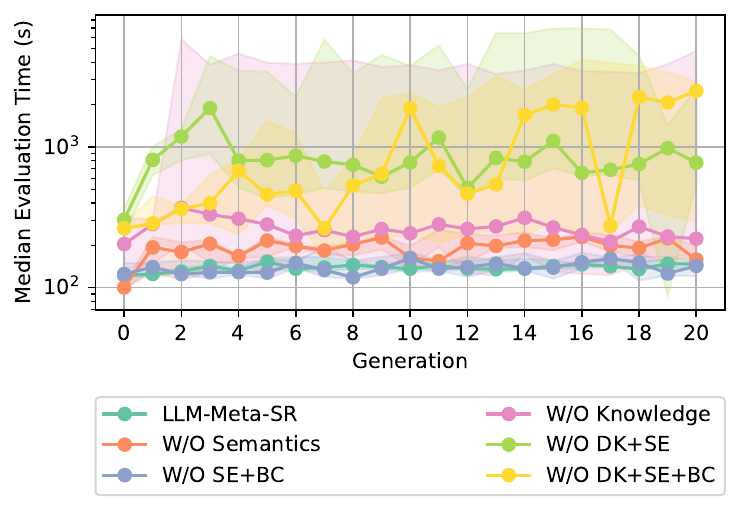}
                \caption{GPT-5-Mini ablation.}
                \label{fig:eval_time_gpt5}
            \end{subfigure}
            \caption{Average evaluation time (in seconds, log scale) per individual across generations for the GPT-4.1-Mini and GPT-5-Mini ablation studies. The shaded regions indicate 95\% bootstrap confidence intervals over three independent runs.}
            \label{fig:eval_time}
        \end{figure}

        In LLM-Meta-SR, the primary computational bottleneck is the evaluation of the evolved selection operators on SR benchmarks, rather than the LLM response generation, which is comparatively fast. To provide insight into the computational cost of the meta-evolution process, we report the average evaluation time per individual across generations for both the GPT-4.1-Mini and GPT-5-Mini ablation studies. As shown in \Cref{fig:eval_time}, each evaluation takes on the order of hundreds of seconds on average, remaining relatively stable across generations despite some variation. This also explains the need for the synthetic evaluation stage in \Cref{sec: Meta-Evolution}: before launching these expensive SR-based evaluations, the framework first filters out operators with syntax errors, runtime failures, or pathological computational behavior using a much cheaper proxy test.

        \par Notably, the choice of inductive bias significantly influences evaluation cost. For GPT-4.1-Mini, incorporating domain knowledge leads to longer evaluation times than the variants without it. A possible reason is that the operators automatically designed with knowledge guidance may exhibit more complex behaviors. For GPT-5-Mini, domain knowledge plays the opposite role: the variants with knowledge guidance yield evaluation times comparable to the GPT-4.1-Mini counterparts, whereas the variants without guidance are substantially more expensive, suggesting that domain knowledge helps constrain the behavioral complexity of the evolved operators. Moreover, without such a constraint, GPT-5-Mini produces operators with inherently higher complexity than GPT-4.1-Mini, indicating that a more capable model, when given full freedom, tends to explore a richer and more computationally demanding design space.

        \section{GPT-5 Prompt}
        \label{sec: GPT-5 Prompt}
        We made several improvements to the prompting strategy to better align the process with real-world algorithm development practices.
        \begin{itemize}
            \item Experts typically have a baseline operator for a given task, and providing this baseline is more consistent with real-world algorithm development practices. Therefore, we use DALex as the baseline implementation in the initialization prompt.
            \item In the code template, both the overall MSE and structural complexity are assigned small weights because they are already widely used in existing selection operators. This adjustment allows the algorithm to focus on designing the specialized selection and complementary components, which can lead to a novel selection operator.
            \item When GPT fails to generate a response containing executable code, the system re-invokes GPT instead of directly using the fallback operator. This avoids redundant evaluations of the fallback operator and reduces unnecessary computation.
        \end{itemize}

        \section{Prompt for Algorithm Evolution}
        \label{sec: Prompt}

        \textbf{System Prompt:} \Cref{fig: system prompt} shows the system prompt used for algorithm evolution, which is added before the specific prompt in all three phases of the automated algorithm design process, including initialization, crossover, and mutation. The system prompt is designed to guide the LLM in generating efficient selection operators by preferring NumPy vectorized operations and avoiding explicit Python for-loops unless absolutely necessary. Python is chosen for its simplicity and strong support from LLMs, but explicit Python for-loops are comparatively slow. In contrast, NumPy vectorized operations are implemented in C++ and offer better computational speed. The system prompt also specifies the maximum number of lines of code that the LLM can generate. This constraint helps control bloat, as discussed in \Cref{sec: Bloat Control}.

        \textbf{Initialization/Mutation Prompt:} \Cref{fig: llm_mutation} presents the prompts used for initialization and mutation. The main goal of these prompts is to encourage the LLM to generate a novel selection operator to explore the search space. The key difference between the prompts used for the two phases is that in the initialization phase, no baseline code is provided to the LLM. In the mutation phase, the elite solution is provided as the baseline code. The prompt format for wrapping the baseline code is shown in \Cref{fig: inspirational_prompt}, and it is intended to encourage the LLM to generate a novel selection operator based on the provided code.

        \textbf{Crossover Prompt:} \Cref{fig: llm_crossover} shows the crossover prompt used for algorithm evolution. The placeholder \texttt{goal} is specified as ``selection operator,'' and its plural form \texttt{goals} is specified as ``selection operators.'' The crossover prompt takes the code of two existing selection operators as input and aims to generate a better operator by combining effective building blocks from both. To support semantic awareness and concise code generation, the prompt includes the score vectors and the corresponding lines of code. The \textit{properties} describe the desired characteristics of the selection operator based on domain knowledge from \Cref{sec: Operator Design Principle}. The specific prompt is shown in \Cref{fig: domain knowledge}. The \textit{template} provides a function signature to ensure that the generated operator can be integrated into the automatic evaluation pipeline, as shown in \Cref{lst:template}. When domain knowledge is not provided, the \textit{properties} are left empty, and a simplified template for the selection operator without knowledge guidance, as shown in \Cref{lst:template without domain knowledge}, is used.

        \textbf{Fallback Selection Operator:} In rare cases where the LLM fails to generate valid Python code, a simple tournament selection operator, as shown in \Cref{lst: tournament}, is used as a default to complete the population.

        \begin{figure}[!ht]
            \centering
            \begin{tcolorbox}[
                colback=white,
                colframe=black,
                arc=4mm,
                boxrule=0.8pt,
                width=0.95\columnwidth,
            ]

                When writing code, prefer NumPy vectorized operations and avoid explicit Python for-loops unless absolutely necessary. Please implement code within \textcolor{varcolor}{\{max\_code\_lines\}} lines.

            \end{tcolorbox}

            \caption{System prompt for evolving selection operators.}
            \label{fig: system prompt}
        \end{figure}
        \begin{figure}[!ht]
            \centering
            \begin{tcolorbox}[
                colback=white,
                colframe=black,
                arc=4mm,
                boxrule=0.8pt,
                width=0.95\columnwidth,
            ]

                Your task is to develop an innovative and novel \textcolor{varcolor}{\{goal\}} for symbolic regression using genetic programming in Python.

                \vspace{1em}

                \textcolor{varcolor}{\{Baseline\}}\\
                \textcolor{varcolor}{\{Properties\}}

                \vspace{1em}

                Ensure that your newly designed function adheres to the following signature:\\
                \textcolor{varcolor}{\{Template\}}

                \vspace{1em}

                You do not need to provide a usage example.

                \vspace{1em}

                Embrace creativity, novelty, and bold experimentation to push the boundaries of the state of the art in \textcolor{varcolor}{\{goals\}} for genetic programming.

            \end{tcolorbox}

            \caption{Initialization and mutation prompts for designing an LLM-based selection operator. For initialization, the baseline is empty.}

            \label{fig: llm_mutation}
        \end{figure}

        \begin{figure}[!ht]
            \centering
            \begin{tcolorbox}[
                colback=white,
                colframe=black,
                arc=4mm,
                boxrule=0.8pt,
                width=0.95\columnwidth,
            ]

                Inspirational Example:

                \vspace{1em}

                \textcolor{varcolor}{\{Code\}}

                \vspace{1em}

                Use this as inspiration to create a distinctly original and inventive \textcolor{varcolor}{\{goal\}}.

            \end{tcolorbox}

            \caption{Prompt embedded in the baseline.}
            \label{fig: inspirational_prompt}
        \end{figure}

        \begin{figure}[!ht]
            \centering
            \begin{tcolorbox}[
                colback=white,
                colframe=black,
                arc=4mm,
                boxrule=0.8pt,
                width=0.95\columnwidth,
            ]
                You are tasked with designing a novel \textcolor{varcolor}{\{goal\}} for symbolic regression using genetic programming.

                \vspace{1em}

                Your goal is to synthesize a new operator that combines the strengths and mitigates the weaknesses of the two operators shown below:

                \vspace{1em}

                --- Operator A ---\\
                Score (Higher is Better): \textcolor{varcolor}{\{RoundToDecimalPlaces(Score\_A, 3)\}}, Lines of Code: \textcolor{varcolor}{\{EffectiveLineCount(OperatorA)\}}\\
                Code:\\
                \textcolor{varcolor}{\{OperatorA\_Code\}}

                \vspace{1em}

                --- Operator B ---\\
                Score (Higher is Better): \textcolor{varcolor}{\{RoundToDecimalPlaces(Score\_B, 3)\}}, Lines of Code: \textcolor{varcolor}{\{EffectiveLineCount(OperatorB)\}}\\
                Code:\\
                \textcolor{varcolor}{\{OperatorB\_Code\}}

                \vspace{1em}

                \textcolor{varcolor}{\{Properties\}}

                \vspace{1em}

                Ensure that your newly designed function adheres to the following signature:\\
                \textcolor{varcolor}{\{Template\}}

                \vspace{1em}

                You do not need to provide a usage example.
            \end{tcolorbox}

            \caption{Prompt used for crossover to design a novel selection operator with an LLM.}

            \label{fig: llm_crossover}
        \end{figure}

        \begin{figure}[!ht]
            \centering
            \begin{tcolorbox}[
                colback=white,
                colframe=black,
                arc=4mm,
                boxrule=0.6pt,
                width=0.95\columnwidth
            ]

                \textbf{1. Diverse \& Specialized Selection}\\
                \hspace{1.5em}-- Choose a varied set of high-performing individuals.\\
                \hspace{1.5em}-- Encourage specialization to maintain a diverse population.

                \vspace{0.5em}
                \textbf{2. Crossover-Aware Pairing}\\
                \hspace{1.5em}-- Promote complementarity between parents.

                \vspace{0.5em}
                \textbf{3. Stage-Specific Pressure}\\
                \hspace{1.5em}-- Vary selection pressure based on the current stage of evolution.

                \vspace{0.5em}
                \textbf{4. Interpretability}\\
                \hspace{1.5em}-- Prefer individuals with fewer nodes and lower tree height to improve model interpretability.

                \vspace{0.5em}
                \textbf{5. Efficiency \& Scalability}\\
                \hspace{1.5em}-- Include clear stopping conditions to avoid infinite loops.

                \vspace{0.5em}
                \textbf{6. Code Simplicity}\\
                \hspace{1.5em}-- Favor clear, concise logic with minimal complexity.

            \end{tcolorbox}
            \caption{Desired properties of good selection operators based on domain knowledge.}
            \label{fig: domain knowledge}
        \end{figure}
        \FloatBarrier
        \begin{lstlisting}[caption={Template for selection operator evolution.}, label={lst:template}]
def selection(population, k=100, status={}):
    # Useful information about individuals:
    # squared_error_vector = individual.case_values
    # predicted_values = individual.predicted_values
    # residual = individual.y-individual.predicted_values
    # number_of_nodes = len(individual)
    # height = individual.height

    # Useful information about evolution:
    # status["evolutionary_stage"]: [0,1], where 0 is the first generation and 1 is the final generation

    parent_a = Select k//2 individuals based on performance over subsets of instances:
        - evaluate individuals on k//2 different random or structured subsets of the data
        - reward those that specialize or perform well on those subsets
        - may also consider overall fitness (e.g., full-dataset MSE)
        - optionally include structural complexity
        - an individual can be selected multiple times

    parent_b = For each parent_a, select a complementary individual:
        - low residual correlation with parent_a
        - may also consider complexity

    # Interleave parent_a and parent_b to form crossover pairs
    selected_individuals = [individual for pair in zip(parent_a, parent_b) for individual in pair]
    return selected_individuals
        \end{lstlisting}

        \begin{lstlisting}[caption={Template for selection operator evolution without domain knowledge.}, label={lst:template without domain knowledge}]
def selection(population, k=100, status={}):
    # Useful information about individuals:
    # squared_error_vector = individual.case_values
    # predicted_values = individual.predicted_values
    # residual = individual.y-individual.predicted_values
    # number_of_nodes = len(individual)
    # height = individual.height

    # Useful information about evolution:
    # status["evolutionary_stage"]: [0,1], where 0 is the first generation and 1 is the final generation

    # Implement selection logic here
    return selected_individuals
        \end{lstlisting}

        \begin{lstlisting}[caption={Default code used when no code can be extracted from the LLM response.}, label={lst: tournament}]
import numpy as np
import random

def selection(individuals, k=100, tour_size=3):
    # Tournament selection
    # Compute mean error for each individual
    mean_errors = [np.mean(ind.case_values) for ind in individuals]
    selected_individuals = []

    # Select `k` individuals
    for _ in range(k):
        # Randomly select competitors
        competitors = random.sample(range(len(individuals)), tour_size)
        # Select the one with the lowest error
        best_idx = min(competitors, key=lambda idx: mean_errors[idx])
        selected_individuals.append(individuals[best_idx])

    return selected_individuals
        \end{lstlisting}

        \section{Computing Infrastructure}
        \label{sec: infrastructure}
        All experiments were run on AMD Milan CPUs. The software packages used in our experiments are listed in \Cref{tab:software}.

        \begin{table}[!tbh]
            \centering
            \caption{Software used for experiments.}
            \label{tab:software}
            \begin{tabular}{lc}
                \toprule
                \textbf{Type} & \textbf{Name}                     \\
                \midrule
                Code          & DEAP~\cite{fortin2012deap}        \\
                Code          & CodeBLEU~\cite{ren2020codebleu}   \\
                Benchmark     & SRBench~\cite{la2021contemporary} \\
                \bottomrule
            \end{tabular}
        \end{table}

        \begin{table*}[!t]
            \centering
            \caption{Statistical results of test \( R^2 \) scores for different selection operators on SR benchmarks (Part I). (``+'', ``='', and ``-'' indicate that the corresponding method is statistically better than, statistically equivalent to, or worse than Omni selection according to the Wilcoxon signed-rank test with \( p = 0.05 \).)}
            \label{tab:stats1}
            \begin{tabular}{lllllllllllllll}
                \toprule
                & Omni  & DALex    & CPS      & DLS      & AutoLex  & D-Split  & RDS-Tour & Boltzmann & Tour-3   & Tour-7   & PLex     \\
                \midrule
                192 & 0.501 & 0.503  (=) & 0.463  (=) & 0.485  (=) & 0.537  (=) & 0.591  (=) & 0.557  (=) & 0.479  (=)  & 0.533  (=) & 0.376  (=) & 0.5  (=) \\
                195 & 0.82  & 0.835  (=) & 0.831  (=) & 0.855  (+) & 0.823  (=) & 0.834  (+) & 0.83  (+)  & 0.839  (=)  & 0.854  (+) & 0.849  (+) & 0.816  (=) \\
                197 & 0.965 & 0.955  (-) & 0.96  (=)  & 0.958  (-) & 0.962  (-) & 0.942  (-) & 0.95  (-)  & 0.942  (-)  & 0.938  (-) & 0.952  (-) & 0.943  (-) \\
                201 & 0.872 & 0.867  (=) & 0.874  (=) & 0.863  (=) & 0.856  (-) & 0.864  (=) & 0.865  (-) & 0.838  (-)  & 0.822  (-) & 0.862  (-) & 0.819  (-) \\
                207 & 0.819 & 0.828  (+) & 0.803  (=) & 0.837  (=) & 0.844  (+) & 0.842  (+) & 0.829  (=) & 0.84  (+)   & 0.851  (+) & 0.842  (+) & 0.853  (+) \\
                210 & 0.773 & 0.82  (=)  & 0.761  (=) & 0.802  (+) & 0.835  (+) & 0.821  (+) & 0.817  (=) & 0.825  (+)  & 0.785  (=) & 0.805  (=) & 0.819  (+) \\
                215 & 0.919 & 0.925  (=) & 0.915  (=) & 0.91  (-)  & 0.909  (-) & 0.916  (=) & 0.885  (-) & 0.868  (-)  & 0.887  (-) & 0.891  (-) & 0.877  (-) \\
                218 & 0.546 & 0.559  (=) & 0.54  (=)  & 0.537  (=) & 0.554  (=) & 0.552  (=) & 0.565  (+) & 0.56  (=)   & 0.56  (+)  & 0.566  (=) & 0.532  (=) \\
                225 & 0.624 & 0.651  (+) & 0.635  (+) & 0.627  (=) & 0.634  (+) & 0.642  (+) & 0.644  (+) & 0.645  (+)  & 0.634  (+) & 0.651  (+) & 0.616  (=) \\
                227 & 0.955 & 0.953  (=) & 0.959  (=) & 0.949  (-) & 0.948  (-) & 0.938  (-) & 0.94  (-)  & 0.933  (-)  & 0.939  (-) & 0.946  (-) & 0.944  (-) \\
                228 & 0.648 & 0.657  (=) & 0.629  (=) & 0.757  (+) & 0.652  (+) & 0.751  (+) & 0.713  (+) & 0.702  (+)  & 0.765  (+) & 0.697  (=) & 0.723  (+) \\
                229 & 0.855 & 0.848  (=) & 0.842  (-) & 0.84  (-)  & 0.862  (=) & 0.824  (-) & 0.817  (-) & 0.796  (-)  & 0.77  (-)  & 0.816  (-) & 0.756  (-) \\
                230 & 0.859 & 0.896  (=) & 0.896  (=) & 0.88  (=)  & 0.888  (=) & 0.887  (=) & 0.873  (=) & 0.853  (=)  & 0.875  (=) & 0.878  (=) & 0.867  (=) \\
                294 & 0.777 & 0.762  (=) & 0.773  (=) & 0.769  (-) & 0.763  (-) & 0.766  (=) & 0.771  (=) & 0.772  (-)  & 0.773  (=) & 0.77  (-) & 0.756  (-) \\
                344 & 0.986 & 0.986  (=) & 0.987  (+) & 0.994  (+) & 0.987  (=) & 0.987  (=) & 0.971  (=) & 0.98  (=)   & 0.968  (=) & 0.987  (=) & 0.975  (=) \\
                485 & 0.687 & 0.599  (=) & 0.536  (-) & 0.645  (=) & 0.617  (=) & 0.58  (-)  & 0.57  (-)  & 0.551  (=)  & 0.548  (=) & 0.542  (-) & 0.523  (=) \\
                503 & 0.767 & 0.778  (+) & 0.776  (+) & 0.759  (-) & 0.767  (=) & 0.768  (=) & 0.778  (+) & 0.767  (=)  & 0.773  (=) & 0.778  (+) & 0.749  (-) \\
                519 & 0.742 & 0.739  (-) & 0.74  (=)  & 0.74  (=)  & 0.739  (=) & 0.734  (=) & 0.733  (-) & 0.734  (=)  & 0.739  (=) & 0.729  (-) & 0.741  (=) \\
                522 & 0.24  & 0.232  (=) & 0.222  (=) & 0.247  (=) & 0.236  (=) & 0.225  (=) & 0.212  (=) & 0.222  (=)  & 0.216  (=) & 0.202  (=) & 0.243  (=) \\
                523 & 0.944 & 0.94  (=)  & 0.939  (=) & 0.945  (=) & 0.937  (=) & 0.944  (=) & 0.939  (=) & 0.939  (=)  & 0.939  (=) & 0.938  (=) & 0.942  (=) \\
                527 & 0.952 & 0.965  (=) & 0.963  (=) & 0.977  (+) & 0.978  (=) & 0.982  (=) & 0.972  (=) & 0.967  (=)  & 0.961  (=) & 0.952  (=) & 0.962  (=) \\
                529 & 0.786 & 0.788  (+) & 0.788  (=) & 0.788  (=) & 0.787  (=) & 0.788  (+) & 0.79  (=)  & 0.787  (=)  & 0.785  (=) & 0.786  (=) & 0.785  (=) \\
                537 & 0.639 & 0.622  (=) & 0.643  (=) & 0.605  (-) & 0.612  (-) & 0.598  (-) & 0.612  (-) & 0.597  (-)  & 0.587  (-) & 0.601  (-) & 0.578  (-) \\
                542 & 0.383 & 0.344  (=) & 0.224  (=) & 0.39  (+)  & 0.306  (=) & 0.345  (=) & 0.409  (+) & 0.289  (=)  & 0.343  (=) & 0.391  (=) & 0.383  (+) \\
                547 & 0.496 & 0.499  (=) & 0.506  (=) & 0.5  (=)   & 0.504  (=) & 0.518  (+) & 0.494  (=) & 0.495  (=)  & 0.475  (=) & 0.487  (=) & 0.477  (=) \\
                556 & 0.891 & 0.903  (+) & 0.892  (=) & 0.904  (=) & 0.887  (=) & 0.905  (=) & 0.887  (=) & 0.899  (=)  & 0.902  (+) & 0.911  (=) & 0.889  (=) \\
                557 & 0.9   & 0.904  (=) & 0.891  (=) & 0.896  (=) & 0.912  (=) & 0.914  (=) & 0.907  (=) & 0.88  (-)   & 0.904  (=) & 0.916  (=) & 0.899  (=) \\
                560 & 0.992 & 0.99  (=)  & 0.993  (=) & 0.991  (=) & 0.99  (=)  & 0.991  (=) & 0.992  (=) & 0.99  (=)   & 0.992  (=) & 0.991  (=) & 0.991  (=) \\
                561 & 0.979 & 0.972  (=) & 0.976  (=) & 0.954  (-) & 0.966  (-) & 0.951  (-) & 0.976  (=) & 0.954  (-)  & 0.951  (-) & 0.954  (-) & 0.949  (-) \\
                562 & 0.955 & 0.953  (=) & 0.959  (=) & 0.949  (-) & 0.948  (-) & 0.938  (-) & 0.94  (-)  & 0.933  (-)  & 0.939  (-) & 0.946  (-) & 0.944  (-) \\
                564 & 0.924 & 0.914  (=) & 0.921  (=) & 0.888  (-) & 0.894  (-) & 0.89  (-)  & 0.894  (-) & 0.892  (-)  & 0.888  (-) & 0.908  (-) & 0.839  (-) \\
                573 & 0.965 & 0.955  (-) & 0.96  (=)  & 0.958  (-) & 0.962  (-) & 0.942  (-) & 0.95  (-)  & 0.942  (-)  & 0.938  (-) & 0.952  (-) & 0.943  (-) \\
                574 & 0.334 & 0.316  (-) & 0.34  (=)  & 0.318  (-) & 0.319  (-) & 0.301  (-) & 0.349  (=) & 0.336  (=)  & 0.326  (-) & 0.347  (=) & 0.302  (-) \\
                579 & 0.9   & 0.903  (-) & 0.89  (=)  & 0.882  (-) & 0.881  (-) & 0.882  (-) & 0.87  (-)  & 0.874  (-)  & 0.869  (-) & 0.876  (-) & 0.825  (-) \\
                581 & 0.842 & 0.824  (=) & 0.793  (=) & 0.785  (-) & 0.775  (-) & 0.747  (-) & 0.733  (-) & 0.706  (-)  & 0.621  (-) & 0.695  (-) & 0.686  (-) \\
                582 & 0.855 & 0.822  (-) & 0.819  (-) & 0.775  (-) & 0.785  (-) & 0.782  (-) & 0.669  (-) & 0.668  (-)  & 0.698  (-) & 0.714  (-) & 0.65  (-) \\
                583 & 0.837 & 0.764  (-) & 0.806  (-) & 0.776  (-) & 0.75  (-)  & 0.756  (-) & 0.68  (-)  & 0.702  (-)  & 0.628  (-) & 0.717  (-) & 0.654  (-) \\
                584 & 0.848 & 0.803  (-) & 0.775  (-) & 0.784  (-) & 0.774  (-) & 0.769  (-) & 0.765  (-) & 0.684  (-)  & 0.662  (-) & 0.734  (-) & 0.679  (-) \\
                586 & 0.861 & 0.839  (-) & 0.828  (-) & 0.794  (-) & 0.817  (-) & 0.812  (-) & 0.712  (-) & 0.71  (-)   & 0.644  (-) & 0.727  (-) & 0.66  (-) \\
                589 & 0.835 & 0.82  (=)  & 0.763  (-) & 0.747  (-) & 0.769  (-) & 0.781  (-) & 0.74  (-)  & 0.742  (-)  & 0.727  (-) & 0.749  (-) & 0.701  (-) \\
                590 & 0.91  & 0.871  (-) & 0.889  (-) & 0.848  (-) & 0.855  (-) & 0.826  (-) & 0.808  (-) & 0.796  (-)  & 0.809  (-) & 0.848  (-) & 0.793  (-) \\
                591 & 0.78  & 0.725  (=) & 0.732  (-) & 0.726  (-) & 0.719  (=) & 0.687  (-) & 0.664  (-) & 0.688  (-)  & 0.677  (-) & 0.702  (-) & 0.646  (-) \\
                592 & 0.848 & 0.824  (-) & 0.784  (-) & 0.788  (-) & 0.813  (-) & 0.746  (-) & 0.696  (-) & 0.655  (-)  & 0.629  (-) & 0.667  (-) & 0.691  (-) \\
                593 & 0.88  & 0.869  (=) & 0.879  (=) & 0.827  (-) & 0.82  (-)  & 0.826  (-) & 0.788  (-) & 0.818  (-)  & 0.771  (-) & 0.808  (-) & 0.767  (-) \\
                594 & 0.725 & 0.696  (=) & 0.674  (=) & 0.676  (=) & 0.651  (=) & 0.675  (=) & 0.681  (=) & 0.674  (-)  & 0.677  (=) & 0.672  (=) & 0.64  (-) \\
                595 & 0.917 & 0.899  (-) & 0.911  (=) & 0.892  (-) & 0.886  (-) & 0.885  (-) & 0.892  (-) & 0.877  (-)  & 0.867  (-) & 0.87  (-) & 0.823  (-) \\
                596 & 0.86  & 0.83  (=)  & 0.79  (-)  & 0.846  (=) & 0.797  (-) & 0.821  (=) & 0.762  (-) & 0.769  (-)  & 0.745  (-) & 0.799  (-) & 0.724  (-) \\
                597 & 0.866 & 0.809  (=) & 0.822  (=) & 0.792  (-) & 0.812  (-) & 0.796  (-) & 0.767  (-) & 0.764  (-)  & 0.757  (-) & 0.785  (-) & 0.756  (-) \\
                598 & 0.908 & 0.894  (-) & 0.901  (=) & 0.872  (-) & 0.875  (-) & 0.885  (-) & 0.882  (-) & 0.842  (-)  & 0.814  (-) & 0.877  (-) & 0.805  (-) \\
                599 & 0.895 & 0.835  (-) & 0.829  (-) & 0.799  (-) & 0.801  (-) & 0.804  (-) & 0.808  (-) & 0.787  (-)  & 0.785  (-) & 0.797  (-) & 0.758  (-) \\
                601 & 0.879 & 0.819  (-) & 0.813  (-) & 0.816  (-) & 0.796  (-) & 0.824  (-) & 0.774  (-) & 0.763  (-)  & 0.773  (-) & 0.754  (-) & 0.758  (-) \\
                602 & 0.854 & 0.81  (=)  & 0.788  (-) & 0.758  (-) & 0.797  (-) & 0.787  (=) & 0.731  (-) & 0.647  (-)  & 0.545  (-) & 0.669  (-) & 0.683  (-) \\
                603 & 0.893 & 0.834  (-) & 0.827  (-) & 0.821  (-) & 0.772  (-) & 0.792  (-) & 0.771  (-) & 0.744  (-)  & 0.782  (-) & 0.758  (-) & 0.734  (-) \\
                604 & 0.858 & 0.846  (=) & 0.833  (-) & 0.804  (-) & 0.815  (-) & 0.837  (=) & 0.756  (-) & 0.771  (-)  & 0.708  (-) & 0.729  (-) & 0.734  (-) \\
                605 & 0.787 & 0.736  (-) & 0.715  (-) & 0.671  (-) & 0.664  (-) & 0.698  (-) & 0.67  (-)  & 0.687  (-)  & 0.672  (-) & 0.652  (-) & 0.652  (-) \\
                606 & 0.844 & 0.791  (-) & 0.782  (-) & 0.751  (-) & 0.778  (-) & 0.779  (-) & 0.752  (-) & 0.748  (-)  & 0.739  (-) & 0.737  (-) & 0.723  (-) \\
                607 & 0.853 & 0.772  (-) & 0.71  (-)  & 0.777  (-) & 0.763  (-) & 0.724  (-) & 0.686  (-) & 0.604  (-)  & 0.633  (-) & 0.693  (-) & 0.667  (-) \\
                608 & 0.88  & 0.858  (=) & 0.801  (-) & 0.797  (-) & 0.808  (-) & 0.813  (-) & 0.78  (-)  & 0.732  (-)  & 0.749  (-) & 0.752  (-) & 0.746  (-) \\\bottomrule
            \end{tabular}

        \end{table*}

        \begin{table*}[!t]
            \centering
            \caption{Statistical results of test \( R^2 \) scores for different selection operators on SR benchmarks (Part II). (``+'', ``='', and ``-'' indicate that the corresponding method is statistically better than, statistically equivalent to, or worse than Omni selection according to the Wilcoxon signed-rank test with \( p = 0.05 \).)}
            \label{tab:stats2}

            \resizebox{\linewidth}{!}{
                \begin{tabular}{lllllllllllllll}
                    \toprule
                    & Omni    & DALex    & CPS      & DLS      & AutoLex  & D-Split  & RDS-Tour & Boltzmann & Tour-3   & Tour-7   & PLex     \\
                    \midrule
                    609     & 0.929   & 0.911  (-) & 0.928  (=) & 0.903  (-) & 0.898  (-) & 0.907  (-) & 0.915  (-) & 0.894  (-)  & 0.894  (-) & 0.914  (-) & 0.862  (-) \\
                    611     & 0.848   & 0.76  (-)  & 0.74  (-)  & 0.781  (-) & 0.761  (-) & 0.761  (=) & 0.678  (-) & 0.736  (-)  & 0.693  (-) & 0.669  (-) & 0.685  (-) \\
                    612     & 0.886   & 0.825  (-) & 0.836  (-) & 0.798  (-) & 0.802  (-) & 0.813  (-) & 0.829  (-) & 0.8  (-)    & 0.782  (-) & 0.813  (-) & 0.724  (-) \\
                    613     & 0.861   & 0.838  (=) & 0.753  (-) & 0.81  (-)  & 0.803  (-) & 0.797  (=) & 0.79  (-)  & 0.711  (-)  & 0.736  (-) & 0.802  (-) & 0.73  (-) \\
                    615     & 0.813   & 0.785  (-) & 0.763  (-) & 0.76  (-)  & 0.774  (-) & 0.735  (-) & 0.676  (-) & 0.629  (-)  & 0.58  (-)  & 0.684  (-) & 0.693  (-) \\
                    616     & 0.825   & 0.695  (-) & 0.763  (=) & 0.74  (-)  & 0.75  (-)  & 0.712  (-) & 0.617  (-) & 0.611  (-)  & 0.577  (-) & 0.578  (-) & 0.612  (-) \\
                    617     & 0.87    & 0.818  (-) & 0.85  (-)  & 0.803  (-) & 0.786  (-) & 0.808  (-) & 0.794  (-) & 0.753  (-)  & 0.751  (-) & 0.783  (-) & 0.737  (-) \\
                    618     & 0.853   & 0.814  (=) & 0.726  (-) & 0.748  (-) & 0.771  (-) & 0.748  (-) & 0.645  (-) & 0.641  (-)  & 0.599  (-) & 0.616  (-) & 0.707  (-) \\
                    620     & 0.855   & 0.843  (=) & 0.822  (-) & 0.8  (-)   & 0.796  (-) & 0.772  (-) & 0.772  (-) & 0.738  (-)  & 0.701  (-) & 0.771  (-) & 0.666  (-) \\
                    621     & 0.85    & 0.834  (=) & 0.851  (=) & 0.85  (=)  & 0.841  (=) & 0.828  (=) & 0.808  (-) & 0.764  (-)  & 0.794  (-) & 0.799  (-) & 0.794  (-) \\
                    622     & 0.807   & 0.799  (=) & 0.751  (-) & 0.757  (-) & 0.753  (=) & 0.747  (-) & 0.712  (-) & 0.724  (-)  & 0.738  (-) & 0.742  (-) & 0.688  (-) \\
                    623     & 0.883   & 0.842  (-) & 0.851  (-) & 0.8  (-)   & 0.794  (-) & 0.812  (-) & 0.815  (-) & 0.782  (-)  & 0.769  (-) & 0.808  (-) & 0.733  (-) \\
                    624     & 0.884   & 0.828  (-) & 0.832  (-) & 0.837  (-) & 0.859  (-) & 0.857  (-) & 0.798  (-) & 0.823  (-)  & 0.791  (-) & 0.792  (-) & 0.818  (-) \\
                    626     & 0.801   & 0.763  (-) & 0.751  (-) & 0.732  (-) & 0.724  (-) & 0.72  (-)  & 0.673  (-) & 0.667  (-)  & 0.704  (-) & 0.733  (-) & 0.618  (-) \\
                    627     & 0.847   & 0.769  (-) & 0.765  (-) & 0.743  (-) & 0.774  (-) & 0.75  (-)  & 0.732  (-) & 0.74  (-)   & 0.754  (-) & 0.738  (-) & 0.711  (-) \\
                    628     & 0.879   & 0.86  (=)  & 0.851  (=) & 0.823  (-) & 0.827  (-) & 0.853  (-) & 0.791  (-) & 0.814  (-)  & 0.779  (-) & 0.805  (-) & 0.721  (-) \\
                    631     & 0.871   & 0.819  (-) & 0.831  (-) & 0.808  (-) & 0.818  (-) & 0.777  (-) & 0.783  (-) & 0.784  (-)  & 0.779  (-) & 0.799  (-) & 0.698  (-) \\
                    633     & 0.906   & 0.87  (-)  & 0.888  (-) & 0.885  (-) & 0.865  (-) & 0.87  (-)  & 0.856  (-) & 0.845  (-)  & 0.834  (-) & 0.859  (-) & 0.816  (-) \\
                    634     & 0.742   & 0.655  (-) & 0.676  (-) & 0.656  (-) & 0.661  (-) & 0.668  (-) & 0.59  (-)  & 0.648  (-)  & 0.601  (-) & 0.605  (-) & 0.595  (-) \\
                    635     & 0.905   & 0.865  (-) & 0.86  (-)  & 0.827  (-) & 0.87  (-)  & 0.846  (-) & 0.852  (-) & 0.82  (-)   & 0.814  (-) & 0.826  (-) & 0.767  (-) \\
                    637     & 0.813   & 0.783  (=) & 0.76  (=)  & 0.742  (-) & 0.781  (-) & 0.776  (-) & 0.698  (-) & 0.674  (-)  & 0.667  (-) & 0.72  (-) & 0.655  (-) \\
                    641     & 0.87    & 0.847  (-) & 0.788  (-) & 0.789  (-) & 0.789  (-) & 0.783  (-) & 0.764  (-) & 0.793  (-)  & 0.718  (-) & 0.77  (-) & 0.746  (-) \\
                    643     & 0.807   & 0.767  (=) & 0.753  (-) & 0.74  (-)  & 0.749  (-) & 0.753  (-) & 0.741  (-) & 0.696  (-)  & 0.722  (-) & 0.719  (-) & 0.686  (-) \\
                    644     & 0.844   & 0.776  (-) & 0.801  (-) & 0.775  (-) & 0.741  (-) & 0.737  (-) & 0.654  (-) & 0.575  (-)  & 0.511  (-) & 0.638  (-) & 0.711  (-) \\
                    645     & 0.692   & 0.767  (=) & 0.751  (=) & 0.746  (=) & 0.748  (=) & 0.677  (=) & 0.54  (-)  & 0.562  (-)  & 0.51  (-)  & 0.504  (-) & 0.562  (-) \\
                    646     & 0.863   & 0.853  (=) & 0.823  (-) & 0.821  (-) & 0.819  (-) & 0.828  (-) & 0.815  (-) & 0.834  (-)  & 0.726  (-) & 0.809  (-) & 0.735  (-) \\
                    647     & 0.859   & 0.825  (-) & 0.783  (-) & 0.827  (-) & 0.805  (-) & 0.838  (=) & 0.744  (-) & 0.756  (-)  & 0.728  (-) & 0.774  (-) & 0.736  (-) \\
                    648     & 0.862   & 0.799  (-) & 0.799  (-) & 0.779  (-) & 0.776  (-) & 0.694  (-) & 0.722  (-) & 0.688  (-)  & 0.611  (-) & 0.67  (-) & 0.611  (-) \\
                    649     & 0.926   & 0.909  (-) & 0.923  (=) & 0.901  (-) & 0.9  (-)   & 0.899  (-) & 0.893  (-) & 0.903  (-)  & 0.885  (-) & 0.911  (-) & 0.867  (-) \\
                    651     & 0.837   & 0.735  (-) & 0.669  (-) & 0.771  (-) & 0.718  (-) & 0.692  (-) & 0.711  (-) & 0.653  (-)  & 0.641  (-) & 0.659  (-) & 0.644  (-) \\
                    653     & 0.872   & 0.824  (-) & 0.823  (-) & 0.82  (-)  & 0.818  (-) & 0.825  (-) & 0.813  (-) & 0.807  (-)  & 0.805  (-) & 0.824  (-) & 0.799  (-) \\
                    654     & 0.903   & 0.886  (=) & 0.911  (=) & 0.874  (-) & 0.886  (=) & 0.881  (-) & 0.89  (-)  & 0.864  (-)  & 0.849  (-) & 0.864  (-) & 0.821  (-) \\
                    656     & 0.825   & 0.637  (-) & 0.693  (-) & 0.623  (-) & 0.621  (-) & 0.596  (-) & 0.577  (-) & 0.582  (-)  & 0.577  (-) & 0.51  (-) & 0.495  (-) \\
                    657     & 0.821   & 0.797  (-) & 0.758  (-) & 0.774  (-) & 0.786  (-) & 0.776  (-) & 0.782  (-) & 0.757  (-)  & 0.789  (-) & 0.739  (-) & 0.73  (-) \\
                    658     & 0.781   & 0.749  (=) & 0.733  (=) & 0.735  (-) & 0.76  (-)  & 0.668  (-) & 0.678  (-) & 0.578  (-)  & 0.616  (-) & 0.624  (-) & 0.6  (-) \\
                    659     & 0.651   & 0.546  (=) & 0.605  (=) & 0.682  (=) & 0.648  (=) & 0.703  (=) & 0.632  (=) & 0.6  (=)    & 0.543  (=) & 0.65  (=) & 0.718  (=) \\
                    663     & 0.994   & 0.997  (+) & 0.996  (+) & 0.995  (+) & 0.996  (+) & 0.996  (=) & 0.996  (+) & 0.994  (=)  & 0.994  (=) & 0.995  (=) & 0.993  (=) \\
                    665     & 0.29    & 0.245  (=) & 0.246  (=) & 0.286  (=) & 0.3  (=)   & 0.275  (+) & 0.315  (+) & 0.235  (=)  & 0.301  (=) & 0.279  (=) & 0.35  (+) \\
                    666     & 0.578   & 0.587  (=) & 0.543  (=) & 0.565  (=) & 0.578  (=) & 0.593  (=) & 0.563  (=) & 0.592  (=)  & 0.591  (=) & 0.561  (=) & 0.578  (=) \\
                    678     & 0.231   & 0.082  (=) & 0.131  (=) & 0.171  (=) & 0.083  (=) & 0.212  (=) & 0.291  (=) & 0.199  (=)  & 0.177  (=) & 0.188  (=) & 0.214  (=) \\
                    687     & 0.28    & 0.309  (=) & 0.176  (=) & 0.349  (=) & 0.325  (=) & 0.374  (=) & 0.43  (=)  & 0.379  (+)  & 0.369  (=) & 0.396  (=) & 0.352  (=) \\
                    690     & 0.968   & 0.968  (=) & 0.968  (=) & 0.964  (=) & 0.965  (=) & 0.966  (=) & 0.967  (=) & 0.963  (-)  & 0.964  (=) & 0.965  (=) & 0.961  (-) \\
                    695     & 0.86    & 0.834  (-) & 0.837  (-) & 0.855  (=) & 0.846  (=) & 0.852  (=) & 0.833  (=) & 0.847  (=)  & 0.832  (-) & 0.817  (-) & 0.857  (=) \\
                    706     & 0.617   & 0.517  (=) & 0.595  (=) & 0.576  (=) & 0.561  (=) & 0.536  (=) & 0.59  (=)  & 0.635  (=)  & 0.647  (=) & 0.519  (=) & 0.639  (=) \\
                    712     & 0.758   & 0.737  (=) & 0.757  (=) & 0.755  (=) & 0.766  (=) & 0.745  (=) & 0.756  (=) & 0.76  (=)   & 0.746  (=) & 0.735  (=) & 0.762  (=) \\
                    1027    & 0.861   & 0.857  (=) & 0.861  (=) & 0.859  (=) & 0.864  (=) & 0.86  (=)  & 0.855  (=) & 0.863  (=)  & 0.857  (=) & 0.862  (=) & 0.855  (=) \\
                    1028    & 0.368   & 0.394  (+) & 0.397  (+) & 0.371  (+) & 0.4  (+)   & 0.395  (+) & 0.403  (+) & 0.408  (+)  & 0.393  (+) & 0.4  (+) & 0.385  (+) \\
                    1029    & 0.559   & 0.545  (=) & 0.561  (=) & 0.556  (=) & 0.55  (=)  & 0.558  (=) & 0.548  (=) & 0.554  (=)  & 0.55  (=) & 0.55  (=) & 0.555  (=) \\
                    1030    & 0.355   & 0.364  (=) & 0.36  (=)  & 0.356  (=) & 0.361  (=) & 0.366  (=) & 0.362  (=) & 0.36  (=)   & 0.362  (=) & 0.354  (=) & 0.354  (=) \\
                    1089    & 0.767   & 0.777  (=) & 0.736  (=) & 0.763  (=) & 0.75  (=)  & 0.735  (=) & 0.753  (=) & 0.741  (=)  & 0.788  (=) & 0.704  (=) & 0.757  (=) \\
                    1096    & 0.791   & 0.818  (=) & 0.749  (=) & 0.861  (=) & 0.774  (=) & 0.828  (+) & 0.835  (=) & 0.756  (=)  & 0.79  (=) & 0.708  (=) & 0.771  (=) \\
                    1191    & 0.303   & 0.318  (+) & 0.309  (=) & 0.304  (=) & 0.307  (=) & 0.308  (+) & 0.315  (+) & 0.307  (=)  & 0.312  (+) & 0.317  (+) & 0.3  (-) \\
                    1193    & 0.573   & 0.571  (=) & 0.571  (=) & 0.567  (-) & 0.568  (-) & 0.571  (=) & 0.581  (+) & 0.57  (-)   & 0.577  (=) & 0.579  (+) & 0.565  (-) \\
                    1196    & 0.46    & 0.467  (+) & 0.458  (=) & 0.459  (=) & 0.46  (=)  & 0.458  (=) & 0.47  (+)  & 0.456  (=)  & 0.463  (=) & 0.464  (=) & 0.449  (-) \\
                    1199    & 0.429   & 0.437  (+) & 0.433  (+) & 0.434  (+) & 0.434  (+) & 0.438  (+) & 0.436  (+) & 0.43  (=)   & 0.436  (+) & 0.435  (+) & 0.431  (=) \\
                    1201    & 0.097   & 0.102  (+) & 0.1  (+)   & 0.097  (=) & 0.098  (=) & 0.096  (=) & 0.099  (=) & 0.093  (=)  & 0.095  (=) & 0.1  (=) & 0.093  (=) \\
                    1203    & 0.561   & 0.574  (+) & 0.575  (=) & 0.567  (=) & 0.565  (+) & 0.566  (=) & 0.556  (=) & 0.543  (-)  & 0.561  (=) & 0.574  (=) & 0.535  (-) \\
                    1595    & 0.108   & 0.202  (+) & 0.115  (+) & 0.109  (+) & 0.201  (+) & 0.179  (+) & 0.129  (+) & 0.117  (+)  & 0.124  (=) & 0.075  (=) & 0.217  (+) \\
                    \midrule
                    Summary & 0/116/0 & 13/60/43 & 8/61/47  & 10/37/69 & 9/41/66  & 13/42/61 & 14/32/70 & 7/34/75   & 9/36/71 & 8/35/73 & 7/31/78 \\\bottomrule
                \end{tabular}
            }
        \end{table*}

        \newpage

\end{document}